\documentclass[lettersize,journal]{IEEEtran}
\usepackage{amsmath,amsfonts}
\usepackage{array}
\usepackage[caption=false,font=normalsize,labelfont=sf,textfont=sf]{subfig}
\usepackage{textcomp}
\usepackage{stfloats}
\usepackage{url}
\usepackage{verbatim}
\usepackage{graphicx}
\usepackage{cite}

\usepackage[utf8]{inputenc} 
\usepackage[T1]{fontenc}    
\usepackage[colorlinks=true, citecolor=blue, urlcolor=black]{hyperref} 
\usepackage{booktabs}       
\usepackage{amsfonts}       
\usepackage{graphicx}       
\usepackage{nicefrac}       
\usepackage{microtype}      
\usepackage{xcolor}         
\usepackage{amsmath}
\usepackage{enumitem}
\usepackage[ruled,vlined]{algorithm2e}
\usepackage{tabularx}
\usepackage{multirow}
\usepackage[table]{xcolor}
\usepackage{adjustbox}
\usepackage{array}
\usepackage{wrapfig}
\usepackage{subcaption}
\usepackage{soul}
\usepackage{colortbl}
\usepackage{pgfplots}
\usepackage{tikz}
\usepackage{ifthen}
\usepackage{makecell}
\usepackage{balance}
\usepackage{tcolorbox}
\usepackage{float}

\begin{document}

\title{ControlGS: Consistent Structural Compression Control for Deployment-Aware Gaussian Splatting}

\author{
Fengdi Zhang,~Yibao Sun,~Hongkun Cao,~Ruqi Huang
\thanks{
This work was supported in part by the National Natural Science Foundation of China under grant No. 62171256, 62331006, 62205178.
\textit{(Corresponding authors: Hongkun Cao; Ruqi Huang)}}
\thanks{Fengdi Zhang is with the Shenzhen International Graduate School, Tsinghua University, Shenzhen 518055, China, and also with the Pengcheng Laboratory, Shenzhen 518055, China (e-mail: zhangfd22@mails.tsinghua.edu.cn).}
\thanks{Yibao Sun and Hongkun Cao are with the Pengcheng Laboratory, Shenzhen 518055, China (e-mail: sunyb01@pcl.ac.cn; caohk@pcl.ac.cn).}
\thanks{Ruqi Huang is with the Shenzhen International Graduate School, Tsinghua University, Shenzhen 518055, China (e-mail: ruqihuang@sz.tsinghua.edu.cn).}
}

\markboth{Journal of \LaTeX\ Class Files,~Vol.~XX, No.~XX, XX~20XX}%
{Shell \MakeLowercase{\textit{et al.}}: A Sample Article Using IEEEtran.cls for IEEE Journals}

\IEEEpubid{0000--0000/00\$00.00~\copyright~20XX IEEE}

\maketitle

\begin{abstract}
3D Gaussian Splatting (3DGS) is a highly deployable real-time method for novel view synthesis. In practice, it requires a universal, consistent control mechanism that adjusts the trade-off between rendering quality and model compression without scene-specific tuning, enabling automated deployment across different device performances and communication bandwidths.
In this work, we present ControlGS, a control-oriented optimization framework that maps the trade-off between Gaussian count and rendering quality to a continuous, scene-agnostic, and highly responsive control axis. Extensive experiments across a wide range of scene scales and types (from small objects to large outdoor scenes) demonstrate that, by adjusting a globally unified control hyperparameter, ControlGS can flexibly generate models biased toward either structural compactness or high fidelity, regardless of the specific scene scale or complexity, while achieving markedly higher rendering quality with the same or fewer Gaussians compared to potential competing methods.
Project~page:~\href{https://zhang-fengdi.github.io/ControlGS/}{\textcolor{magenta}{https://zhang-fengdi.github.io/ControlGS}}.
\end{abstract}


\section{Introduction}

\IEEEPARstart{3D} Gaussian splatting (3DGS) \cite{3DGS} has recently shown remarkable performance in novel view synthesis~(NVS). By projecting anisotropic Gaussians onto the image plane and employing efficient $\alpha$-blending, 3DGS achieves a good balance between rendering efficiency and high-fidelity reconstruction, making it a highly deployable method for real-time NVS across a wide range of computing devices, such as smartphones~\cite{phone}, VR/AR headsets~\cite{VR1,VR2}, and edge devices~\cite{edge}. However, to cover fine details and preserve fidelity, 3DGS typically requires millions of Gaussians, causing the model size to expand dramatically, which in turn brings storage overhead, computational burden, and deployment challenges.

To tackle this problem, the community has focused on reducing the number of Gaussians while maintaining rendering quality\cite{LP-3DGS, EAGLES, Mini-Splatting, GaussianSpa, Taming3DGS}, \textit{i.e.}, structural compression. This process naturally introduces the trade-off between \textit{Gaussian count} and \textit{rendering quality}, which has become a central challenge in structural compression. However, in real applications, compression alone is insufficient. Different scenarios, hardware, and deployment goals require different balances between model size and fidelity. Therefore, compared to static compression, a controllable mechanism that allows flexible adjustment between the two is better aligned with practical needs.

\begin{figure*}[!t]
  \centering
  \includegraphics[width=0.97\linewidth]{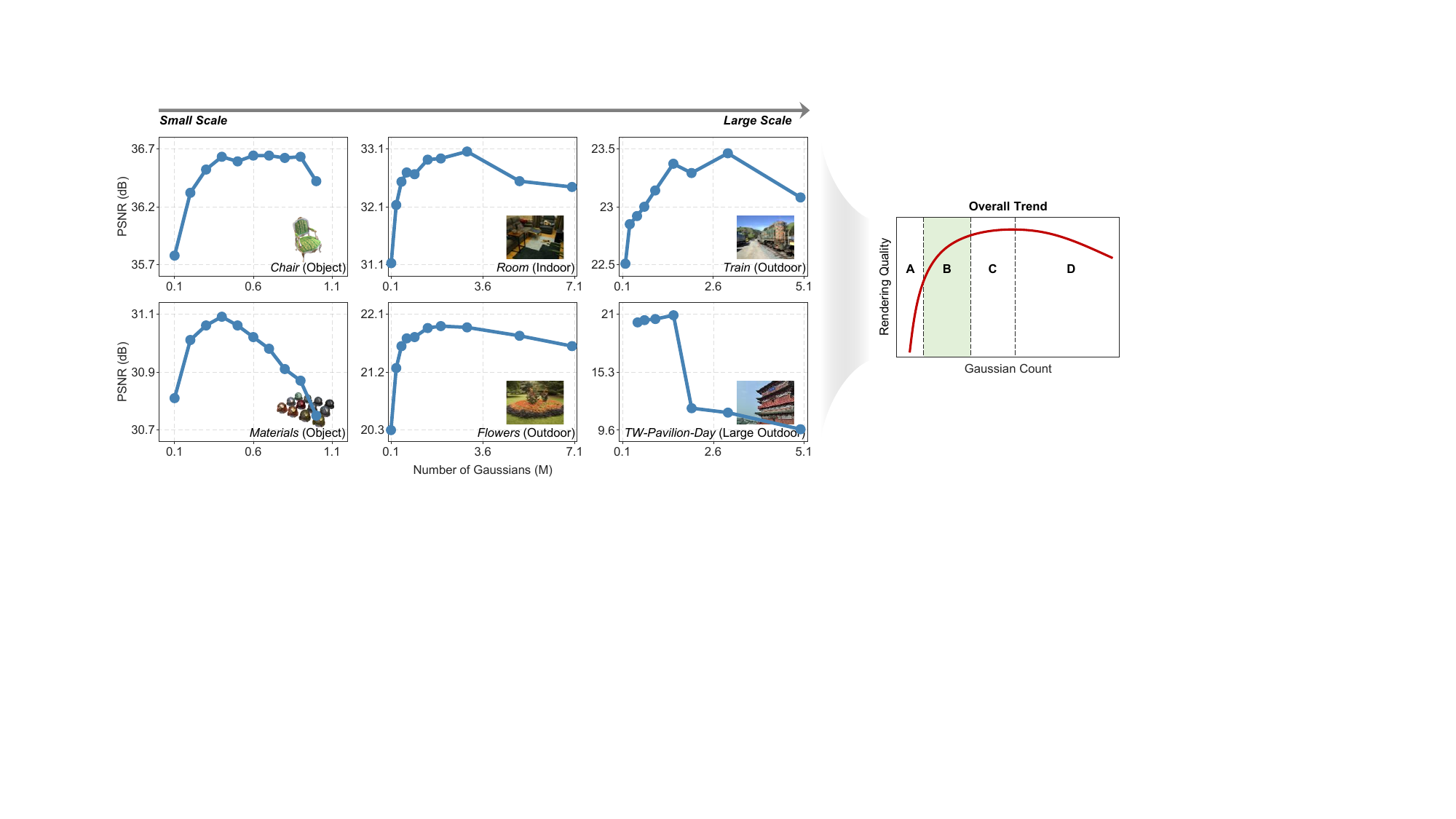}
  \caption{Gaussian count--rendering quality relationship across scenes. \textit{Left six panels}: Empirical curves obtained using the top-performing budget-based method 3DGS-MCMC~\cite{3DGS-MCMC}, covering representative object, indoor, and outdoor scenes from multiple benchmark datasets~\cite{GigaNVS, Mip-NeRF360, TanksAndTemples, NeRF}. Although the absolute number of Gaussians varies across scenes due to differences in scale, the resulting relationships consistently exhibit a universal four-phase pattern.
  \textit{Right}: Conceptual illustration of the four-phase pattern: (A) underfitting, (B) efficient regime, (C) saturation, and (D) overfitting.}
  \label{fig:control}
\end{figure*}

Achieving such controllability requires answering a key question: where should control be applied? Through systematic experiments, as shown in Fig.~\ref{fig:control}, we analyze the Gaussian count--rendering quality curve across scenes of varying scales and complexities, and observe a consistent four-phase pattern: a sharp quality drop under insufficient Gaussians~(\textit{underfitting}), a cost-effective region where quality improves rapidly with relatively few Gaussians~(\textit{efficient regime}), a stage where quality gain diminishes despite rapidly growing Gaussians~(\textit{saturation}), and finally a regime where excessive Gaussians may even harm quality~(\textit{overfitting}). This structure suggests that control is most effectively applied within the \textit{efficient regime}, where the balance between resources and rendering quality is most favorable and rendering quality is most responsive to changes in Gaussian count, ensuring that control remains both meaningful and~impactful.
\IEEEpubidadjcol

While existing approaches have made notable progress toward structural compression, challenges remain in achieving controllability. Budget-based methods constrain only the hard-coded Gaussian count budget or similar counterparts, with no guarantee that training converges to the \textit{efficient regime}. In practice, this also requires repeated trial-and-error per scene to determine the appropriate Gaussian budget~\cite{Mini-Splatting,Mini-Splatting2,SafeguardGS,GaussianSpa,Taming3DGS,3DGS-MCMC}, which is impractical for automated deployment. 
Meanwhile, non-budget-based methods attempt to target the \textit{efficient regime}, but their effectiveness is sensitive to scene scale and complexity, making cross-scene consistency difficult to maintain~\cite{Color-cuedGS,GoDe,EAGLES,LP-3DGS}, thus limiting their practicality in real-world applications without specific scene assumptions. This indicates that the community still lacks a unified mechanism that can consistently hit the \textit{efficient regime} while supporting flexible preference~control.

Motivated by this, we present ControlGS, an optimization framework designed with scene-agnostic structural compression controllability as its primary goal.
ControlGS introduces three key designs: 
(1) \textit{Uniform Splitting}: A global, octree-style splitting strategy replaces heuristic local densification, building a coarse-to-fine optimization path that eliminates sensitivity to scene scale and complexity; 
(2) \textit{Opacity-based Sparsification}: an opacity-based $L_1$ regularization that progressively suppresses and prunes redundant Gaussians while correcting over-splitting;
An opacity-based $L_1$ regularization method progressively suppresses and prunes redundant Gaussians and corrects over-splitting, only based on their contribution to the rendered image;
(3) \textit{Single Hyperparameter Control}: By adjusting the relative weight, \textit{i.e.,} the control hyperparameter, of the above two components in the loss function, ControlGS enables scene-agnostic structural compression control.

Experiments on instances from multiple datasets across a wide range of scales and types (from small objects to large outdoor scenes) demonstrate that, by tuning the control hyperparameter within a predefined globally bounded range, 
ControlGS automatically learns an appropriate number of Gaussians across diverse scenes, consistently anchors the solution in the \textit{efficient regime}, and provides users with continuous, scene-agnostic, and highly responsive preference adjustment between \textit{fewer Gaussians} and \textit{higher rendering quality}. 
Moreover, ControlGS achieves a better Pareto frontier between Gaussian count and rendering quality than potential competing methods.

The contributions of this work are summarized as follows:  
\begin{enumerate}
    \item We investigate the Gaussian count--rendering quality behavior of 3DGS models across a wide range of scene scales, from small objects to large outdoor scenes, revealing a universal four-phase pattern and identifying the \textit{efficient regime} as the optimal control target;
    \item We present a control-oriented 3DGS optimization framework that consistently converges to the \textit{efficient regime} across scenes of varying scales and types;
    \item Our method enables continuous, scene-agnostic, and highly responsive preference control between model compactness and rendering quality within the \textit{efficient regime} across diverse scenes;
    \item Our method achieves markedly higher rendering quality with the same or fewer Gaussians, outperforming potential competing methods.
\end{enumerate}

\section{Related Work}

\subsection{Novel View Synthesis}

Novel view synthesis~(NVS) aims to generate images of a scene or object from unseen viewpoints using existing images.
NeRF~\cite{NeRF} employs MLP-based implicit 3D representations and differentiable volume rendering for consistent multi-view synthesis, but at high computational cost. Although later works improve speed~\cite{FastNeRF, PlenOctrees, Plenoxels, InstantNGP}, they still depend on dense sampling and costly neural inference, limiting their ability to balance efficiency and fidelity in high-resolution or large-scale scenes.
3DGS~\cite{3DGS} mitigates this by introducing anisotropic 3D Gaussians and replacing ray marching with Gaussian projection and $\alpha$-blending, substantially improving efficiency while enabling real-time, high-quality NVS.

\subsection{3DGS Compression}

While 3DGS offers clear advantages in speed and rendering quality, its explicit representation leads to high storage overhead, now a key bottleneck. This has made 3DGS compression a major research focus.
Current methods fall into two approaches: structural compression and attribute compression.
Structural compression~\cite{Mini-Splatting,Mini-Splatting2,SafeguardGS,GaussianSpa,Taming3DGS,Color-cuedGS,GoDe,EAGLES,LP-3DGS} focuses on reducing the number of Gaussians to fundamentally shrink model size.

Attribute compression includes adding neural components~\cite{Scaffold-GS,Octree-GS,CompGSLiu,Compact3D}, simplifying SH~\cite{Reduced-3DGS,LightGaussian,EfficientGS,RDOGaussian,GoDe}, applying quantization~\cite{SOG,Reduced-3DGS,LightGaussian,Compact3D,CompGSNavaneet,ELMGS,RDOGaussian,PUP3D-GS,TrimmingTheFat,GoDe}, and using entropy coding~\cite{Compact3D,ELMGS,RDOGaussian} to reduce the storage overhead of each Gaussian's attributes.

\subsection{3DGS Structural Compression Control}

3DGS models face an inherent trade-off between Gaussian count and rendering quality: more Gaussians improve rendering quality but reduce compressibility, while fewer enhance compression at the cost of rendering quality.
Based on this, structural compression control aims to adjust the preference between Gaussian count and rendering quality by tuning hyperparameters during training or post-processing, enabling deployable models tailored to specific resource or application needs. Existing approaches fall into budget-based and non-budget-based categories.
Budget-based methods~\cite{Mini-Splatting,Mini-Splatting2,SafeguardGS,GaussianSpa,Taming3DGS,3DGS-MCMC} prune less important Gaussians using a hard-coded Gaussian count budget or similar counterparts, often requiring repeated tuning for specific scenes to select a suitable model. This limits their practicality, as such tuning processes are impractical in automated deployment.
Non-budget-based methods~\cite{SOG,LightGaussian,RDOGaussian,Compact3D,Reduced-3DGS,EAGLES,CompGSNavaneet,Color-cuedGS,GoDe,LP-3DGS} simplify this scene-specific tuning process, improving automation and efficiency. However, they often struggle to maintain consistent behavior across scenes, compromising their generalizability in real-world applications. Additionally, they tend to under-utilize the contribution of each Gaussian, leading to low Gaussian efficiency, \textit{i.e.,} using more Gaussians but achieving suboptimal rendering quality.

\section{Motivation}

Our goal is to allow users to consistently and highly responsively adjust preferences across scenes, choosing between larger models for higher rendering quality and smaller, lightweight models, with the Gaussian count automatically determined by the algorithm. We also aim to ensure high Gaussian efficiency, \textit{i.e.,} achieving better rendering quality with fewer Gaussians.

To this end, we first conduct experiments across a wide range of scene scales and types to systematically summarize the relationship between the number of Gaussians and rendering quality, as shown in Fig.~\ref{fig:control}. We denote the curve as $\mathcal{R}(N,S)$, where $N$ is the number of Gaussians and $S$ is the scene.
We observe the existence of an \textit{efficient regime} $[N^\star_{\min}(S), N^\star_{\max}(S)]$, within which Gaussian efficiency is highest, and rendering quality is most sensitive to $N$. 
Therefore, the question becomes: how can we make the model stably converge to the \textit{efficient regime} and support continuous, predictable preference control?

For simplicity, we first fix a scene $S$ and consider single-scene control. During 3DGS optimization, the number of Gaussians is jointly determined by two opposing mechanisms: the densification mechanism $\mathcal{D}$ encourages adding new Gaussians to improve rendering quality, while the pruning mechanism $\mathcal{P}$ tends to remove redundant Gaussians to compress the model. We introduce a control hyperparameter $\lambda$ to balance the relative strengths of the two mechanisms, so that the final Gaussian count naturally emerges from their dynamic interplay:
\begin{equation}
\Psi(N;\lambda) = F_{\mathcal{D}}(N) + \lambda F_{\mathcal{P}}(N),
\label{eq:control_function}
\end{equation}
where $F_{\mathcal{D}}(N)$ and $F_{\mathcal{P}}(N)$ represent the objectives associated with $\mathcal{D}$ and $\mathcal{P}$, respectively. The equilibrium Gaussian count is then given by:
\begin{equation}
N_{\mathrm{eq}}(\lambda) = \arg\min_{N \in [0, N_{\max}]} \Psi(N; \lambda).
\end{equation}
When $\lambda \to 0$, densification dominates and $N_{\mathrm{eq}} \to N_{\max}$; when $\lambda$ is large, pruning dominates and $N_{\mathrm{eq}} \to 0$. Thus, $N_{\mathrm{eq}}(\lambda)$ is a continuous and monotonically decreasing function of $\lambda$. By the Intermediate Value Theorem, for any target $N^\star \in [N^\star_{\min}, N^\star_{\max}]$, there exists a unique $\lambda^\star \in [\lambda^\star_{\min}, \lambda^\star_{\max}]$ such that $N_{\mathrm{eq}}(\lambda^\star) = N^\star$. By tuning $\lambda$ within this interval, continuous and predictable preference adjustment can be achieved in the \textit{efficient regime}.

To extend this mechanism across scenes, it is sufficient to reduce the dependence of $\mathcal{D}$ and $\mathcal{P}$ on the scene $S$, such that:
\begin{equation}
F_{\mathcal{D}}(N,S) \approx F_{\mathcal{D}}(N), \quad F_{\mathcal{P}}(N,S) \approx F_{\mathcal{P}}(N).
\end{equation}
In other words, similar adjustment behavior should be maintained across different scenes. To this end, we design $\mathcal{D}$ and $\mathcal{P}$ to be scene-independent, thereby ensuring consistent control responses across $S$. Consequently, a single global $\lambda$ is sufficient to anchor models in their respective \textit{efficient regime} $[N^\star_{\min}(S), N^\star_{\max}(S)]$ and achieve cross-scene consistent structural compactness--fidelity preference control.

Therefore, our task reduces to formulating scene-independent strategies for (1) densification and (2) pruning, and (3) unifying their objectives under a single relative weight $\lambda$ as the control hyperparameter during training.

\section{Method}

\subsection{Preliminaries}
\label{sec:preliminaries}
3DGS~\cite{3DGS} explicitly represents a scene using anisotropic 3D Gaussians and enables real-time rendering through efficient differentiable splatting. The process begins by reconstructing a sparse point cloud using structure-from-motion~(SfM)~\cite{SfM}, which is then used to initialize a set of 3D Gaussians.
Each Gaussian is defined by a set of attribute parameters: center position $p$, opacity $\alpha$, spherical harmonic coefficients $c$ for color representation, and a covariance matrix $\Sigma$ that encodes its spatial extent. For differentiable optimization, the covariance matrix $\Sigma$ is further parameterized by a scaling matrix $S$ and a rotation matrix $R$.

During rendering, 3D Gaussians are projected onto the 2D image plane, and blended via $\alpha$-blending to produce the final pixel color. The pixel color $C$ is computed by blending $N$ overlapping Gaussians~as: 
\begin{equation}
C = \sum_{i \in N} c_i \alpha_i \prod_{j=1}^{i-1}(1 - \alpha_j), 
\label{eq:alpha-blending}
\end{equation} 
where $c_i$ is the color of the $i$-th Gaussian determined by its spherical harmonic coefficients, and $\alpha_i$ is obtained by evaluating a 2D Gaussian from its covariance matrix $\Sigma_i$ scaled by a learned opacity.
The Gaussian parameters are then optimized by minimizing a loss that combines an $\mathcal{L}_1$ term and a differentiable structural similarity index metric~(D-SSIM)~\cite{SSIM} between the rendered outputs and the ground-truth views: 
\begin{equation}
\mathcal{L}_\mathrm{RGB} = (1 - \lambda_{w}) \mathcal{L}_1 + \lambda_{w}\mathcal{L}_{\mathrm{D\text{-}SSIM}},
\end{equation}
where the weight $\lambda_{w}$ is set to 0.2 in 3DGS~\cite{3DGS}.

\begin{figure*}[!t]
  \centering
  \includegraphics[width=0.99\linewidth]{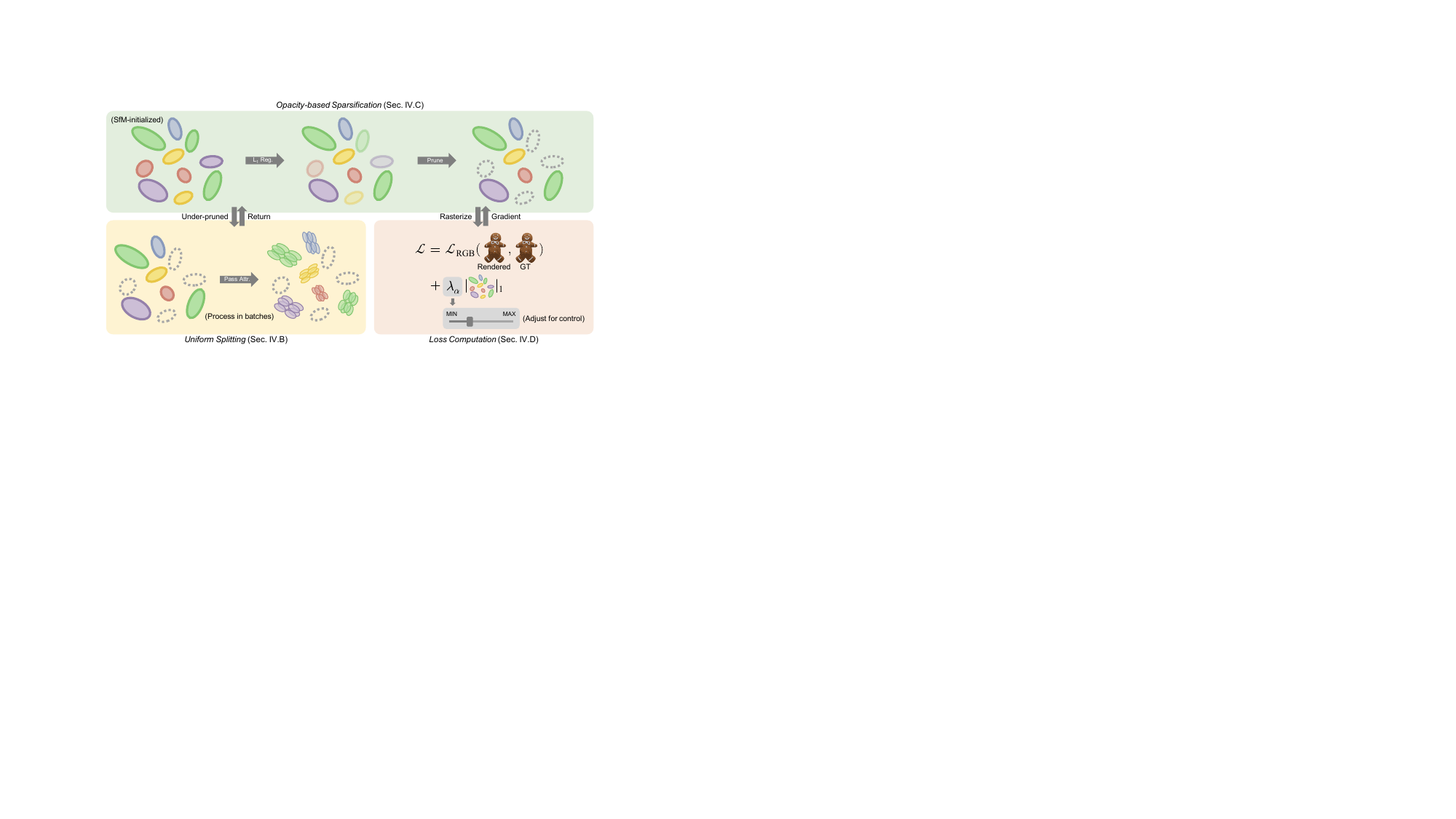}
  \caption{Overview of the ControlGS pipeline. Training starts from an SfM-initialized Gaussian set and proceeds with RGB reconstruction and opacity-based sparsification, which prune low-contribution Gaussians and compact the representation. When pruning saturates, indicating convergence at the current resolution, uniform splitting expands candidates and refines structure, after which optimization resumes and the cycle repeats. A smaller $\lambda_\alpha$ retains more candidates for higher rendering quality, whereas a larger $\lambda_\alpha$ enforces stronger sparsification with fewer Gaussians, enabling consistent structural compression control across scenes.}
  \label{fig:overview}
\end{figure*}

\subsection{Densification: Uniform Splitting}
\label{sec:uniform_Gaussian_branching}


Developing a scene-independent densification method requires first identifying why existing approaches remain scene-dependent. In 3DGS and its variants \cite{3DGS, LP-3DGS, EAGLES}, densification is typically applied to only a subset of Gaussians, selected based on local heuristics. These include thresholds on accumulated normalized gradients and axis lengths relative to the scene’s maximum radius. Such criteria inherently depend on the scene’s scale and complexity, making the densification behavior vary across scenes and thereby undermining consistency across different settings.

Our solution is straightforward: we uniformly split \textit{all} existing Gaussians, without selecting specific subsets or distinguishing between cloning and splitting. This uniform strategy effectively decouples the densification process from scene-dependent heuristics, ensuring consistent behavior across diverse scenes.

Specifically, in each densification step, we apply an octree-style split to every existing Gaussian to ensure visual consistency and maintain a stable optimization trajectory. As bisecting each spatial axis naturally yields $2^3 = 8$ subdivisions, this approach provides uniform spatial coverage and unbiased directional exploration. The positions of the eight child Gaussians are given by
\begin{equation}
p_{\text{child}, i} = p_{\text{parent}} + R_{\text{parent}}(\Delta_i \odot S_{\text{parent}}),
\end{equation}
where $\Delta_i$ is an offset vector with components $\{\pm 0.25\}^3$; $S_{\text{parent}}$ and $R_{\text{parent}}$ are the parent’s scaling and rotation matrices, respectively; and $\odot$ denotes element-wise multiplication.
Each child Gaussian inherits the scaling matrix from its parent and applies a shrinkage factor following vanilla 3DGS \cite{3DGS}:
\begin{equation}
S_{\text{child}} = S_{\text{parent}}/1.6.
\end{equation}
Since the octree-style splitting causes viewing rays from different directions to intersect approximately two child Gaussians, we compute the opacity of child Gaussians based on $\alpha$-blending to ensure consistent composite opacity before and after the split. Specifically, solving 
$(1 - \alpha_{\text{child}})^2 = 1 - \alpha_{\text{parent}}, $
yields:
\begin{equation}
\alpha_{\text{child}} = 1 - \sqrt{1 - \alpha_{\text{parent}}}.
\end{equation}
Finally, the rotation matrix $R_{\text{child}}$ and spherical harmonic coefficients $c_{\text{child}}$ are directly inherited from the parent Gaussian.

To avoid memory overflow from splitting too many Gaussians at once, we perform the uniform splitting in batches by randomly selecting $N_{\text{batch}}$ Gaussians without replacement. After each batch is split, a brief optimization phase prunes redundant Gaussians to free memory. This process iterates until all Gaussians are processed in the current splitting step.

\subsection{Pruning: Opacity-based Sparsification}
\label{sec:Gaussian_atrophy}

In 3DGS and its variants, Gaussian pruning is also guided by scene-dependent heuristics, such as thresholds on projection radius or axis length relative to the scene scale. However, the ultimate goal of 3DGS optimization is to improve rendering quality. Thus, whether a Gaussian should be retained should be decided directly by its actual contribution to the rendered image. As shown in Eq.~\eqref{eq:alpha-blending}, the opacity $\alpha$ serves as the blending weight of a Gaussian, directly reflecting its contribution to pixel colors. Based on this observation, we adopt opacity-based sparsification as a unified pruning criterion by introducing the following $L_1$ regularization term:
\begin{equation}
\mathcal{L}_\alpha = \sum_i |\alpha_i|.
\end{equation}
During training, we further set a small opacity threshold $\tau_\alpha$ and periodically remove Gaussians with $\alpha_i < \tau_\alpha$, thereby transitioning from soft sparsity~($\mathcal{L}_\alpha$) to hard pruning~($\alpha < \tau_\alpha$).

\subsection{Unifying for Consistent Structural Compression Control}
\label{sec:structural_compression_control}

At this point, increasing and reducing Gaussians are handled by two complementary mechanisms: \textit{uniform splitting}, which enriches details by generating candidates, and \textit{opacity-based sparsification}, which prunes redundancies to maintain compactness. Both eliminate heuristic dependencies on scene scale or complexity, thereby ensuring cross-scene consistency. Accordingly, as discussed in Eq.~\ref{eq:control_function}, we unify them under a single relative weight $\lambda_\alpha$ as the final loss:
\begin{equation}
\mathcal{L}=\mathcal{L}_{\mathrm{RGB}}+\lambda_\alpha\mathcal{L}_\alpha.
\label{eq:final_loss}
\end{equation}
A smaller $\lambda_\alpha$ favors retaining more split candidates and leads to higher rendering quality, while a larger $\lambda_\alpha$ enforces stronger sparsification and yields a more compact representation. Thus, adjusting $\lambda_\alpha$ can navigate the Gaussian count--rendering quality trade-off in a scene-agnostic manner, enabling consistent structural compression control across scenes.

The loss alone is insufficient without a compatible optimization schedule. To this end, as shown in Fig.~\ref{fig:overview}, we adopt a \textit{optimize~(with pruning) $\rightarrow$ split $\rightarrow$ re-optimize $\rightarrow$ re-split} rhythm. Training begins with an SfM-initialized Gaussian set and standard optimization at the current resolution. During optimization, we periodically record the number of Gaussians removed due to $\alpha_i < \tau_\alpha$, denoted $N_{\text{remove}}$. If $N_{\text{remove}}$ remains below a threshold $\tau_{\text{remove}}$, it indicates that redundant Gaussians have been pruned and the model has nearly converged at this scale. At this point, we perform \textit{uniform splitting}, resume optimization, and trigger the next split once $N_{\text{remove}}<\tau_{\text{remove}}$ again. This iterative process drives the model along a ``first prune, then refine'' trajectory, with the rhythm determined by sparsification progress rather than scene-specific tuning.
We summarize the optimization workflow of our method in Algorithm~\ref{alg:controlgs}.

In summary, \textit{uniform splitting} provides unbiased expansion, while \textit{opacity-based sparsification} enforces demand-driven contraction. Together, under a single hyperparameter $\lambda_\alpha$, they form a stable, scene-independent closed loop: first expanding the candidate space to capture detail, then retracting redundant parts. With a single knob, the model can consistently and predictably transition across scenes between a more compact representation and higher rendering quality.

\begin{algorithm}[!h]
\small
\caption{ControlGS Optimization}
\label{alg:controlgs}
\DontPrintSemicolon
\SetArgSty{textnormal}
\SetKwFor{ForEach}{for each}{do}{end}
\SetKwComment{Comment}{$\triangleright$\ }{}

$\mathcal{G} \leftarrow \textsc{InitFromSfM}()$ \;
$t \leftarrow 0$ \Comment*[r]{Iteration Count}
$N_{\text{split}} \leftarrow 0$ \Comment*[r]{Splitting Count}

\While{$t < t_{\text{max}}$}{
    $V,\hat{I} \leftarrow \textsc{SampleView}()$ \Comment*[r]{Camera and Image}
    $I \leftarrow \textsc{Rasterize}(\mathcal{G},V)$ \;
    $\mathcal{L} \leftarrow 
    \textsc{Loss}(I,\hat{I},\lambda_\alpha,\alpha_i)$ \Comment*[r]{Eq.(10)}
    $\mathcal{G} \leftarrow \textsc{Update}(\mathcal{G}, \mathcal{L})$\;
    \If{\textsc{IsPruneStep}$(t)$ \textbf{and} $t \ge t_{\text{until}}$}{
        $\mathcal{G}, N_{\text{remove}} \leftarrow \textsc{Prune}(\mathcal{G}, \tau_\alpha)$ \Comment*[r]{Remove $\alpha_i{<}\tau_\alpha$}
        \If{$N_{\text{remove}} < \tau_{\text{remove}}$ \textbf{or} hasNextBatch}{
            \uIf{$N_{\text{split}} < \tau_{\text{split}}$}{
                $\mathcal{B},\ \text{hasNextBatch} \leftarrow \textsc{NextBatch}(\mathcal{G}, N_{\text{batch}})$\;
                $\mathcal{G}_{\text{child}} \leftarrow \textsc{UniformSplit}(\mathcal{B})$

                $\mathcal{G} \leftarrow \mathcal{G} \setminus \mathcal{B} \cup \mathcal{G}_{\text{child}}$\;
                $t_{\text{until}} \leftarrow t + t_{\text{delay}}$ \Comment*[r]{Delay Pruning}
            }\Else{
                $\lambda_\alpha \leftarrow 0$ \Comment*[r]{Disable Reg.}
            }
            \If{\textbf{not} hasNextBatch}{$N_{\text{split}} \leftarrow N_{\text{split}} + 1$}
        }
    }
    $t \leftarrow t + 1$
}
\Return $\mathcal{G}$
\end{algorithm}

\begin{figure*}[!t]
  \centering
  \includegraphics[width=0.9\linewidth]{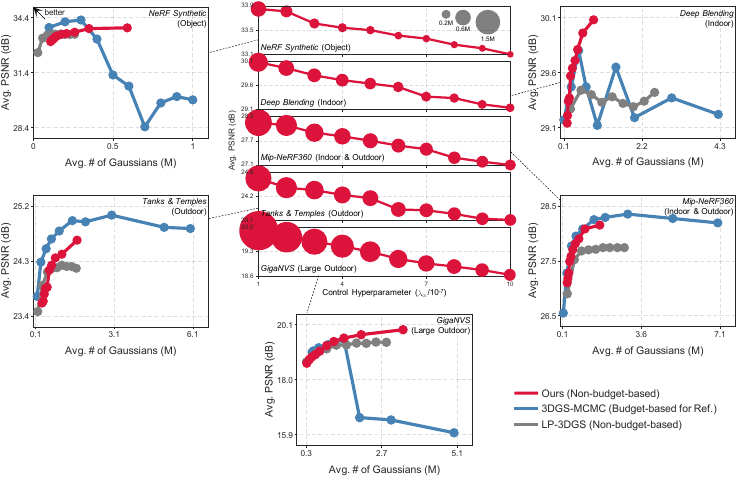}
  \caption{Cross-scene structural compression control results.
  \textit{Center:} Average PSNR of our method versus the control hyperparameter $\lambda_\alpha$; bubble size denotes the average number of Gaussians. As $\lambda_\alpha$ increases, the model becomes more compact, while smaller $\lambda_\alpha$ preserves more Gaussians for higher fidelity. By tuning $\lambda_\alpha$, our method enables smooth and predictable preference control between model compactness and rendering quality.
  \textit{Surrounding plots:} Average PSNR versus average Gaussian count on NeRF Synthetic~\cite{NeRF}~(object), Mip-NeRF360~\cite{Mip-NeRF360}~(indoor/outdoor), Tanks \& Temples~\cite{TanksAndTemples}~(outdoor), Deep Blending~\cite{DeepBlending}~(indoor), and GigaNVS~\cite{GigaNVS}~(large outdoor), comparing our method (\textcolor[HTML]{DC143C}{red}), 3DGS-MCMC~\cite{3DGS-MCMC}~(\textcolor[HTML]{4682B4}{blue}), which is used to obtain the Gaussian count--rendering quality curve \textbf{for reference}, and LP-3DGS~\cite{LP-3DGS}~(\textcolor[HTML]{808080}{gray}).
  All methods are trained for 100k iterations. For ours, $\lambda_\alpha$ ranges from $1\text{e-7}$ to $1\text{e-6}$ with a step of $1\text{e-7}$~(10 control points); LP-3DGS varies its pruning ratio from 0.1 to 0.9 in 0.1 steps~(9 control points). 
  3DGS-MCMC fits the overall Gaussian count--quality curve with scene-dependent budgets~(50k--100k for NeRF Synthetic, 100k--700k for others). 
  ControlGS consistently reaches the \textit{efficient regime} and delivering comparable or higher PSNR with fewer Gaussians than LP-3DGS, while LP-3DGS saturates early and 3DGS-MCMC requires scene-specific~tuning.}
  \label{fig:count_quality_control}
\end{figure*}

\begin{figure*}[!t]
  \centering
  \includegraphics[width=0.99\linewidth]{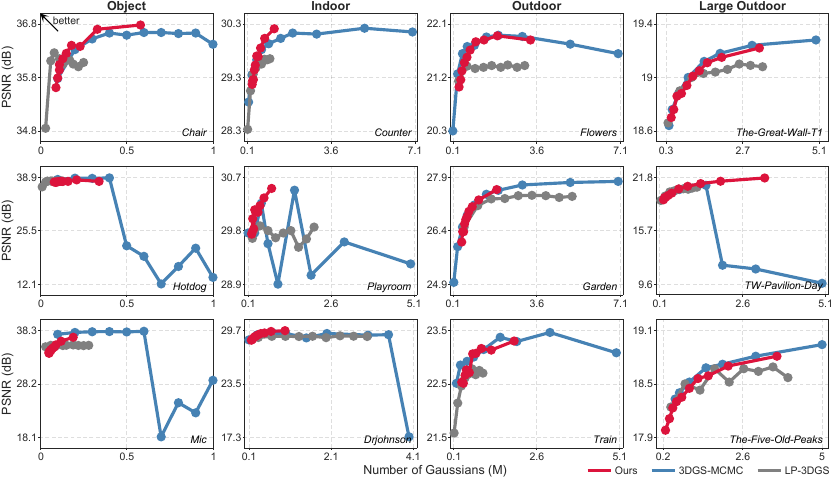}
  \vspace{-4pt}
  \caption{Gaussian count--rendering quality~(PSNR) relationships across representative scenes of different scales from multiple benchmark datasets~\cite{GigaNVS, Mip-NeRF360, TanksAndTemples, NeRF}. Results are shown for our method~(\textcolor[HTML]{DC143C}{red}), the budget-based method 3DGS-MCMC~\cite{3DGS-MCMC} (\textcolor[HTML]{4682B4}{blue}), and the non-budget-based method LP-3DGS~\cite{LP-3DGS} (\textcolor[HTML]{808080}{gray}). 
  The experimental settings follow the same configuration as in Fig.~\ref{fig:count_quality_control}.
  }
  \vspace{-2pt}
  \label{fig:control_performance}
\end{figure*}

\section{Experiments}

\subsection{Experimental Settings}
\label{sec:experimental_settings}
\subsubsection{Dataset and Metrics}  

We comprehensively evaluate our method across 24~instances spanning diverse spatial scales and types, including objects, bounded indoor scenes, unbounded outdoor scenes, and large cross-scale outdoor scenes. The evaluation includes 8~objects from the NeRF synthetic dataset~\cite{NeRF}, 9~indoor/outdoor scenes from the Mip-NeRF360 dataset~\cite{Mip-NeRF360}, 2~outdoor scenes from the Tanks \& Temples dataset~\cite{TanksAndTemples}, 2~indoor scenes from the Deep Blending dataset~\cite{DeepBlending}, and 3~large outdoor scenes from the GigaNVS dataset~\cite{GigaNVS}.
Following the 3DGS evaluation protocol, we adopt the Mip-NeRF360 data split, selecting every eighth frame for testing.
We report peak signal-to-noise ratio (PSNR), structural similarity index metric (SSIM)~\cite{SSIM}, learned perceptual image patch similarity (LPIPS)~\cite{LPIPS}, and the number of Gaussians used in each model to assess the trade-off between model compactness and rendering quality.

\subsubsection{Baselines}

We compare ControlGS with vanilla 3DGS~\cite{3DGS}, a representative budget-based method 3DGS-MCMC~\cite{3DGS-MCMC}~(SfM-initialized), and a range of non-budget-based methods, including LP-3DGS~\cite{LP-3DGS} (using RadSplat scores~\cite{RadSplat}), SOG~\cite{SOG}, LightGaussian~\cite{LightGaussian}, RDOGaussian~\cite{RDOGaussian}, Compact3D~\cite{Compact3D}, Reduced-3DGS~\cite{Reduced-3DGS}, EAGLES~\cite{EAGLES}, CompGS~\cite{CompGSNavaneet}, Color-cued GS~\cite{Color-cuedGS}, GoDe~\cite{GoDe} (GoDe post-processing vanilla 3DGS models~\cite{3DGS}), and GoDe-M~\cite{GoDe} (GoDe post-processing 3DGS-MCMC models~\cite{3DGS-MCMC}).
For the budget-based method 3DGS-MCMC, we report performance under various Gaussian budget settings to fit the Gaussian count--rendering quality curve as a reference; for LP-3DGS, we present results with different pruning ratios ($\rho$); and for GoDe, we evaluate performance under various level-of-detail (LoD) settings.

\subsubsection{Implementation Details}

Our method is implemented on top of the 3DGS framework~\cite{3DGS}. We follow default 3DGS settings for data loading, parameter initialization, learning rate scheduling, optimizer selection, dynamic SH degree promotion, and rendering, with exposure compensation disabled. Experiments are conducted on an Intel Core i9-10980XE CPU and an NVIDIA RTX 3090 GPU.
For our method, a \textit{single} hyperparameter configuration is used across \textit{all} experiments. The hyperparameter $\lambda_\alpha$ controls the structural compression strength, with specific values reported in the experimental~results. 

For our method, pruning is performed every 100 iterations, removing Gaussians with opacity below $\tau_\alpha = 0.005$. When the number of removed Gaussians falls below $N_{\text{remove}} = 2000$, a uniform splitting step is triggered. The splitting is done in batches, with 100k Gaussians processed per batch, and pruning is applied between every two splitting batches. To prevent unstable pruning after splitting, pruning is delayed by 200 iterations after each splitting. The maximum number of splitting rounds is set to 6. Once the maximum splitting rounds are reached, if pruning removes fewer than $N_{\text{remove}}$ Gaussians again, $\lambda_\alpha$ is set to zero until the optimization reaches the maximum number of iterations.

\subsection{Results and Evaluation}

\subsubsection{Structural Compression Control Analysis}

To systematically validate ControlGS in structural compression control, we first identify that when the control hyperparameter $\lambda_\alpha \in [\text{1e-7}, \text{1e-6}]$, the optimization results are anchored in the \textit{efficient regime}. Within this range, we analyze its effect on PSNR and Gaussian count, as shown in Fig.~\ref{fig:count_quality_control} and~\ref{fig:control_performance}.

The analysis reveals that ControlGS exhibits consistent and predictable control behavior across all scenes. As $\lambda_\alpha$ increases, it smoothly strengthens the structural compression with high responsiveness, leading to a monotonic decrease in Gaussian count and a corresponding decline in rendering quality, without stagnation or abrupt fluctuations.

In contrast, LP-3DGS~\cite{LP-3DGS}, though capable of performing non-budget-based structural compression control by adjusting the pruning ratio, struggles to consistently anchor optimization within the \textit{efficient regime} across scenes. It often shows performance saturation within its control range, and in some cases even performance degradation due to overfitting.
As a budget-based method, 3DGS-MCMC~\cite{3DGS-MCMC} requires manual specification of the exact Gaussian count, making cross-scene consistent compression control infeasible. In practice, it demands per-scene parameter tuning to achieve optimal results; otherwise, it easily falls into underfitting or overfitting regions, leading to a significant performance drop.

We further deployed ControlGS models trained with different $\lambda_\alpha$ values to a browser-based client runtime (see our project page) and tested their real-time rendering performance on three widely used integrated GPUs representing high-, mid-, and low-end hardware tiers, as summarized in Table~\ref{tab:igpu_fps}. Results show that assigning $\lambda_\alpha\text{=1e-7}$, $\text{4e-7}$, and $\text{7e-7}$ to the respective tiers yields stable rendering above 25 frames per second (FPS), a typical lower bound for perceptually smooth playback~\cite{FPS25}, without any scene-specific Gaussian budget tuning. These results demonstrate the potential of ControlGS for automated cross-device deployment, enabled by its consistent structural compression control across diverse~scenes.

\newcolumntype{C}[1]{>{\centering\arraybackslash}m{#1}}
\setlength{\tabcolsep}{0pt}
\setlength{\abovetopsep}{1.5pt}    
\setlength{\belowrulesep}{1.5pt}   
\setlength{\aboverulesep}{1.5pt}   
\setlength{\belowbottomsep}{1.5pt} 
\renewcommand{\arraystretch}{0.75}   
\setlength{\extrarowheight}{3pt}     
\begin{table}[h]
\vspace{5pt}
\caption{Rendering speed (FPS) of our method under different $\lambda_\alpha$ on integrated GPUs~(iGPUs): high-end~(AMD Radeon 780M), mid-range~(Intel UHD Graphics 770), and entry-level~(AMD Radeon RX Vega 7).}
\label{tab:igpu_fps}
\centering
\scriptsize
\setlength{\tabcolsep}{0pt}
\renewcommand{\arraystretch}{0.75}
\setlength{\extrarowheight}{3pt}
\begin{tabular}{C{1.4cm}C{2.1cm}|C{0.7cm}C{0.7cm}C{0.7cm}C{0.7cm}C{0.7cm}C{0.7cm}C{0.7cm}}
\toprule
\textbf{GPU Tier} & \textbf{Instance $\mid \lambda_\alpha$}
& \textbf{1e-7} & \textbf{2e-7} & \textbf{3e-7} & \textbf{4e-7} 
& \textbf{5e-7} & \textbf{6e-7} & \textbf{7e-7} \\
\midrule
\multirow{6}{*}{\makecell[c]{\textbf{High-end}\\\textbf{iGPU}}}
  & Bicycle (Outdoor) & \underline{42.2} & 56.2 & 65.8 & 77.5 & 78.7 & 88.6 & 96.3 \\
  & Truck (Outdoor)   & \underline{43.8} & 51.2 & 60.0 & 63.0 & 68.0 & 73.8 & 77.4 \\
  & Bonsai (Indoor)   & \underline{41.3} & 51.2 & 55.1 & 60.9 & 67.6 & 66.9 & 70.8 \\
  & Room (Indoor)     & \underline{44.5} & 51.4 & 57.0 & 61.0 & 62.9 & 65.8 & 68.8 \\
  & Hotdog (Object)   & \underline{74.0} & 94.0 & 106.2 & 116.0 & 118.0 & $>$120 & $>$120 \\
  & Lego (Object)     & \underline{55.6} & 96.4 & $>$120 & $>$120 & $>$120 & $>$120 & $>$120 \\
\midrule
\multirow{6}{*}{\makecell[c]{\textbf{Mid-range}\\\textbf{iGPU}}}
  & Bicycle (Outdoor) & 13.9 & 21.1 & 26.4 & \underline{31.7} & 34.1 & 39.0 & 44.0 \\
  & Truck (Outdoor)   & 11.9 & 17.9 & 22.5 & \underline{25.1} & 26.6 & 27.7 & 29.6 \\
  & Bonsai (Indoor)   & 15.6 & 20.3 & 23.6 & \underline{26.2} & 29.0 & 29.5 & 31.8 \\
  & Room (Indoor)     & 17.1 & 21.3 & 24.7 & \underline{27.7} & 28.1 & 30.0 & 31.2 \\
  & Hotdog (Object)   & 37.5 & 45.9 & 49.8 & \underline{54.0} & 54.7 & 58.5 & 59.6 \\
  & Lego (Object)     & 28.3 & 41.8 & 50.0 & \underline{58.5} & 65.9 & 73.0 & 74.9 \\
\midrule
\multirow{6}{*}{\makecell[c]{\textbf{Entry-level}\\\textbf{iGPU}}}
  & Bicycle (Outdoor) & 17.1 & 21.5 & 24.2 & 28.9 & 29.7 & 32.9 & \underline{37.0} \\
  & Truck (Outdoor)   & 16.1 & 18.6 & 21.8 & 23.1 & 24.5 & 26.4 & \underline{28.2} \\
  & Bonsai (Indoor)   & 14.4 & 17.8 & 20.2 & 21.8 & 24.3 & 24.7 & \underline{26.2} \\
  & Room (Indoor)     & 15.9 & 18.8 & 20.6 & 22.0 & 22.5 & 24.2 & \underline{25.2} \\
  & Hotdog (Object)   & 28.1 & 35.3 & 39.2 & 44.9 & 45.4 & 47.2 & \underline{48.1} \\
  & Lego (Object)     & 28.1 & 39.3 & 48.0 & 56.4 & 57.1 & 59.3 & \underline{60.4} \\
\bottomrule
\end{tabular}
\vspace{-15pt}
\end{table}

\definecolor{first}{HTML}{FF9999}
\definecolor{second}{HTML}{FFCC99}
\definecolor{third}{HTML}{FFF8AD}
\definecolor{green}{HTML}{A9D08E}

\newcommand{\barChartA}[1]{%
  \smash{\pgfmathsetmacro{\barlen}{#1/2.63*0.915}%
  \begin{tikzpicture}[baseline=(base)] 
    \node[inner sep=0pt] (base) at (0.9cm, 0.053cm) {};
    \fill[green] (0,0) rectangle (\barlen cm, 0.27cm);
    \node[anchor=center, font=\scriptsize] at (0.45cm, 0.14cm) {#1};
  \end{tikzpicture}%
}}

\newcommand{\barChartB}[1]{%
  \smash{\pgfmathsetmacro{\barlen}{#1/1.68*0.915}%
  \begin{tikzpicture}[baseline=(base)] 
    \node[inner sep=0pt] (base) at (0.9cm, 0.053cm) {};
    \fill[green] (0,0) rectangle (\barlen cm, 0.27cm);
    \node[anchor=center, font=\scriptsize] at (0.45cm, 0.14cm) {#1};
  \end{tikzpicture}%
}}

\newcommand{\barChartC}[1]{%
  \smash{\pgfmathsetmacro{\barlen}{#1/2.48*0.915}%
  \begin{tikzpicture}[baseline=(base)] 
    \node[inner sep=0pt] (base) at (0.9cm, 0.053cm) {};
    \fill[green] (0,0) rectangle (\barlen cm, 0.27cm);
    \node[anchor=center, font=\scriptsize] at (0.45cm, 0.14cm) {#1};
  \end{tikzpicture}%
}}

\newcolumntype{C}[1]{>{\centering\arraybackslash}m{#1}}
\setlength{\tabcolsep}{0pt}
\setlength{\abovetopsep}{1.5pt}    
\setlength{\belowrulesep}{1.5pt}   
\setlength{\aboverulesep}{1.5pt}   
\setlength{\belowbottomsep}{1.5pt} 
\renewcommand{\arraystretch}{0.75}   
\setlength{\extrarowheight}{3pt}     

\newtcbox{\topone}{on line, colback=first, colframe=first, boxrule=0pt, arc=0pt, boxsep=1pt, left=2pt, right=2pt, top=1pt, bottom=1pt}
\newtcbox{\toptwo}{on line, colback=second, colframe=second, boxrule=0pt, arc=0pt, boxsep=1pt, left=2pt, right=2pt, top=1pt, bottom=1pt}
\newtcbox{\topthree}{on line, colback=third, colframe=third, boxrule=0pt, arc=0pt, boxsep=1pt, left=2pt, right=2pt, top=1pt, bottom=1pt}
\newtcbox{\greenbar}{on line, colback=green, colframe=green, boxrule=0pt, arc=0pt, boxsep=1pt, left=2pt, right=2pt, top=1pt, bottom=1pt}

\begin{table*}[!t]
\centering
\caption{Comparison with non-budget-based structural compression methods on Mip-NeRF360, Tanks \& Temples, and Deep Blending datasets using PSNR, SSIM, LPIPS, and Gaussian count in millions. \topone{\textbf{Best}}, \toptwo{second-best}, and \topthree{third-best} results are highlighted in color. \greenbar{Horizontal bars} indicate the relative number of Gaussians used. ``$\downarrow$'' or ``$\uparrow$'' indicate
lower or higher values are better. Methods with ``*'' further employ attribute compression.}
\label{tab:scene_comparison}
\scriptsize
\begin{tabular}{C{1.45cm}C{1.3cm}|
  C{0.9cm}C{0.9cm}C{0.9cm}C{1.0cm}|
  C{0.9cm}C{0.9cm}C{0.9cm}C{1.0cm}|
  C{0.9cm}C{0.9cm}C{0.9cm}C{1.0cm}
}
\toprule
\multicolumn{2}{c|}{\textbf{Dataset}}
 & \multicolumn{4}{c|}{\textbf{Mip-NeRF360 (Indoor/Outdoor)}}
 & \multicolumn{4}{c|}{\textbf{Tanks \& Temples (Outdoor)}}
 & \multicolumn{4}{c}{\textbf{Deep Blending (Indoor)}} \\
\multicolumn{2}{c|}{Method $\mid$ Metrics}
 & PSNR$\uparrow$ & SSIM$\uparrow$ & LPIPS$\downarrow$ & Num(M)
 & PSNR$\uparrow$ & SSIM$\uparrow$ & LPIPS$\downarrow$ & Num(M)
 & PSNR$\uparrow$ & SSIM$\uparrow$ & LPIPS$\downarrow$ & Num(M) \\
\midrule
\multicolumn{2}{c|}{3DGS~\cite{3DGS}} & 27.63 & 0.814 & 0.222 & \barChartA{2.63}
     & 23.70 & 0.853 & 0.171 & \barChartB{1.58}
     & 29.88 & \cellcolor{third}0.908 & \cellcolor{second}0.242 & \barChartC{2.48} \\

\multicolumn{2}{c|}{SOG*~\cite{SOG}} & 27.08 & 0.799 & 0.277 & \barChartA{2.18}
    & 23.56 & 0.837 & 0.221 & \barChartB{1.24}
    & 29.26 & 0.894 & 0.336 & \barChartC{0.89} \\

\multicolumn{2}{c|}{LightGaussian*~\cite{LightGaussian}} & 27.24 & 0.810 & 0.273 & \barChartA{2.20}
               & 23.55 & 0.839 & 0.235 & \barChartB{1.21}
               & 29.41 & 0.904 & 0.329 & \barChartC{0.96} \\

\multicolumn{2}{c|}{RDOGaussian*~\cite{RDOGaussian}} & 27.05 & 0.801 & 0.288 & \barChartA{1.86}
            & 23.32 & 0.839 & 0.232 & \barChartB{0.91}
            & 29.72 & 0.906 & 0.318 & \barChartC{1.48} \\

\multicolumn{2}{c|}{Compact3D*~\cite{Compact3D}} & 27.08 & 0.798 & 0.247 & \barChartA{1.39}
          & 23.32 & 0.831 & 0.201 & \barChartB{0.84}
          & 29.79 & 0.901 & 0.258 & \barChartC{1.06} \\

\multicolumn{2}{c|}{Reduced-3DGS*~\cite{Reduced-3DGS}} & 27.19 & 0.810 & 0.267 & \barChartA{1.44}
             & 23.55 & 0.843 & 0.223 & \barChartB{0.66}
             & 29.70 & 0.907 & 0.315 & \barChartC{0.99} \\

\multicolumn{2}{c|}{EAGLES*~\cite{EAGLES}} & 27.18 & 0.810 & 0.231 & \barChartA{1.33}
       & 23.26 & 0.837 & 0.201 & \barChartB{0.65}
       & 29.83 & \cellcolor{second}0.910 & \cellcolor{third}0.246 & \barChartC{1.20} \\

\multicolumn{2}{c|}{CompGS*~\cite{CompGSNavaneet}} 
  & 27.12 & 0.806 & 0.240 & \barChartA{0.85}
  & 23.44 & 0.838 & 0.198 & \barChartB{0.52}
  & \cellcolor{third}29.90 & 0.907 & 0.251 & \barChartC{0.55} \\

\multicolumn{2}{c|}{Color-cued GS~\cite{Color-cuedGS}} & 27.07 & 0.797 & 0.249 & \barChartA{0.65}
              & 23.18 & 0.830 & 0.198 & \barChartB{0.37}
              & 29.71 & 0.902 & 0.255 & \barChartC{0.64} \\
\midrule
\multirow{3}{*}{\hspace{1.4em}GoDe*~\cite{GoDe}}
   & \hspace{0.2em}LoD6 & 27.27 & 0.807 & 0.273 & \barChartA{1.55}
          & 23.76 & 0.839 & 0.231 & \barChartB{0.94}
          & 29.73 & 0.904 & 0.327 & \barChartC{0.93} \\

   & \hspace{0.2em}LoD4 & 27.16 & 0.801 & 0.295 & \barChartA{0.60}
          & 23.66 & 0.832 & 0.245 & \barChartB{0.44}
          & 29.73 & 0.903 & 0.334 & \barChartC{0.49} \\

   & \hspace{0.2em}LoD3 & 26.93 & 0.791 & 0.315 & \barChartA{0.38}
          & 23.48 & 0.824 & 0.259 & \barChartB{0.30}
          & 29.74 & 0.902 & 0.340 & \barChartC{0.36} \\
\midrule
\multirow{3}{*}{\hspace{0.8em}GoDe-M*~\cite{GoDe}} 
     & \hspace{0.2em}LoD6 & 27.42 & 0.815 & 0.263 & \barChartA{1.55}
            & 23.97 & 0.842 & 0.220 & \barChartB{0.94}
            & 29.71 & 0.901 & 0.323 & \barChartC{0.93} \\

     & \hspace{0.2em}LoD4 & 27.23 & 0.804 & 0.289 & \barChartA{0.60}
            & 23.76 & 0.831 & 0.241 & \barChartB{0.44}
            & 29.70 & 0.901 & 0.326 & \barChartC{0.49} \\

     & \hspace{0.2em}LoD3 & 26.99 & 0.790 & 0.312 & \barChartA{0.38}
            & 23.49 & 0.821 & 0.259 & \barChartB{0.30}
            & 29.66 & 0.901 & 0.331 & \barChartC{0.36} \\
\midrule
\multirow{5}{*}{\hspace{0.8em}LP\text{-}3DGS~\cite{LP-3DGS}}
  & $\rho\text{=0.2}$
  & 27.74 & 0.814 & \cellcolor{third}0.217 & \barChartA{2.54}
  & 24.21 & 0.854 & 0.167 & \barChartB{1.47}
  & 29.34 & 0.898 & 0.251 & \barChartC{2.26} \\
  & $\rho\text{=0.4}$
  & 27.74 & 0.814 & 0.218 & \barChartA{1.90}
  & 24.23 & 0.855 & 0.167 & \barChartB{1.10}
  & 29.32 & 0.899 & 0.250 & \barChartC{1.69} \\
  & $\rho\text{=0.6}$
  & 27.70 & 0.815 & 0.222 & \barChartA{1.26}
  & 24.18 & 0.853 & 0.172 & \barChartB{0.74}
  & 29.33 & 0.898 & 0.251 & \barChartC{1.12} \\
  & $\rho\text{=0.8}$
  & 27.52 & 0.809 & 0.242 & \barChartA{0.63}
  & 23.90 & 0.845 & 0.193 & \barChartB{0.36}
  & 29.44 & 0.902 & 0.254 & \barChartC{0.56} \\
  & $\rho\text{=0.9}$
  & 26.90 & 0.789 & 0.279 & \barChartA{0.32}
  & 23.47 & 0.825 & 0.232 & \barChartB{0.18}
  & 29.27 & 0.901 & 0.263 & \barChartC{0.28} \\
\midrule
\multirow{5}{*}{\hspace{0.6em}\makecell[c]{\textbf{ControlGS}\\\hspace{0.45em}\textbf{(Ours)}}} 
  & $\lambda_\alpha\text{=1e-7}$
  & \cellcolor{first}\textbf{28.15} & \cellcolor{first}\textbf{0.831} & \cellcolor{first}\textbf{0.195} & \barChartA{1.76}
  & \cellcolor{first}\textbf{24.64} & \cellcolor{first}\textbf{0.869} & \cellcolor{first}\textbf{0.140} & \barChartB{1.68}
  & \cellcolor{first}\textbf{30.08} & \cellcolor{first}\textbf{0.911} & \cellcolor{first}\textbf{0.240} & \barChartC{0.90} \\

  & $\lambda_\alpha\text{=2e-7}$
  & \cellcolor{second}28.08 & \cellcolor{second}0.827 & \cellcolor{second}0.209 & \barChartA{1.10}
  & \cellcolor{second}24.41 & \cellcolor{second}0.863 & \cellcolor{second}0.152 & \barChartB{1.10}
  & \cellcolor{second}29.96 & \cellcolor{second}0.910 & 0.248 & \barChartC{0.61} \\

  & $\lambda_\alpha\text{=3e-7}$
  & \cellcolor{third}27.90 & \cellcolor{third}0.821 & 0.221 & \barChartA{0.83}
  & \cellcolor{third}24.35 & \cellcolor{third}0.857 & \cellcolor{third}0.162 & \barChartB{0.85}
  & 29.81 & 0.907 & 0.257 & \barChartC{0.47} \\

  & $\lambda_\alpha\text{=5e-7}$
  & 27.70 & 0.810 & 0.242 & \barChartA{0.56}
  & 24.15 & 0.849 & 0.176 & \barChartB{0.62}
  & 29.64 & 0.900 & 0.273 & \barChartC{0.33} \\

  & $\lambda_\alpha\text{=1e-6}$
  & 27.10 & 0.780 & 0.284 & \barChartA{0.31}
  & 23.61 & 0.828 & 0.207 & \barChartB{0.34}
  & 29.14 & 0.889 & 0.295 & \barChartC{0.19} \\
\bottomrule
\end{tabular}
\vspace{-12pt}
\end{table*}

\newcommand{\barChartD}[1]{%
  \smash{\pgfmathsetmacro{\barlen}{#1/0.59*0.915}%
  \begin{tikzpicture}[baseline=(base)]
    \node[inner sep=0pt] (base) at (0.9cm, 0.053cm) {};
    \fill[green] (0,0) rectangle (\barlen cm, 0.27cm);
    \node[anchor=center, font=\scriptsize] at (0.45cm, 0.14cm) {#1};
  \end{tikzpicture}%
}}

\newcommand{\barChartE}[1]{%
  \smash{\pgfmathsetmacro{\barlen}{#1/3.38*0.915}%
  \begin{tikzpicture}[baseline=(base)]
    \node[inner sep=0pt] (base) at (0.9cm, 0.053cm) {};
    \fill[green] (0,0) rectangle (\barlen cm, 0.27cm);
    \node[anchor=center, font=\scriptsize] at (0.45cm, 0.14cm) {#1};
  \end{tikzpicture}%
}}

\begin{table}[!t]
\caption{Comparison with non-budget-based structural compression methods on the NeRF synthetic (NeRF) and GigaNVS datasets, following the format of Table~\ref{tab:scene_comparison}.}
\label{tab:synthetic_comparison}
\centering
\scriptsize
\setlength{\tabcolsep}{0pt}
\renewcommand{\arraystretch}{0.75}
\setlength{\extrarowheight}{3pt}
\begin{tabular}{C{1.45cm}C{1.3cm}|C{0.9cm}C{1.0cm}|C{0.9cm}C{0.9cm}C{0.9cm}C{1.0cm}}
\toprule
\multicolumn{2}{c|}{\textbf{Dataset}}
  & \multicolumn{2}{c|}{\textbf{NeRF (Object)}}
  & \multicolumn{4}{c}{\textbf{GigaNVS (Large Outdoor)}} \\
\multicolumn{2}{c|}{Method $\mid$ Metrics}
  & PSNR$\uparrow$ & Num(M)
  & PSNR$\uparrow$ & SSIM$\uparrow$ & LPIPS$\downarrow$ & Num(M) \\
\midrule
\multicolumn{2}{c|}{3DGS~\cite{3DGS}}
  & 33.55 & \barChartD{0.26} 
  & 19.30 & 0.741 & 0.284 & \barChartE{3.18} \\
\multicolumn{2}{c|}{EAGLES*~\cite{EAGLES}}
  & 32.27 & \barChartD{0.09}
  & 19.02 & 0.705 & 0.331 & \barChartE{1.39} \\
\midrule
\multirow{5}{*}{\hspace{0.5em}LP-3DGS~\cite{LP-3DGS}}
  & $\rho\text{=0.2}$
    & 33.49 & \barChartD{0.23}
    & 19.44 & \cellcolor{second}0.748 & \cellcolor{second}0.273 & \barChartE{2.52} \\
  & $\rho\text{=0.4}$
    & 33.51 & \barChartD{0.17}
    & 19.40 & \cellcolor{third}0.744 & \cellcolor{third}0.280 & \barChartE{1.90} \\
  & $\rho\text{=0.6}$
    & 33.50 & \barChartD{0.12}
    & 19.32 & 0.735 & 0.295 & \barChartE{1.26} \\
  & $\rho\text{=0.8}$
    & 33.29 & \barChartD{0.06}
    & 19.02 & 0.701 & 0.340 & \barChartE{0.63} \\
  & $\rho\text{=0.9}$
    & 32.50 & \barChartD{0.03}
    & 18.68 & 0.644 & 0.404 & \barChartE{0.31} \\
\midrule
\multirow{5}{*}{\hspace{0.2em}\makecell[c]{\textbf{ControlGS}\\\hspace{0.2em}\textbf{(Ours)}}} 
  & $\lambda_\alpha\text{=1e-7}$
    & \cellcolor{first}\textbf{33.85} & \barChartD{0.59}
    & \cellcolor{first}\textbf{19.91} & \cellcolor{first}\textbf{0.763} & \cellcolor{first}\textbf{0.260} & \barChartE{3.38} \\
  & $\lambda_\alpha\text{=2e-7}$
    & \cellcolor{second}33.81 & \barChartD{0.35}
    & \cellcolor{second}19.72 & 0.743 & 0.284 & \barChartE{2.05} \\
  & $\lambda_\alpha\text{=3e-7}$
    & \cellcolor{third}33.61 & \barChartD{0.26}
    & \cellcolor{third}19.59 & 0.723 & 0.307 & \barChartE{1.50} \\
  & $\lambda_\alpha\text{=5e-7}$
    & 33.50 & \barChartD{0.18}
    & 19.30 & 0.688 & 0.345 & \barChartE{0.95} \\
  & $\lambda_\alpha\text{=1e-6}$
    & 33.10 & \barChartD{0.11}
    & 18.63 & 0.601 & 0.434 & \barChartE{0.32} \\
\bottomrule
\end{tabular}
\end{table}

\subsubsection{Quantitative Analysis}  

We compare ControlGS with more non-budget-based structural compression methods. 
As shown in Tables~\ref{tab:scene_comparison} and~\ref{tab:synthetic_comparison}, 
although existing non-budget-based approaches can reduce the number of Gaussians compared to the vanilla 3DGS~\cite{3DGS}, 
they generally suffer from a decline in rendering quality, as reflected by PSNR, SSIM, and LPIPS.

In contrast, ControlGS overcomes this limitation by simultaneously reducing the Gaussian count while improving rendering quality. 
For example, compared to 3DGS, ControlGS achieves 28.08 dB with 1.10M Gaussians on Mip-NeRF360 (a 58.1\% reduction and a 0.45 dB gain), 
24.41 dB with 1.10M on Tanks and Temples (a 30.4\% reduction and a 0.71 dB gain), 
and 30.08 dB with 0.90M on Deep Blending (a 63.7\% reduction and a 0.20 dB gain).

Overall, ControlGS provides a more efficient scene representation and clearly outperforms existing methods 
in the trade-off between Gaussian count and rendering quality, 
forming a better Pareto frontier where similar quality is achieved with fewer Gaussians, 
or higher quality is achieved with a comparable model size. 
This advantage is consistent across multiple datasets and scene scales.

\begin{figure}[h]
  \centering
  \vspace{-7pt}
  \includegraphics[width=0.99\linewidth]{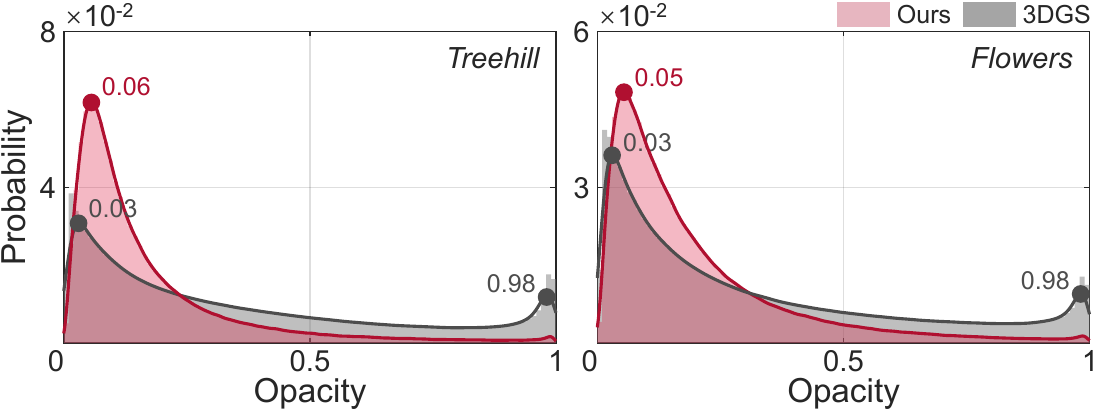}
  \vspace{-16pt}
  \caption{Opacity distributions of our method (\textcolor[HTML]{DC143C}{red}) and vanilla 3DGS (\textcolor[HTML]{808080}{gray}) on \textit{Treehill} and \textit{Flowers} scenes from Mip-NeRF360~\cite{Mip-NeRF360}, with highlighted peaks and opacities.}
  \label{fig:opacity_comp}
  \vspace{-8pt}
\end{figure}

\begin{figure}[h]
  \centering
  \vspace{-7pt}
  \includegraphics[width=0.99\linewidth]{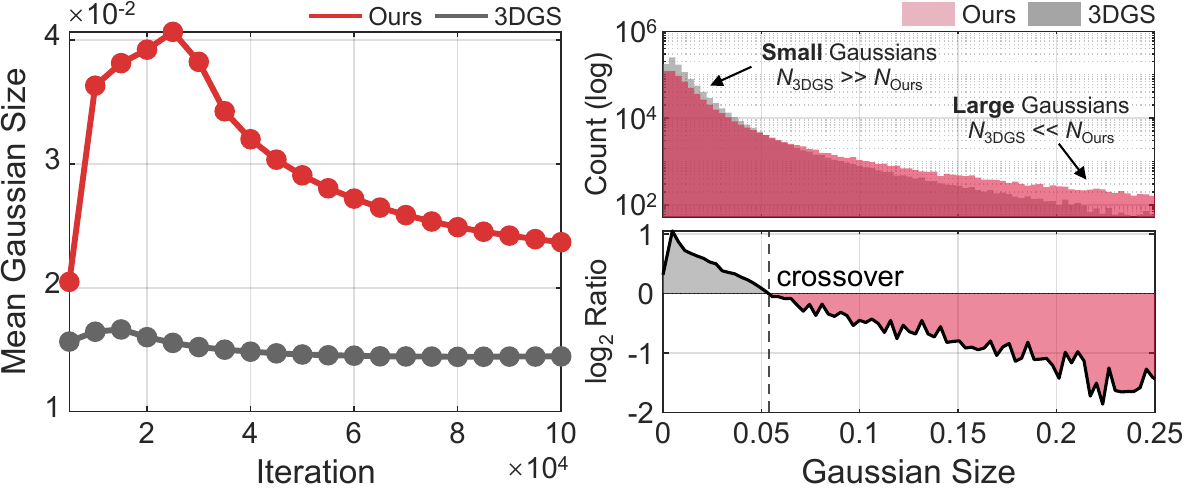}
  \vspace{-16pt}
  \caption{\textit{Left}: mean Gaussian size over training; \textit{Right}: final size distribution (log-scale histogram and log$_2(N_\text{3DGS}/N_\text{Ours})$ curve) of our method~(\textcolor[HTML]{DC143C}{red}) and vanilla~3DGS~(\textcolor[HTML]{808080}{gray}) on \textit{Counter} scene from Mip-NeRF360~\cite{Mip-NeRF360}.}
  \label{fig:size_evolution_comp}
  \vspace{-8pt}
\end{figure}

\begin{figure}[h]
  \centering
  \vspace{-2pt}
  \setlength{\tabcolsep}{1pt}  
  \setlength{\fboxrule}{0.2pt}
  \setlength{\fboxsep}{-0.2pt}
  \renewcommand{\arraystretch}{0.6} 
  \newcommand{\imgwidth}{0.32\linewidth}
  \begin{tabular}{@{}ccc@{}}
    \fbox{\includegraphics[width=\imgwidth]{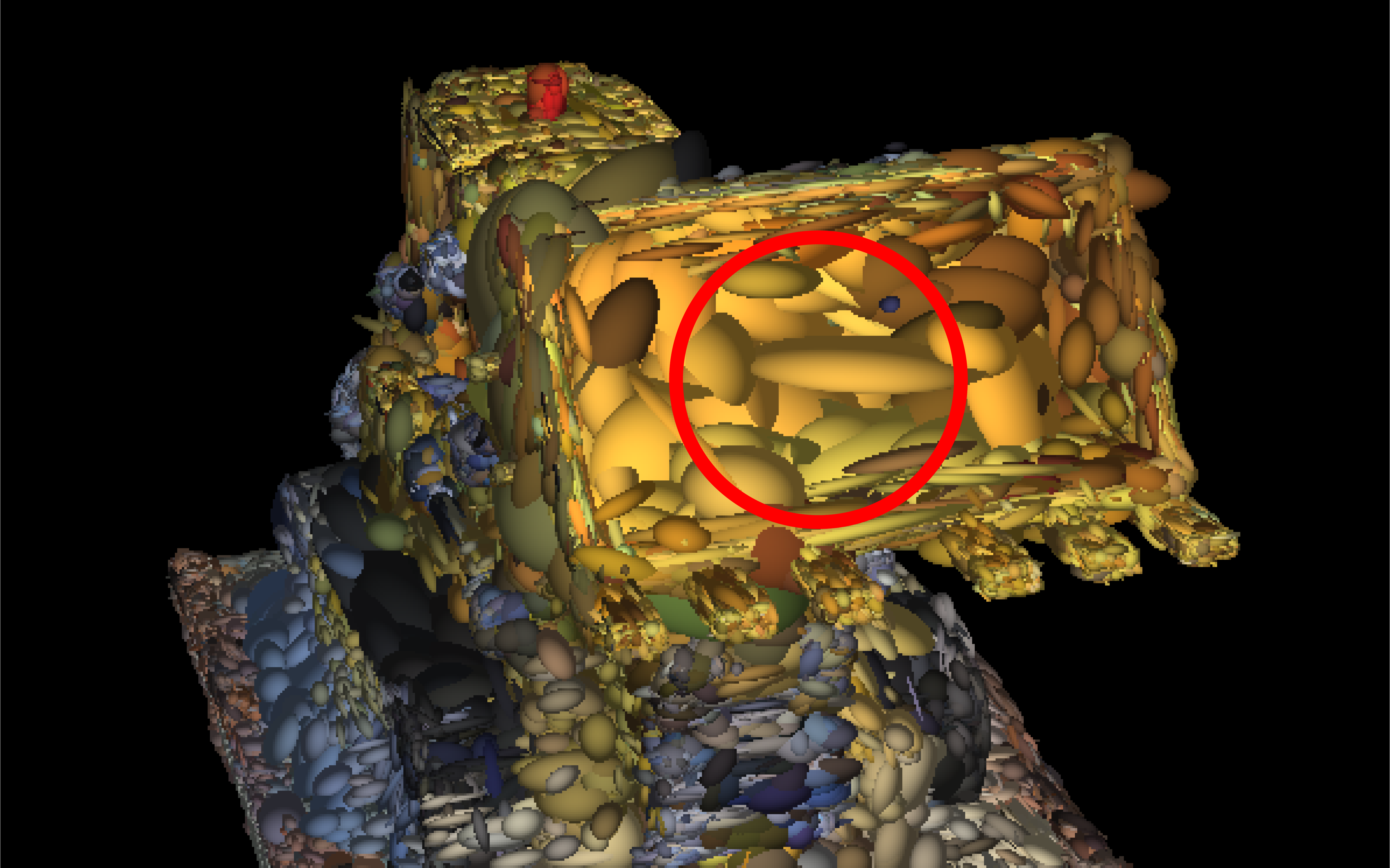}} &
    \fbox{\includegraphics[width=\imgwidth]{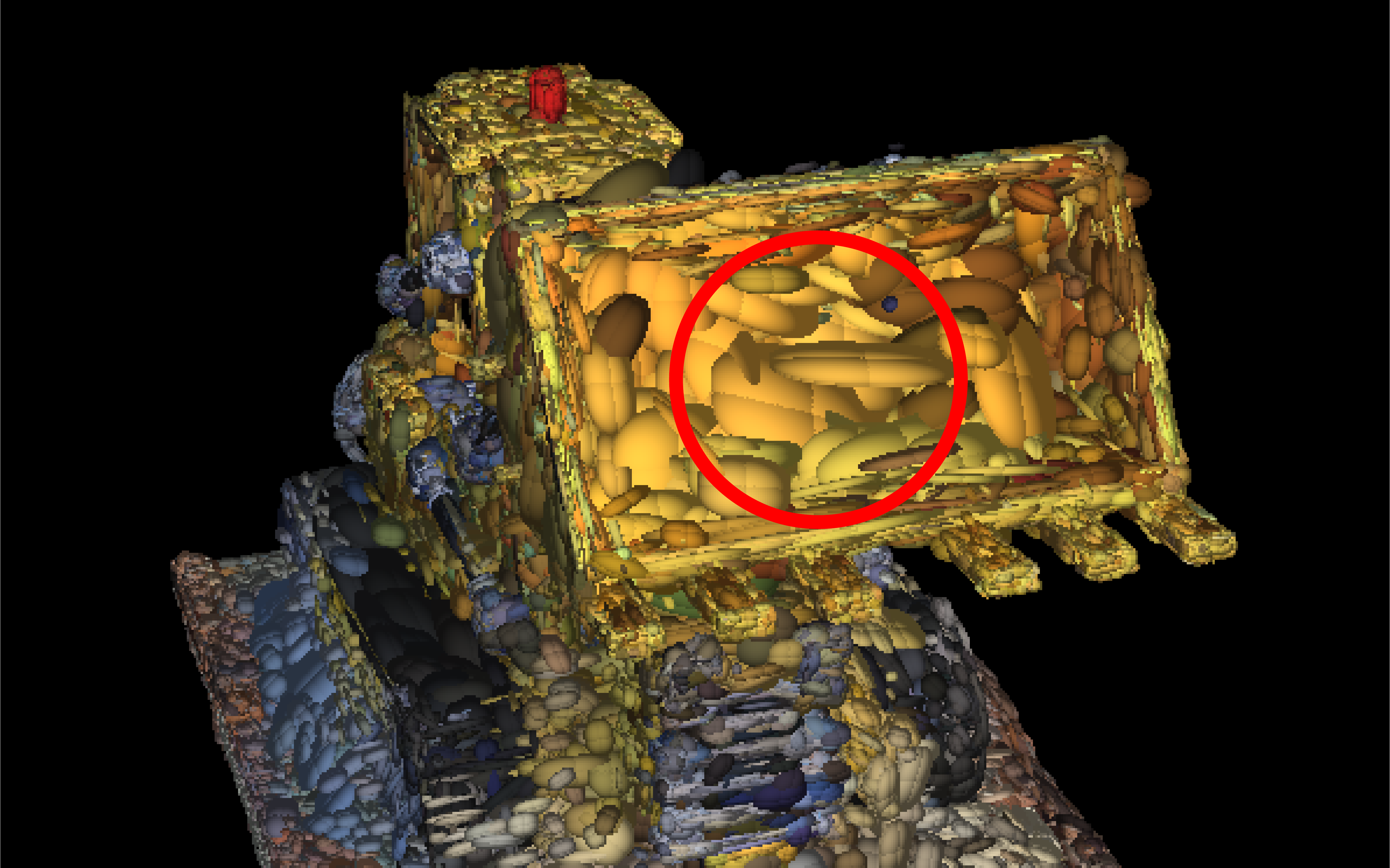}} &
    \fbox{\includegraphics[width=\imgwidth]{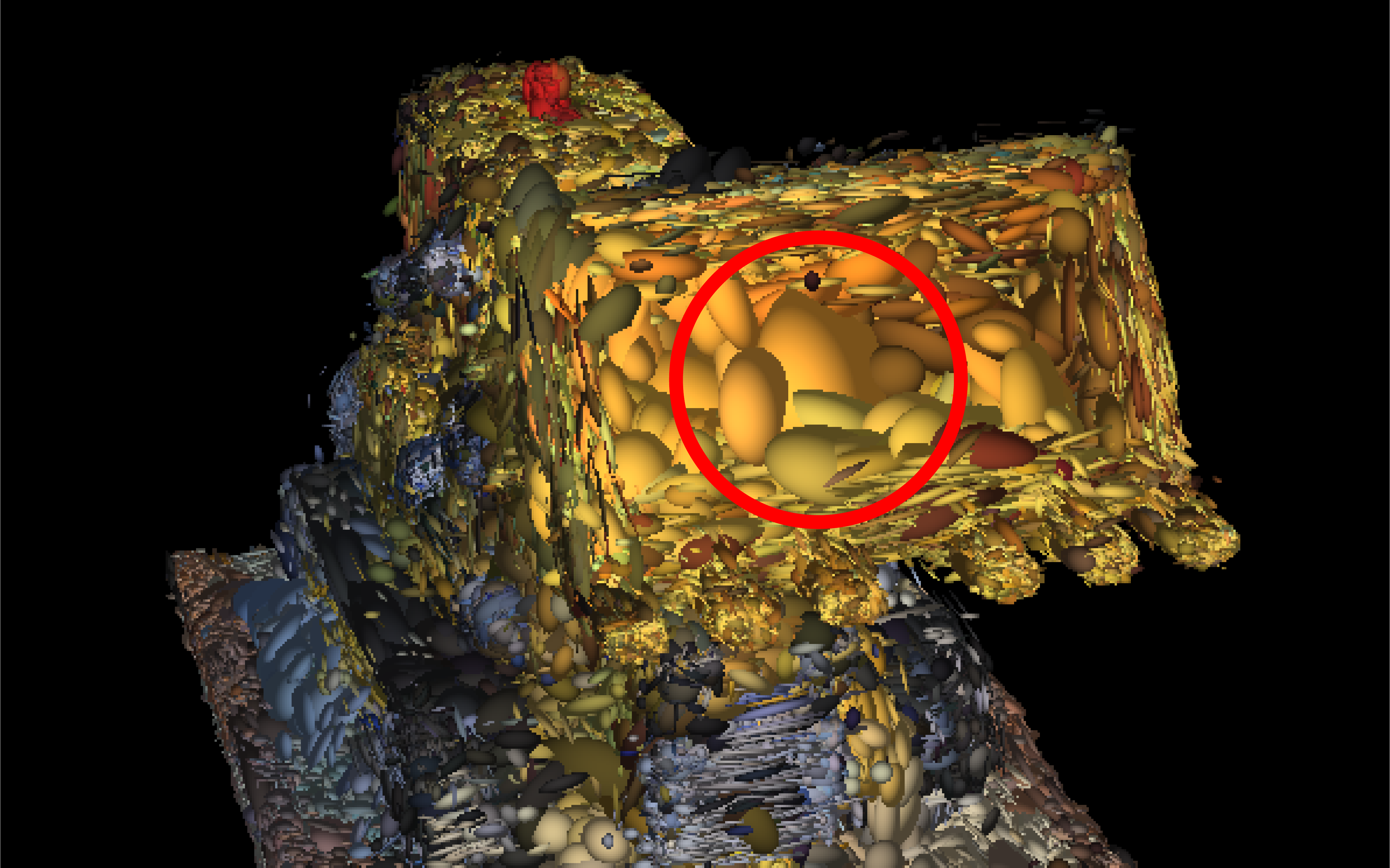}} \\
    [-2pt]
    {\scriptsize Iter.~399 (Before Split)} &
    {\scriptsize Iter.~400 (Split)} &
    {\scriptsize Iter.~600 (Pruned)} \\
    \fbox{\includegraphics[width=\imgwidth]{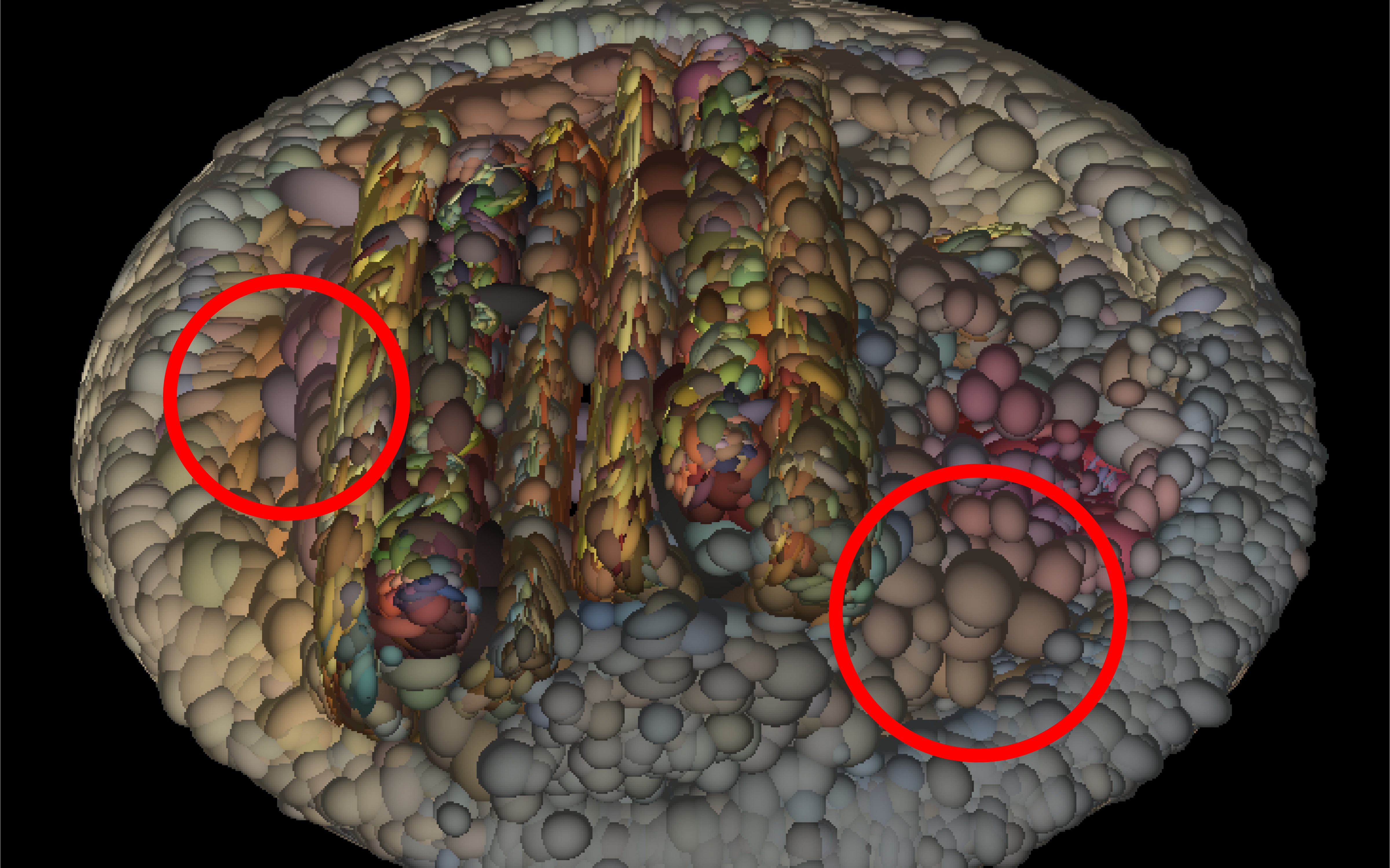}} &
    \fbox{\includegraphics[width=\imgwidth]{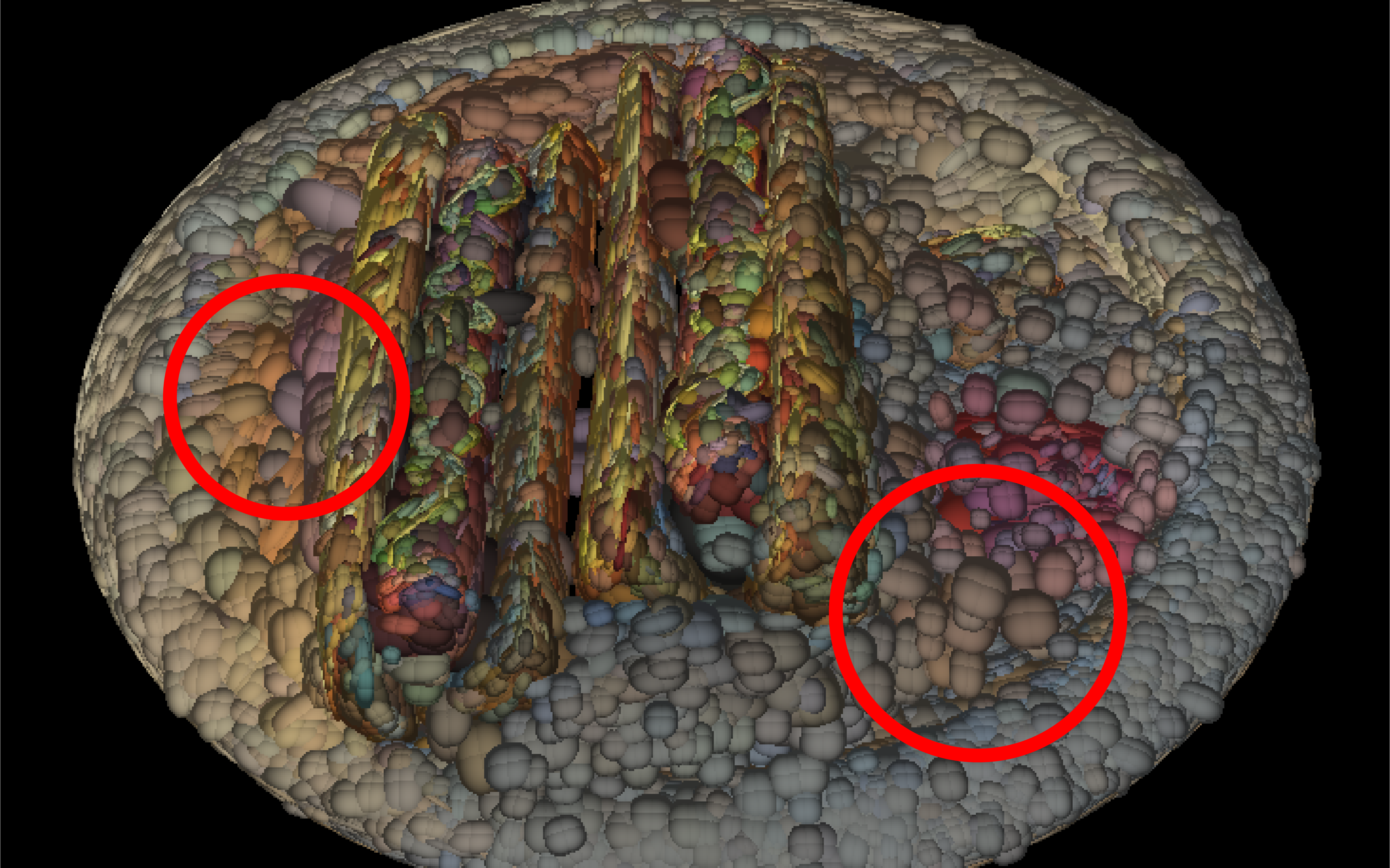}} &
    \fbox{\includegraphics[width=\imgwidth]{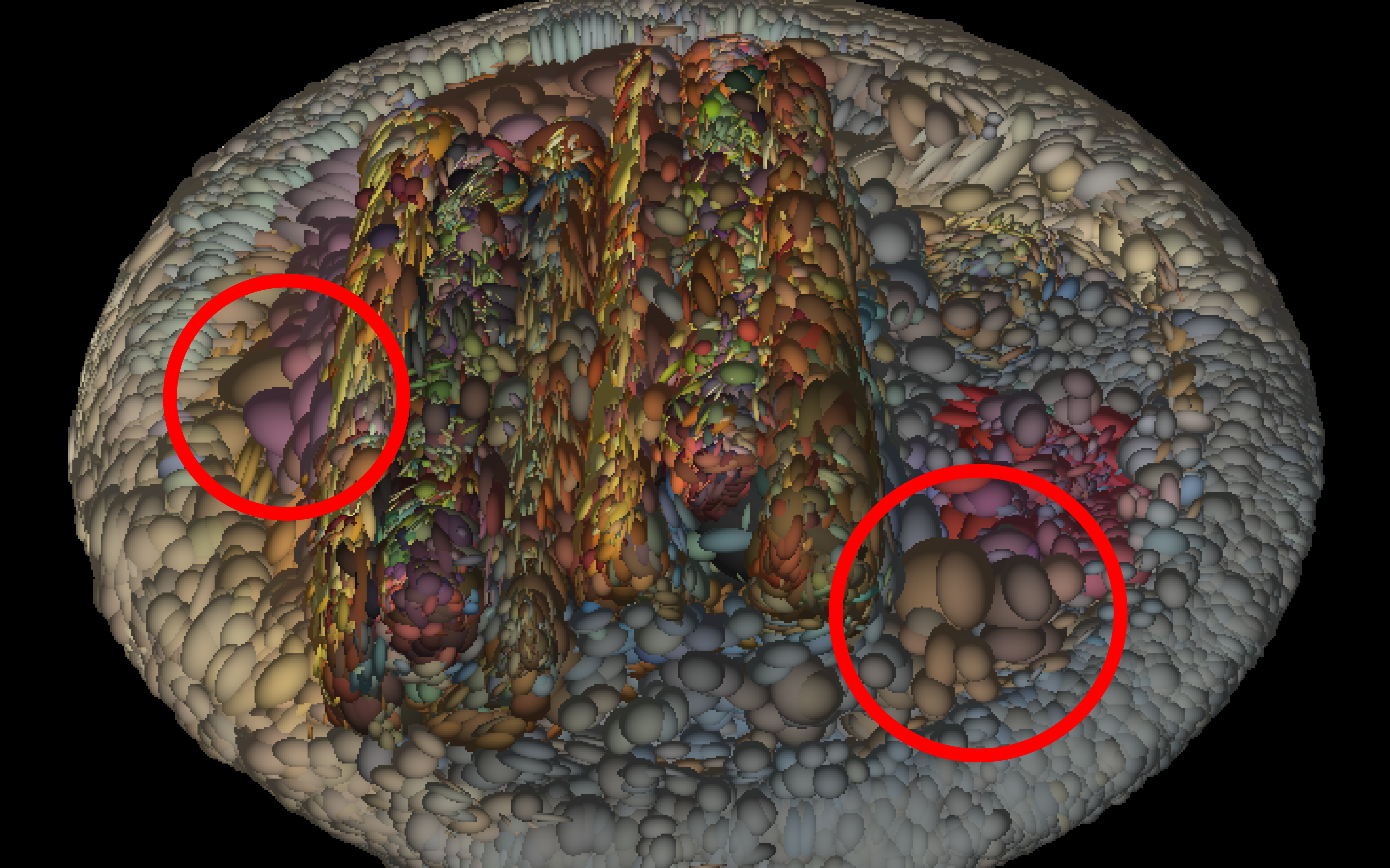}} \\
    [-2pt]
    {\scriptsize Iter.~799 (Before Split)} &
    {\scriptsize Iter.~800 (Split)} &
    {\scriptsize Iter.~1000 (Pruned)} \\
  \end{tabular}
  \vspace{-3pt}
  \caption{Illustration of the self-correcting mechanism. \textit{Uniform splitting} may over-split, whereas \textit{opacity-based sparsification} prunes non-contributing Gaussians, restoring sparsity.
  }
  \label{fig:self_correcting_mechanism}
  \vspace{-8pt}
\end{figure}

To better understand why our method achieves superior rendering quality with fewer Gaussians, we further analyze the training dynamics and structural characteristics of ControlGS.

\newcommand{\imgwidth}{2.45cm} 
\newcommand{\fixedstack}[2]{%
\parbox[c][1.5em][c]{\imgwidth}{%
    \centering
    {\scriptsize #1}\\[-0.8ex]
    {\scriptsize #2}%
  }%
}
\setlength{\fboxrule}{0.2pt}   
\setlength{\fboxsep}{-0.2pt}    
\begin{figure*}[!t]
    \centering

    \renewcommand{\arraystretch}{0.5} 
    \newcolumntype{M}[1]{>{\centering\arraybackslash}m{#1}}
    \hspace*{-0.8cm}
    \begin{tabular}{@{}M{0.8cm}@{}%
                    M{\imgwidth}@{\hspace{0.6mm}}%
                    M{\imgwidth}@{\hspace{0.6mm}}%
                    M{\imgwidth}@{\hspace{0.6mm}}%
                    M{\imgwidth}@{\hspace{0.6mm}}%
                    M{\imgwidth}@{\hspace{0.6mm}}%
                    M{\imgwidth}@{\hspace{0.6mm}}%
                    M{\imgwidth}@{}}
        \adjustbox{angle=0,lap=0em}{\scriptsize\textsf{\textbf{}}} &
        \adjustbox{angle=0,lap=0em}{\scriptsize\textsf{\textbf{Large Outdoor}}} &
        \adjustbox{angle=0,lap=0em}{\scriptsize\textsf{\textbf{Outdoor}}} &
        \adjustbox{angle=0,lap=0em}{\scriptsize\textsf{\textbf{Outdoor}}} &
        \adjustbox{angle=0,lap=0em}{\scriptsize\textsf{\textbf{Indoor}}} &
        \adjustbox{angle=0,lap=0em}{\scriptsize\textsf{\textbf{Indoor}}} &
        \adjustbox{angle=0,lap=0em}{\scriptsize\textsf{\textbf{Indoor}}} &
        \adjustbox{angle=0,lap=0em}{\scriptsize\textsf{\textbf{Object}}} \\
    
        \adjustbox{angle=90,lap=0.8em}{\scriptsize\textsf{\textbf{GT}}} &
        \fbox{\includegraphics[width=\imgwidth]{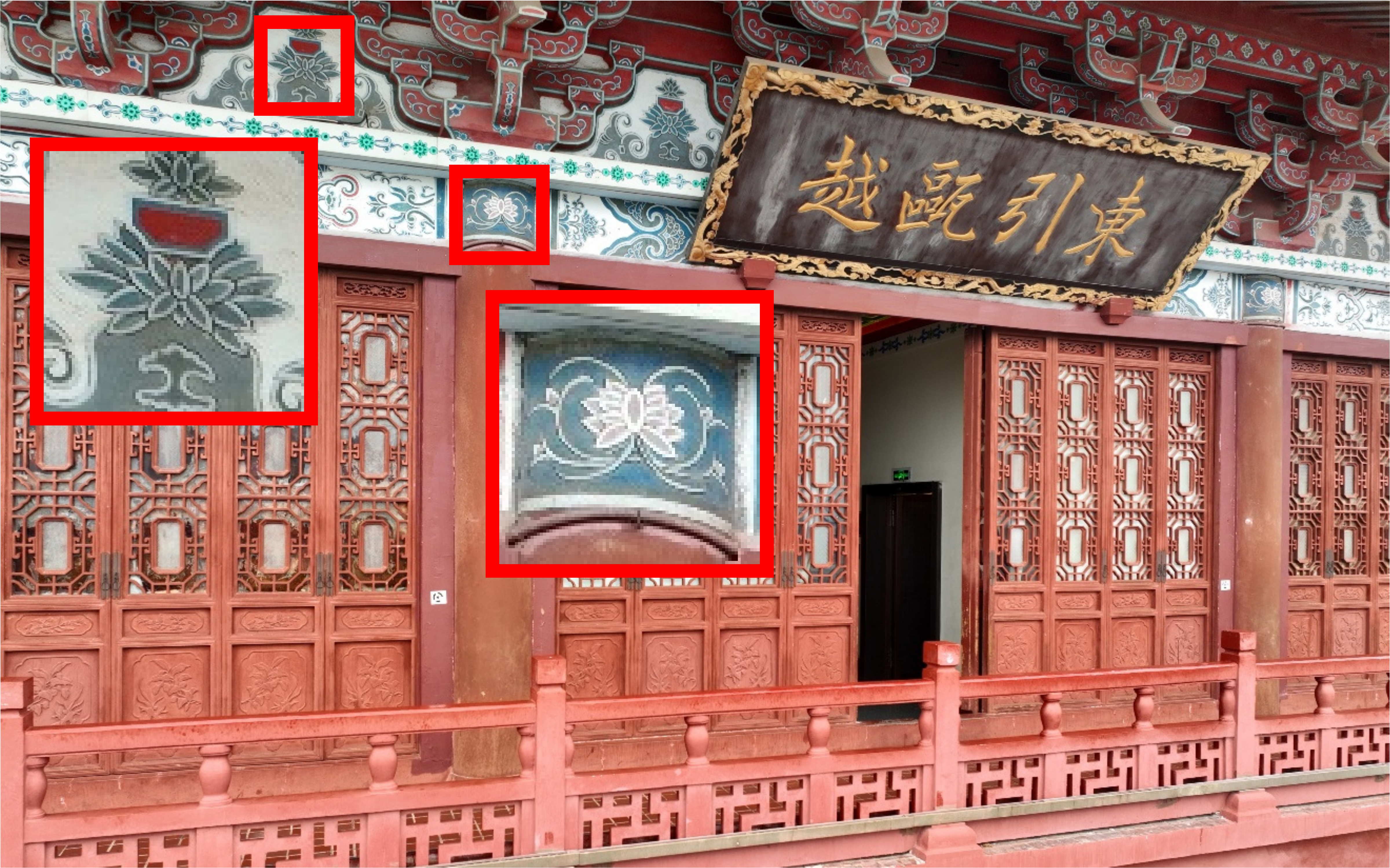}} &
        \fbox{\includegraphics[width=\imgwidth]{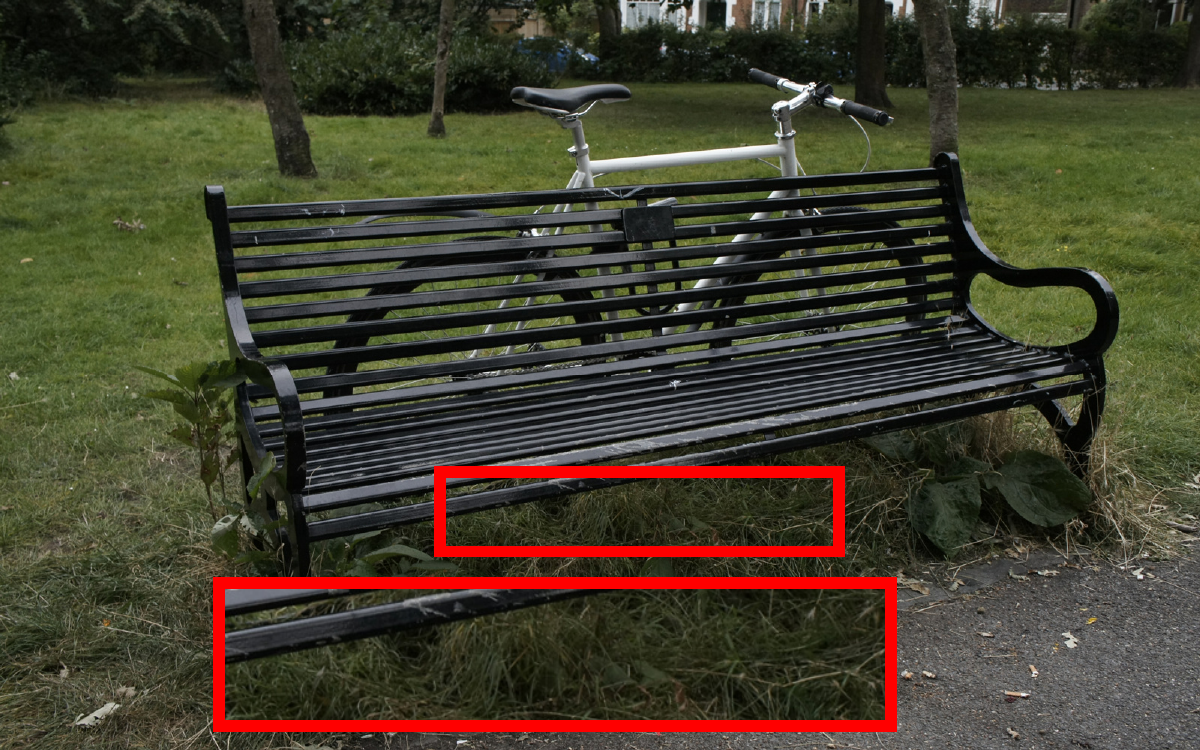}} &
        \fbox{\includegraphics[width=\imgwidth]{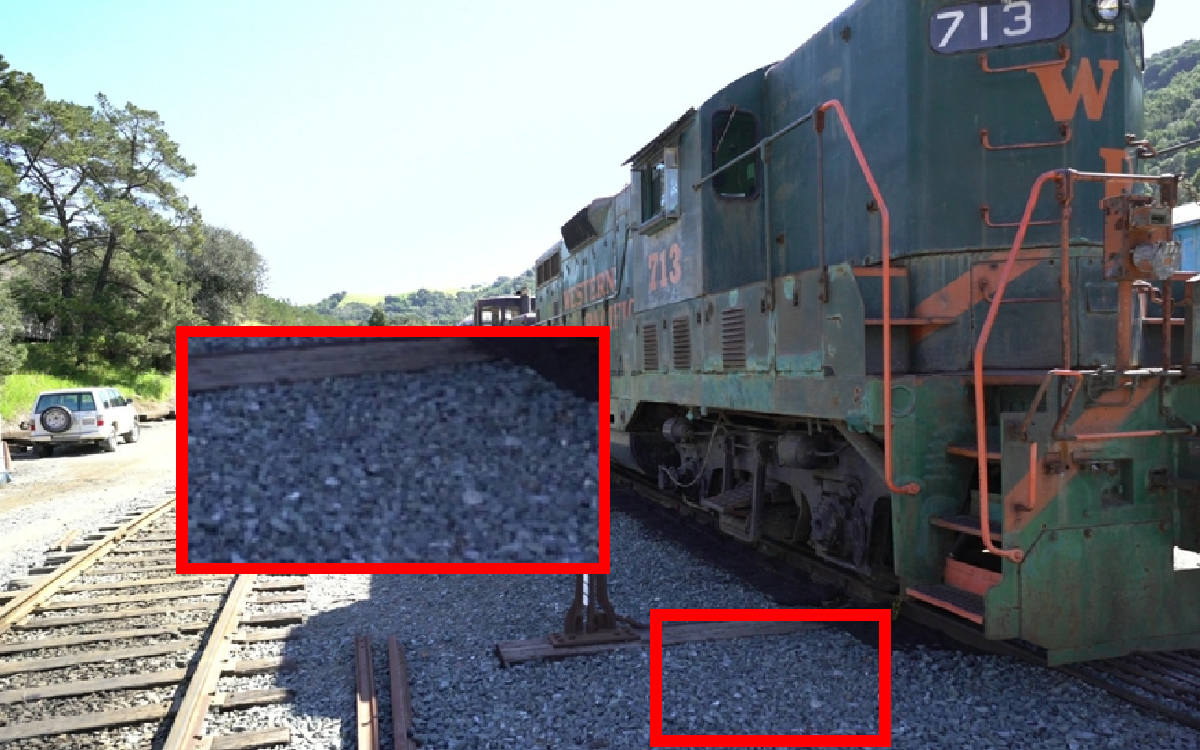}} &
        \fbox{\includegraphics[width=\imgwidth]{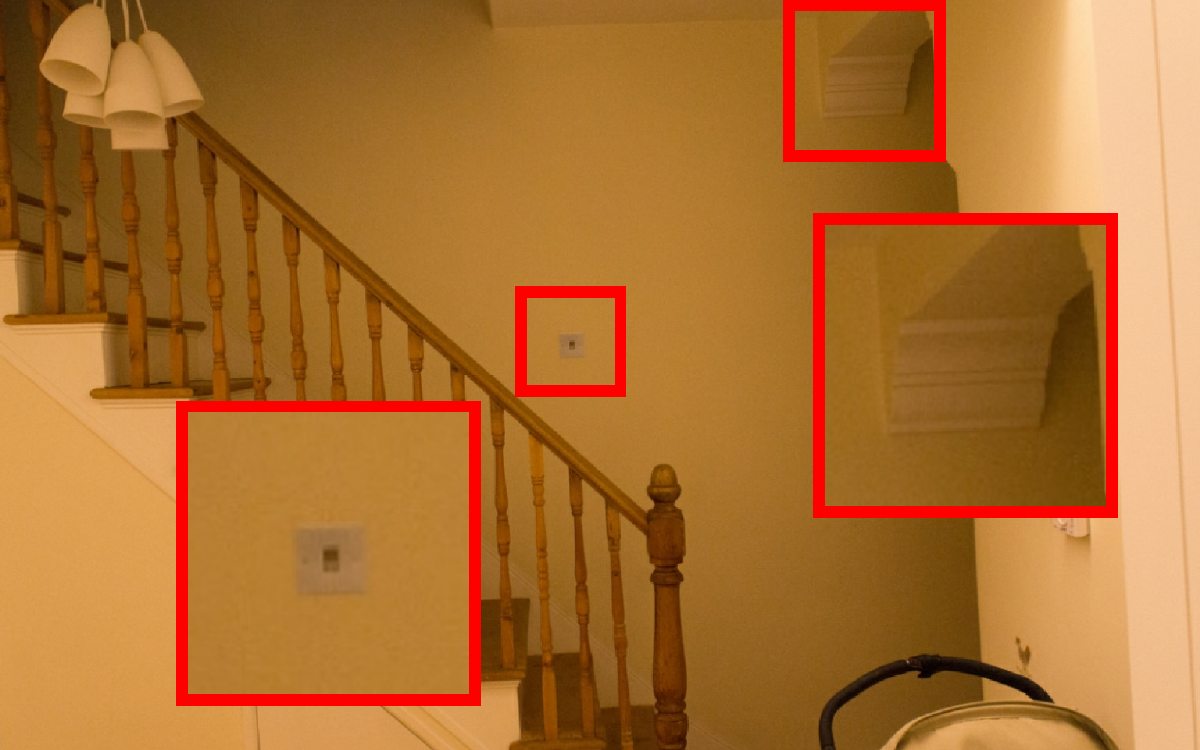}} &
        \fbox{\includegraphics[width=\imgwidth]{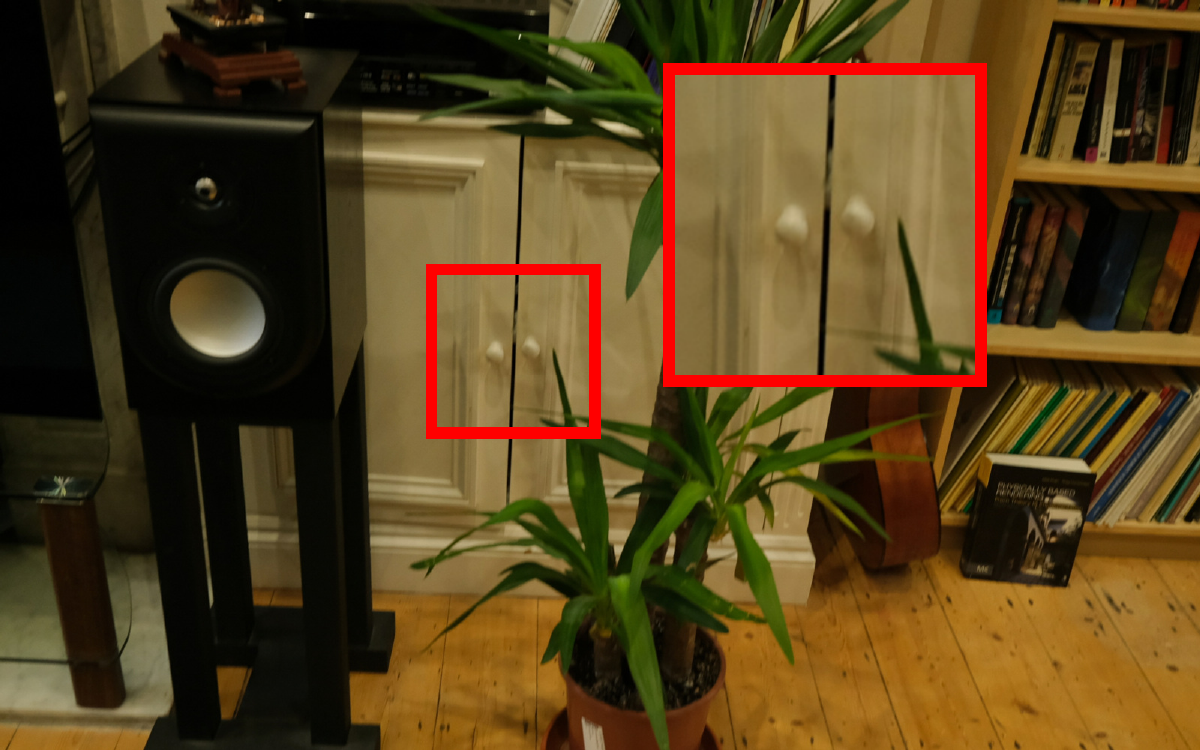}} &
        \fbox{\includegraphics[width=\imgwidth]{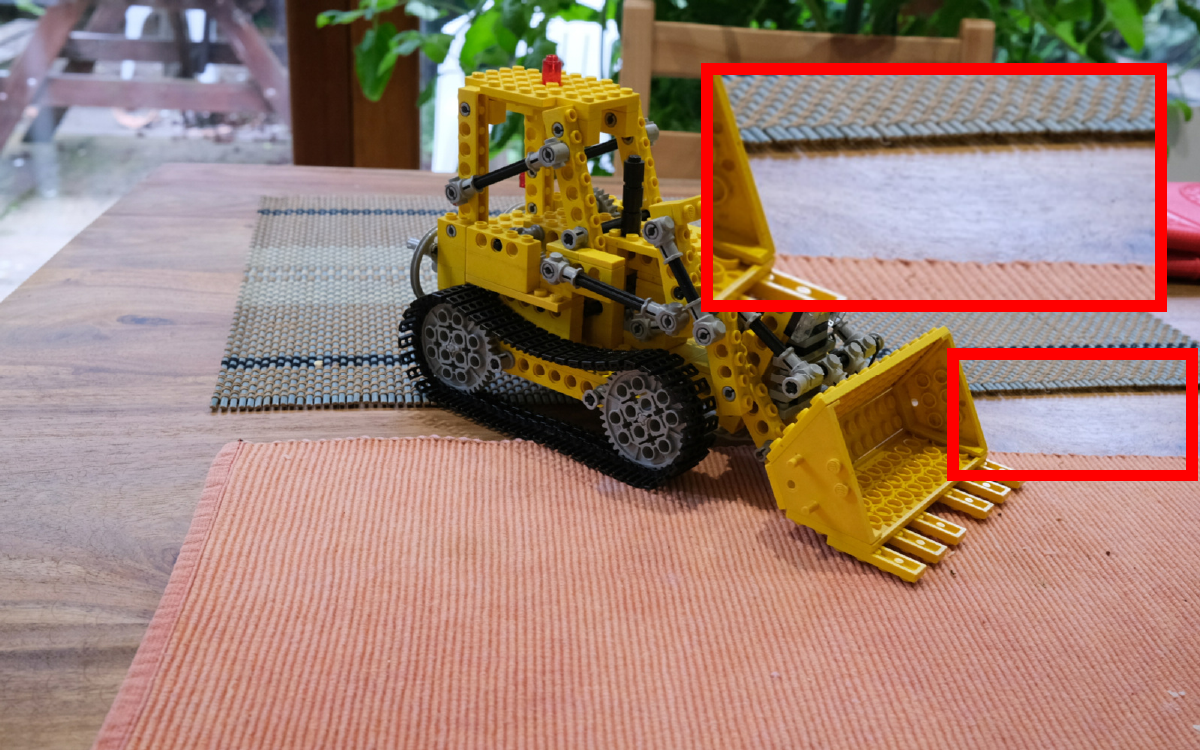}} &
        \fbox{\includegraphics[width=\imgwidth]{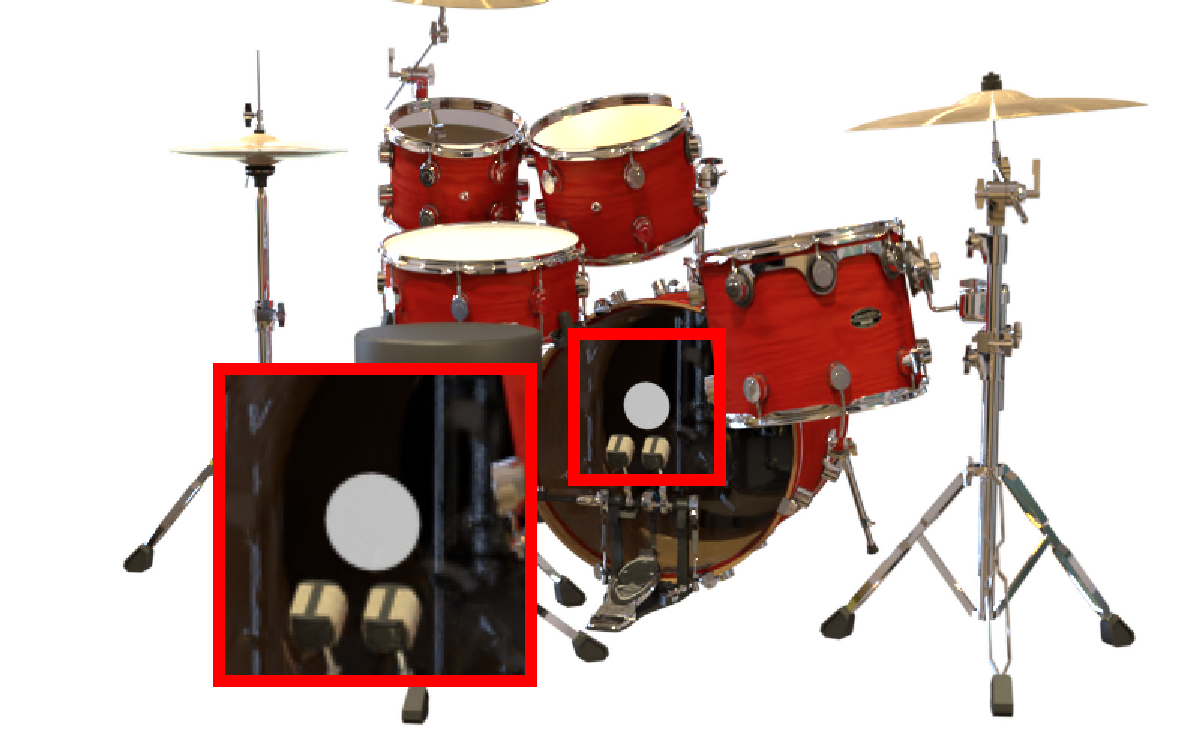}} \\

        \adjustbox{angle=90,lap=0.8em}{\scriptsize\textsf{\textbf{Ours}}} &
        \fbox{\includegraphics[width=\imgwidth]{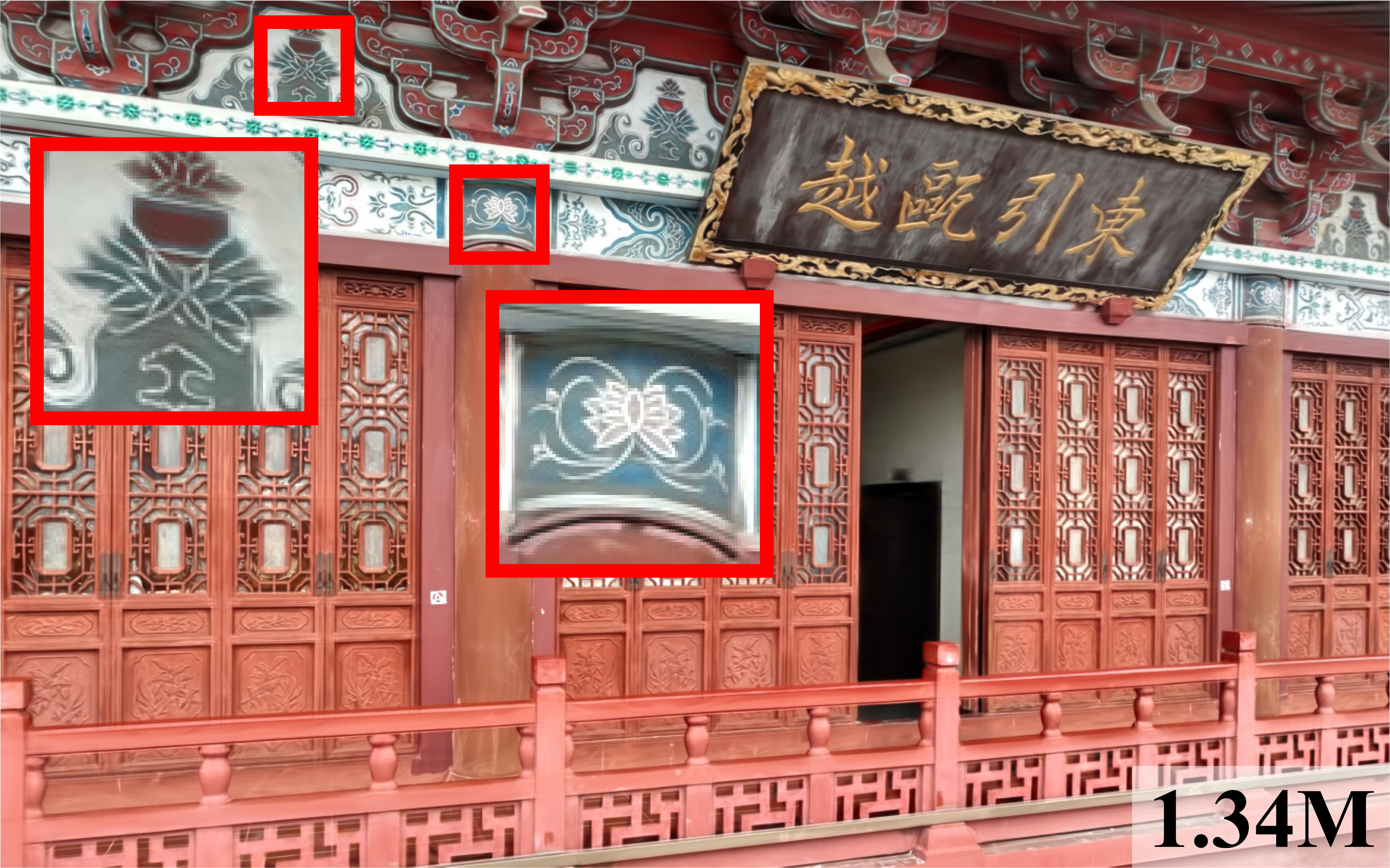}} &
        \fbox{\includegraphics[width=\imgwidth]{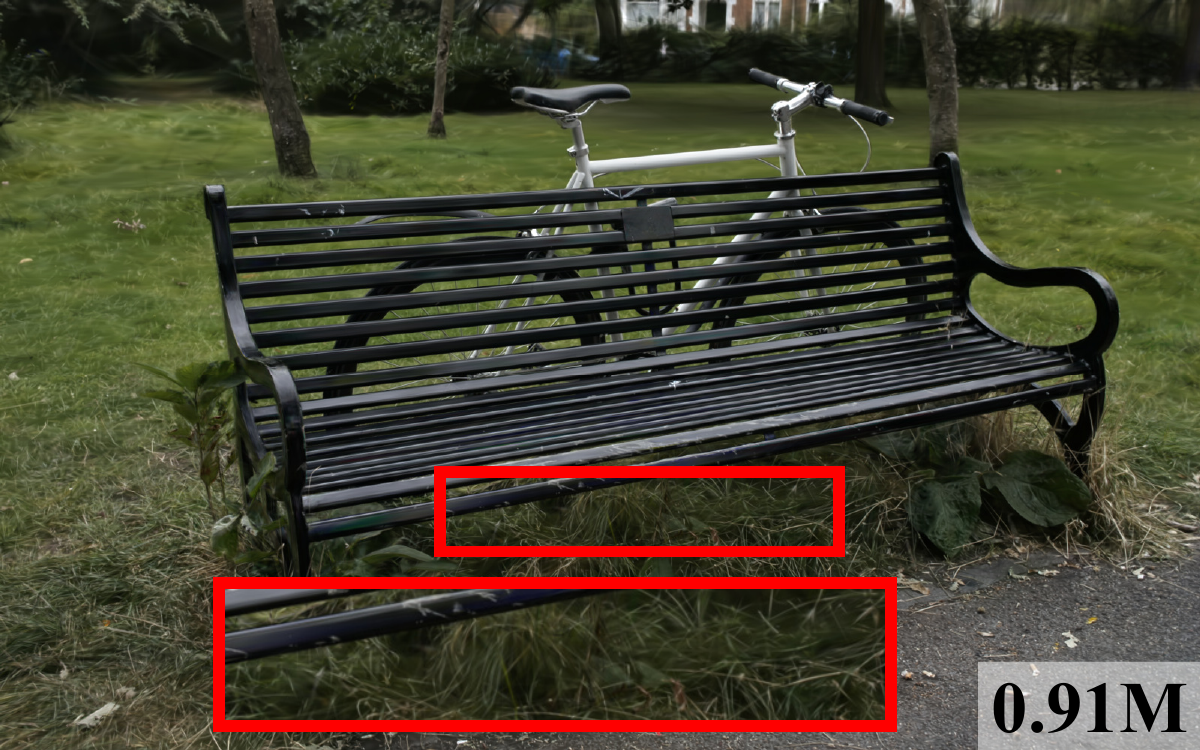}} &
        \fbox{\includegraphics[width=\imgwidth]{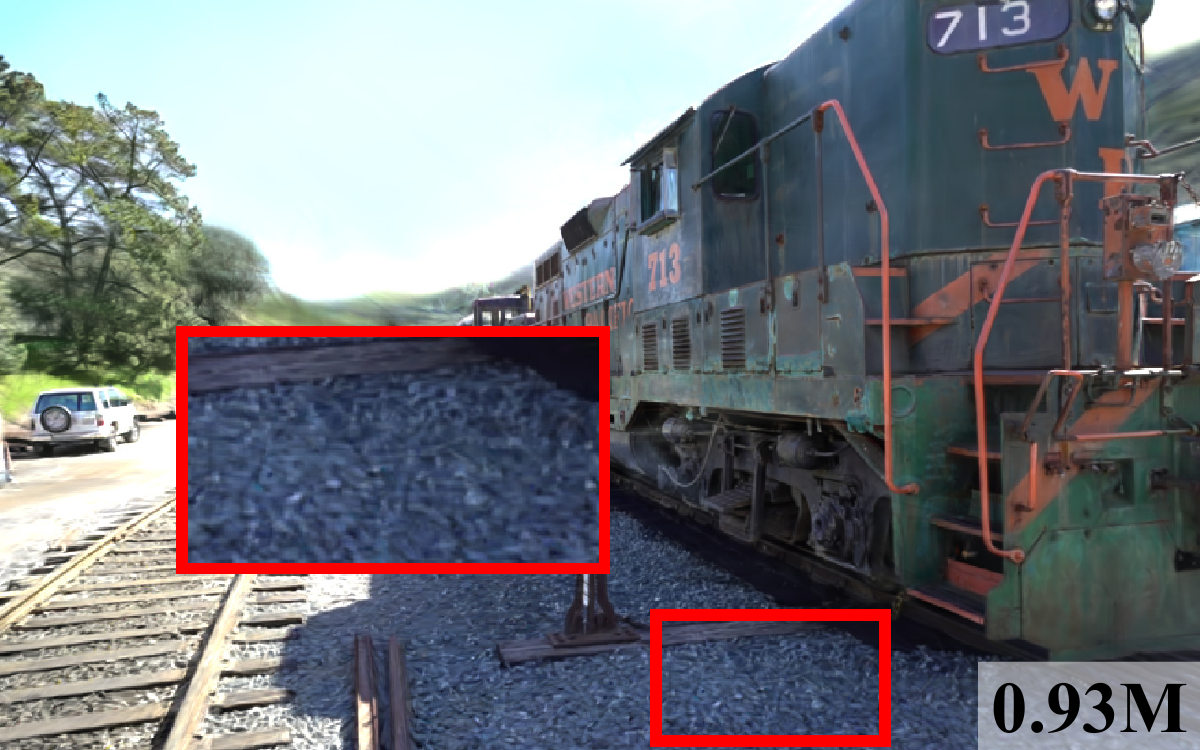}} &
        \fbox{\includegraphics[width=\imgwidth]{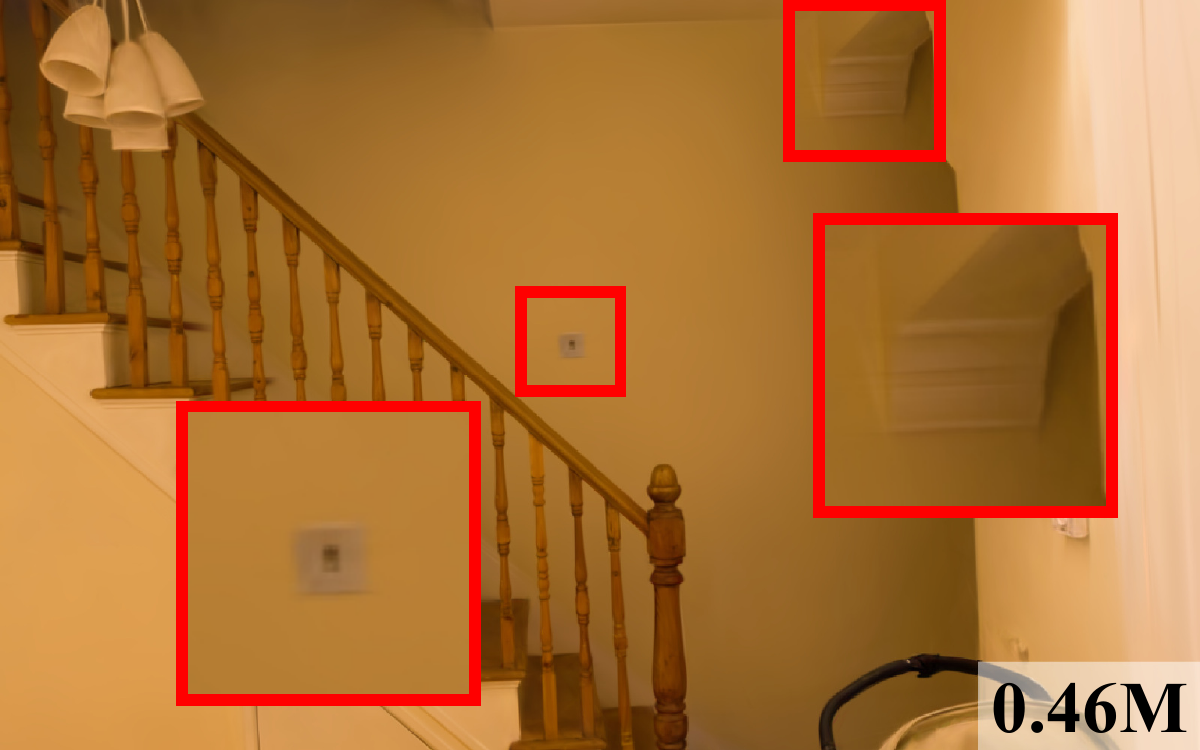}} &
        \fbox{\includegraphics[width=\imgwidth]{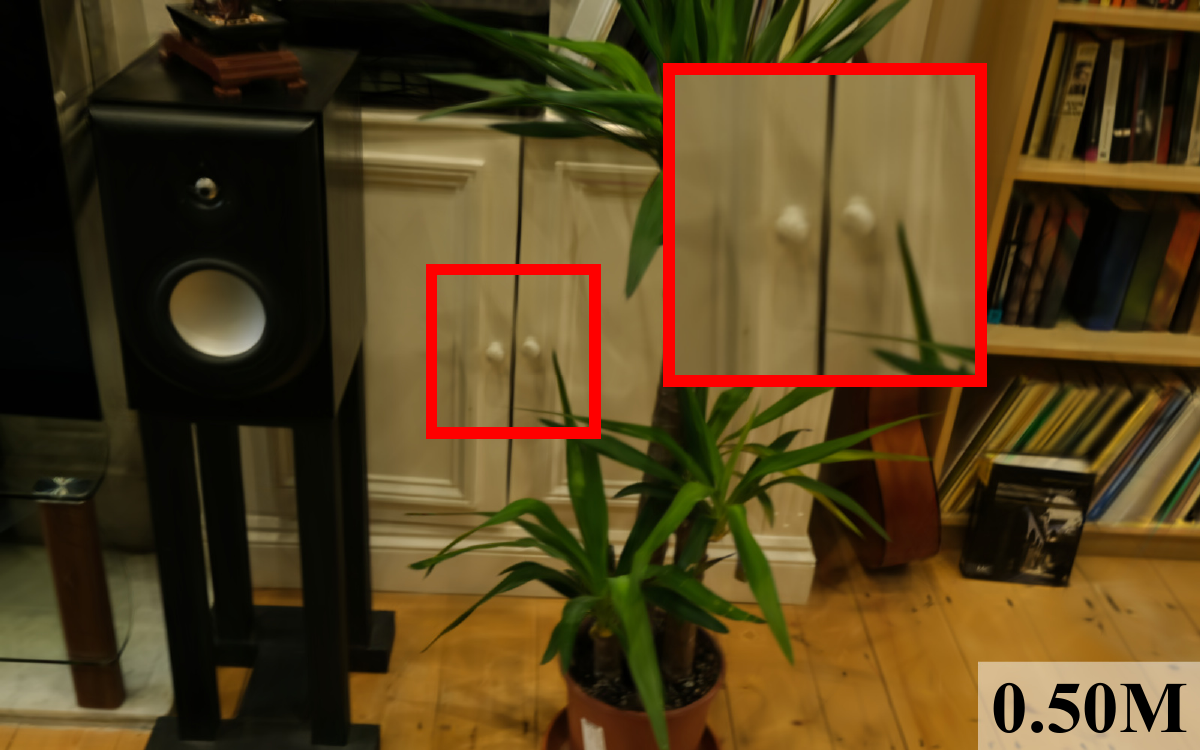}} &
        \fbox{\includegraphics[width=\imgwidth]{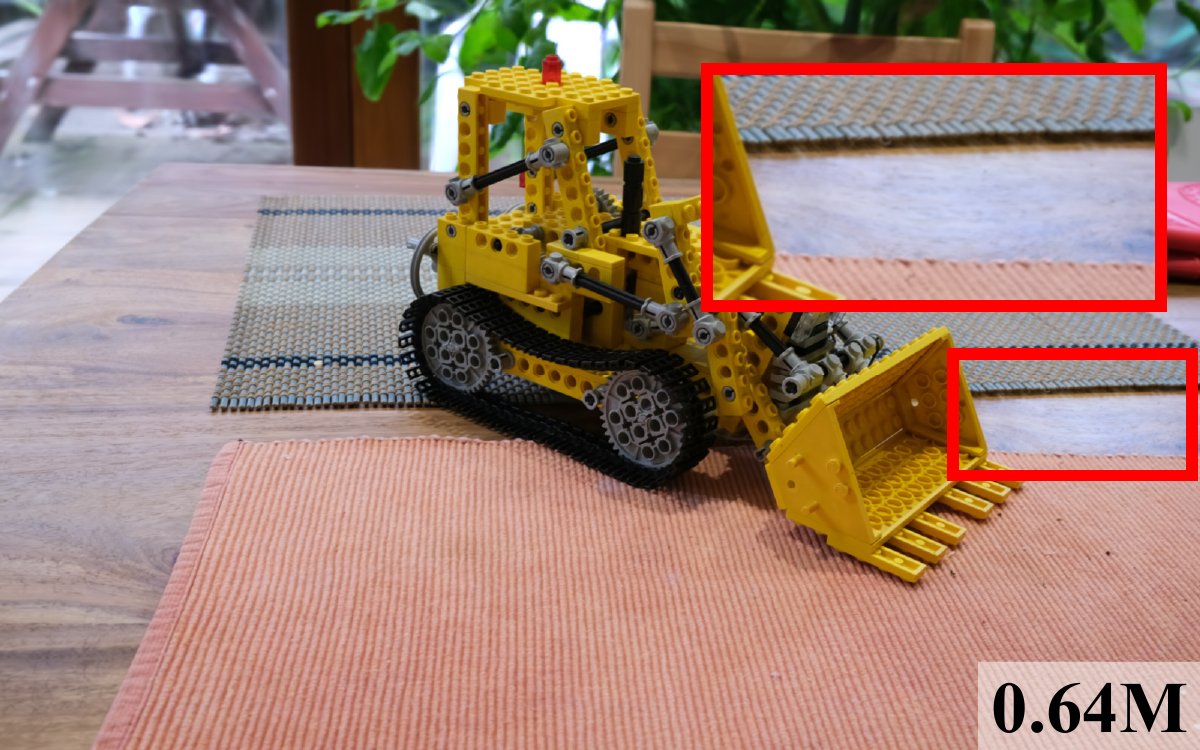}} &
        \fbox{\includegraphics[width=\imgwidth]{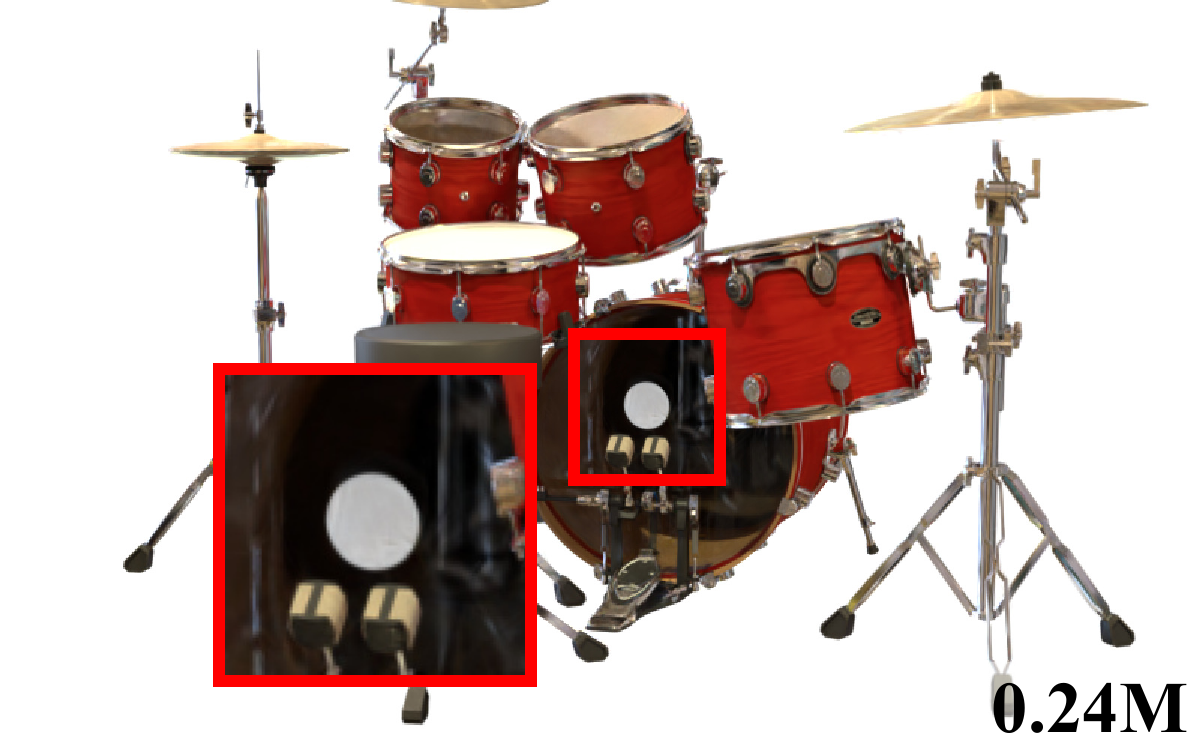}} \\

        \adjustbox{angle=90,lap=0.8em}{\scriptsize\textsf{\textbf{3DGS}}} &
        \fbox{\includegraphics[width=\imgwidth]{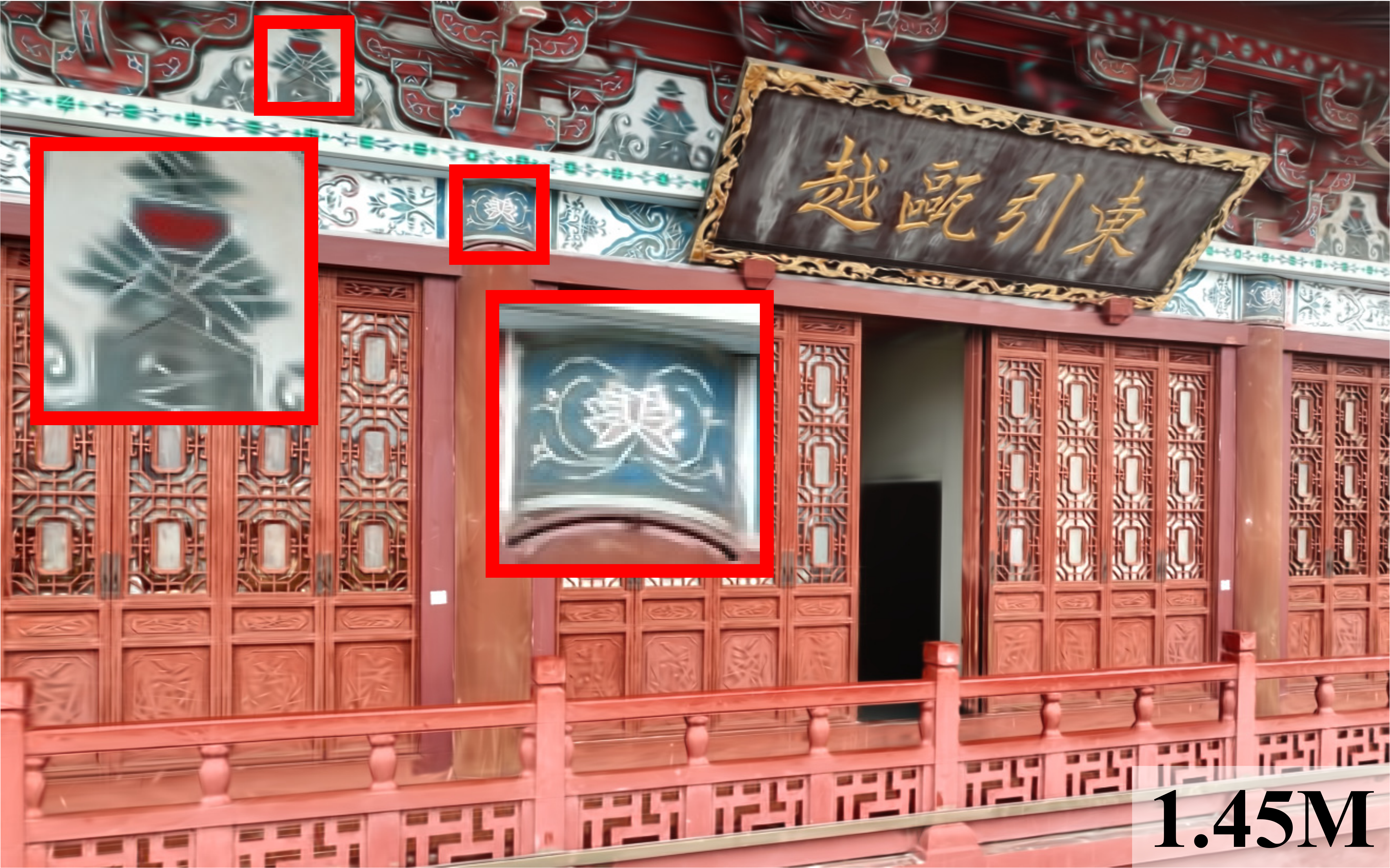}} &
        \fbox{\includegraphics[width=\imgwidth]{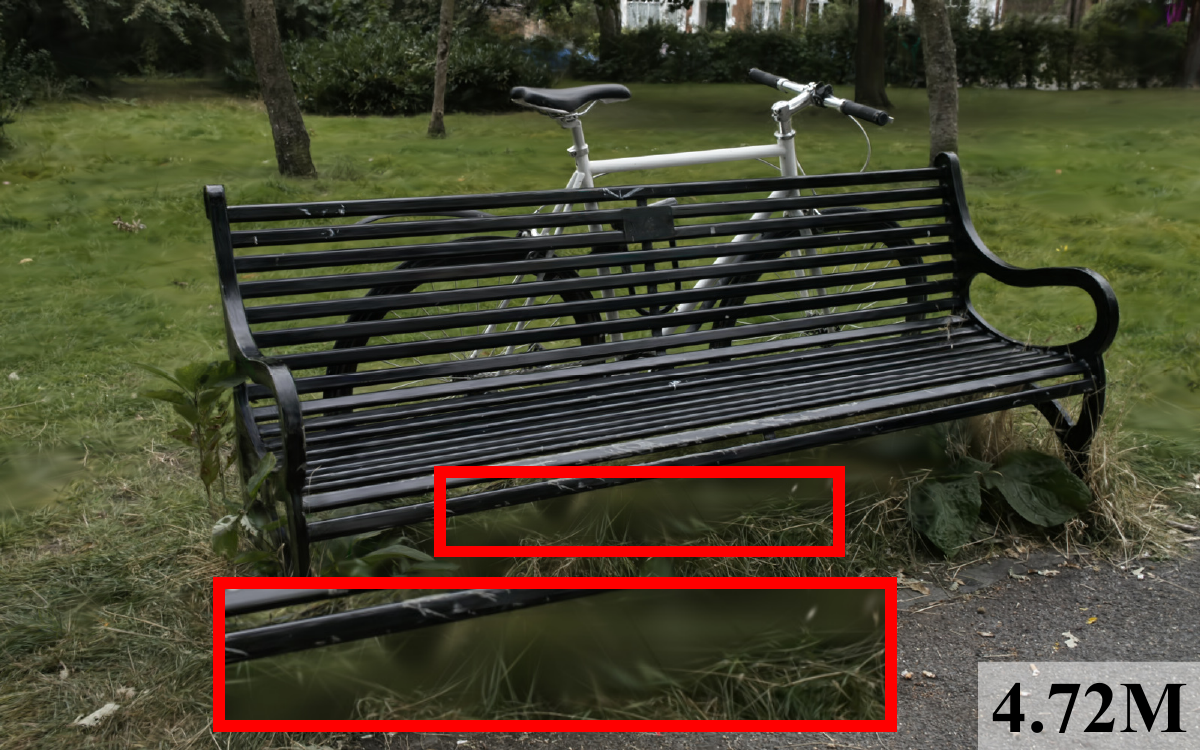}} &
        \fbox{\includegraphics[width=\imgwidth]{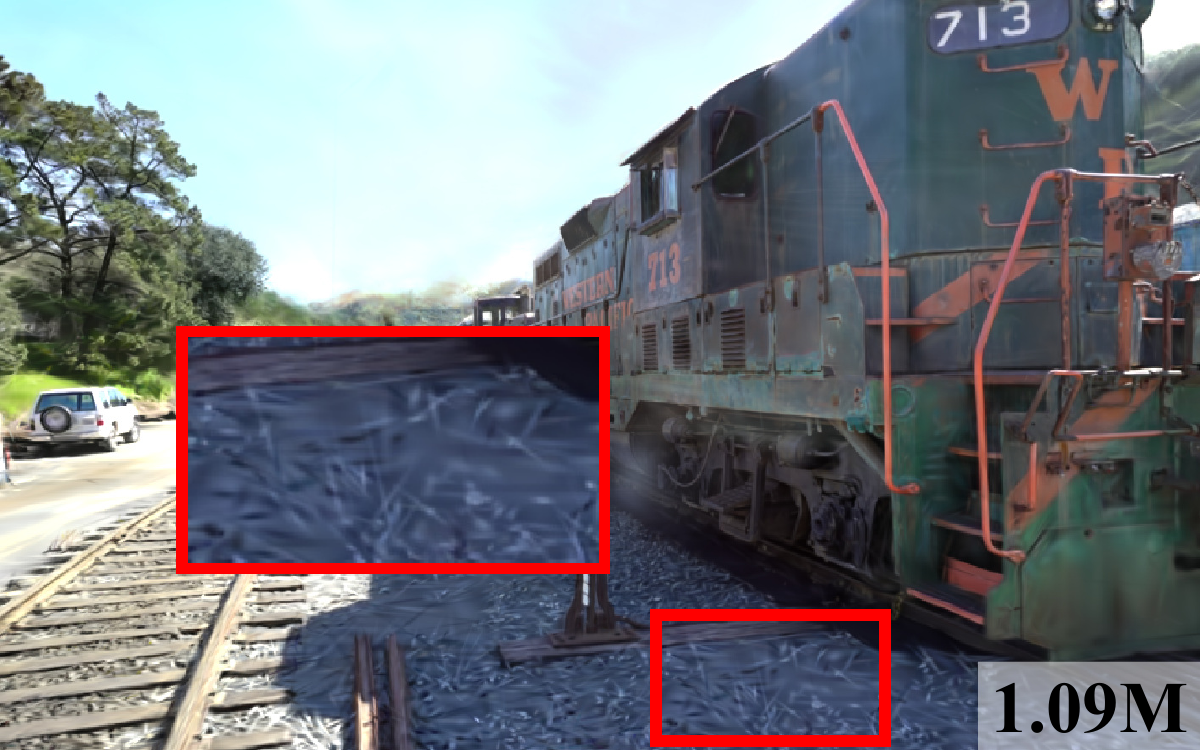}} &
        \fbox{\includegraphics[width=\imgwidth]{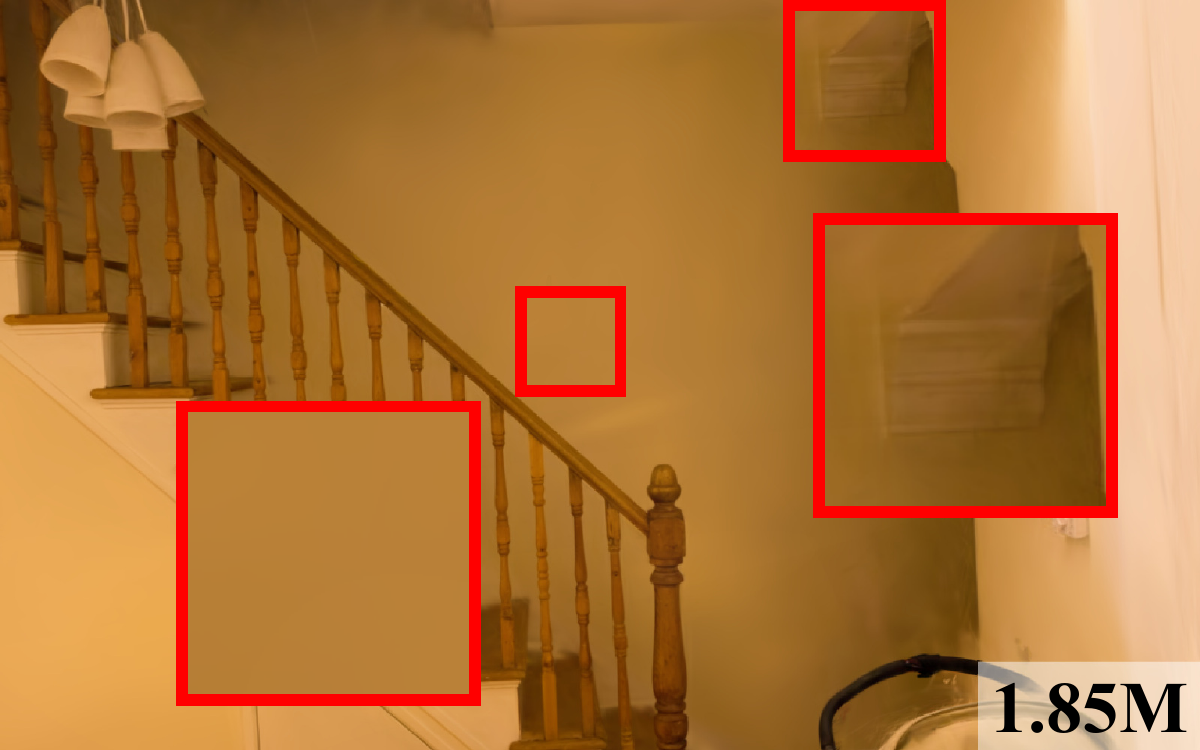}} &
        \fbox{\includegraphics[width=\imgwidth]{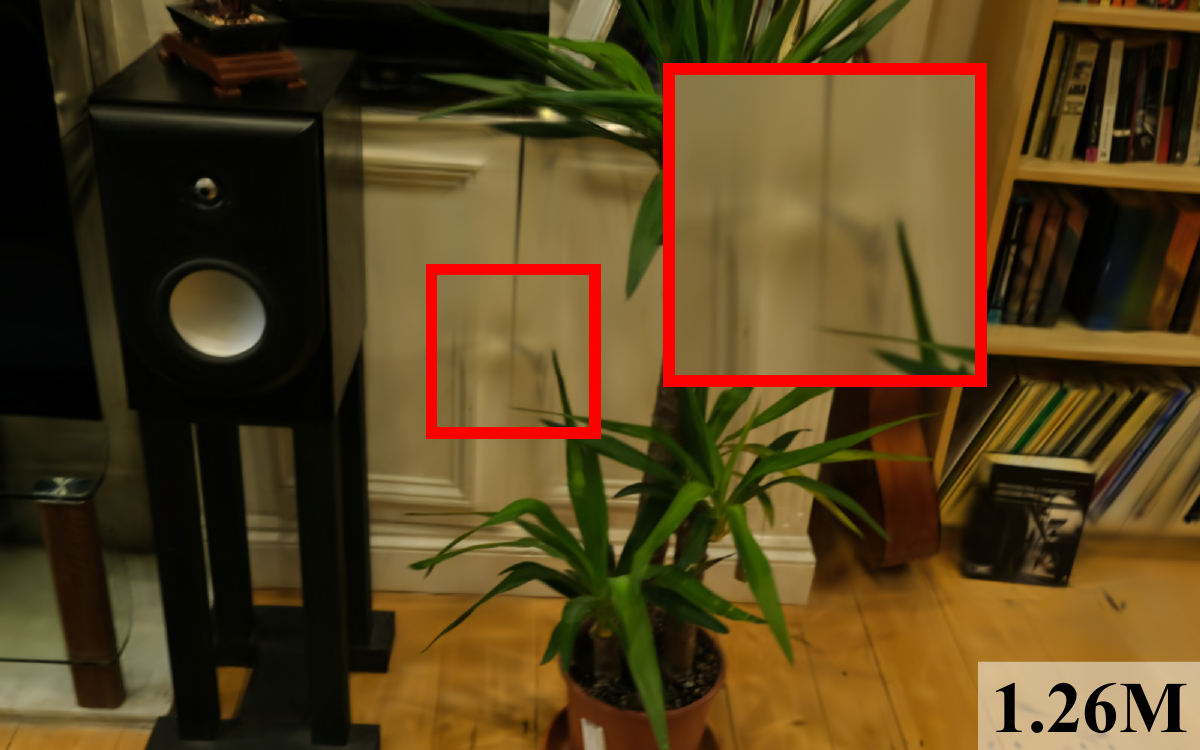}} &
        \fbox{\includegraphics[width=\imgwidth]{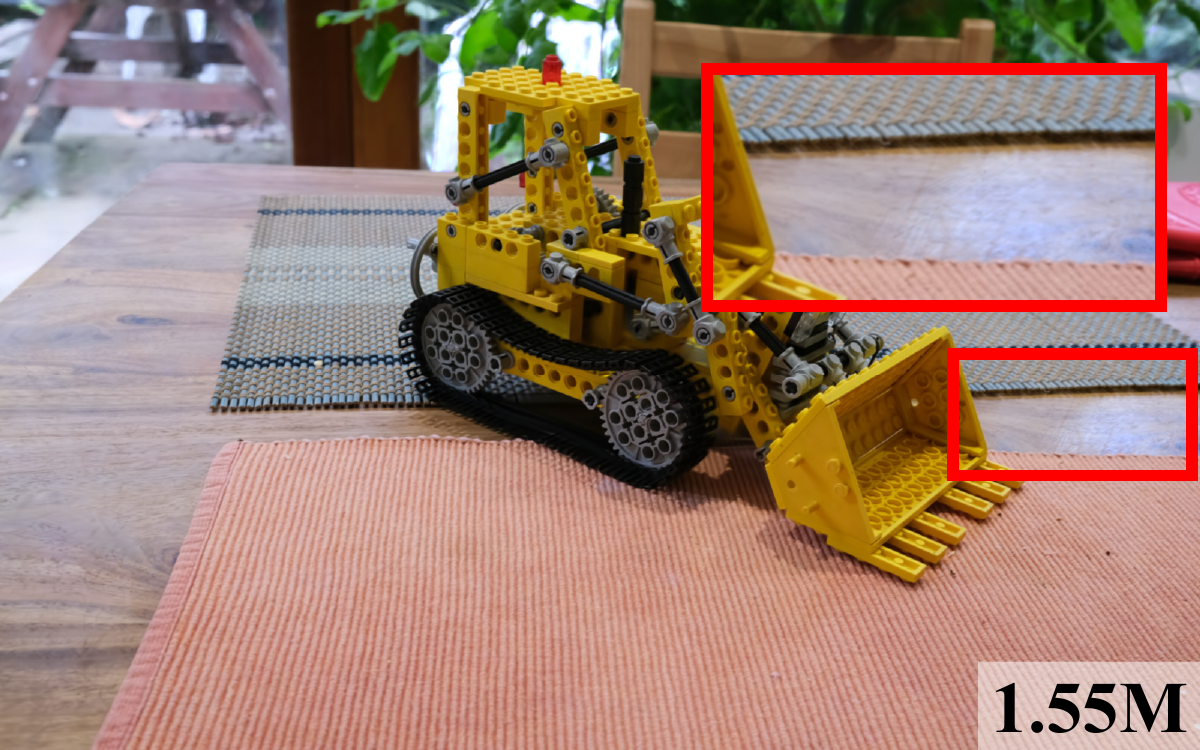}} &
        \fbox{\includegraphics[width=\imgwidth]{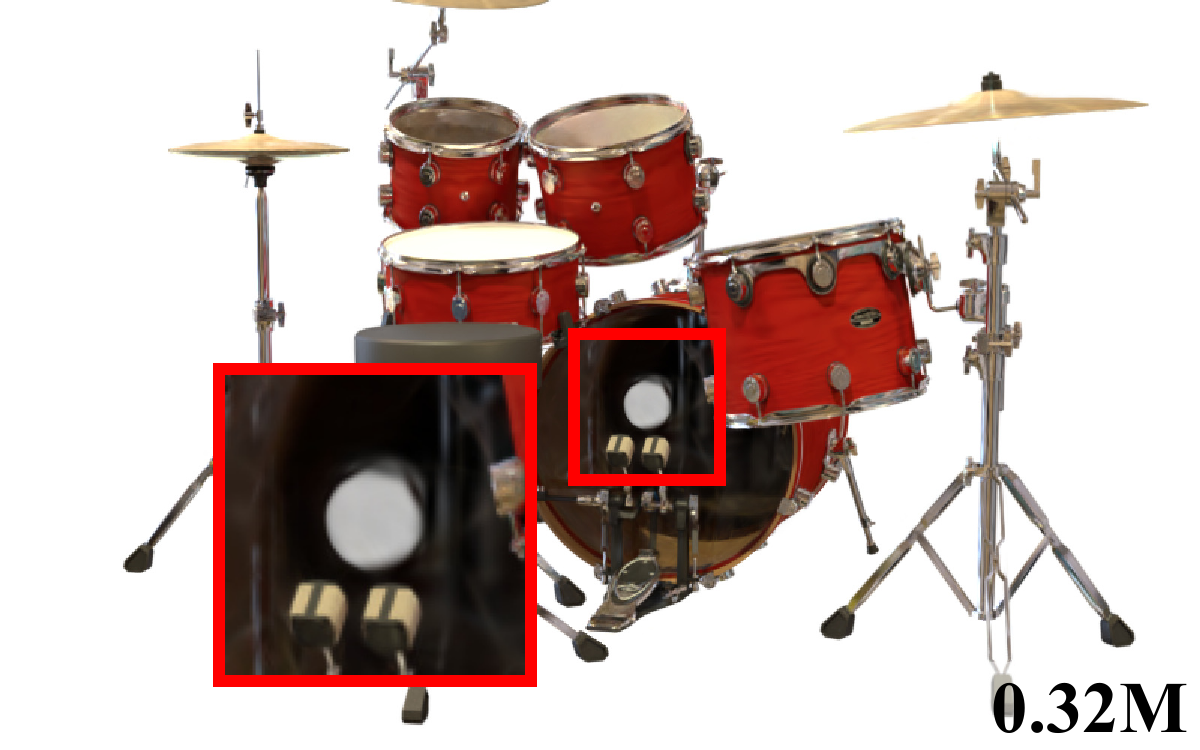}} \\

        \adjustbox{angle=90,lap=0.8em}{\scriptsize\textsf{\textbf{LP-3DGS}}} &
        \fbox{\includegraphics[width=\imgwidth]{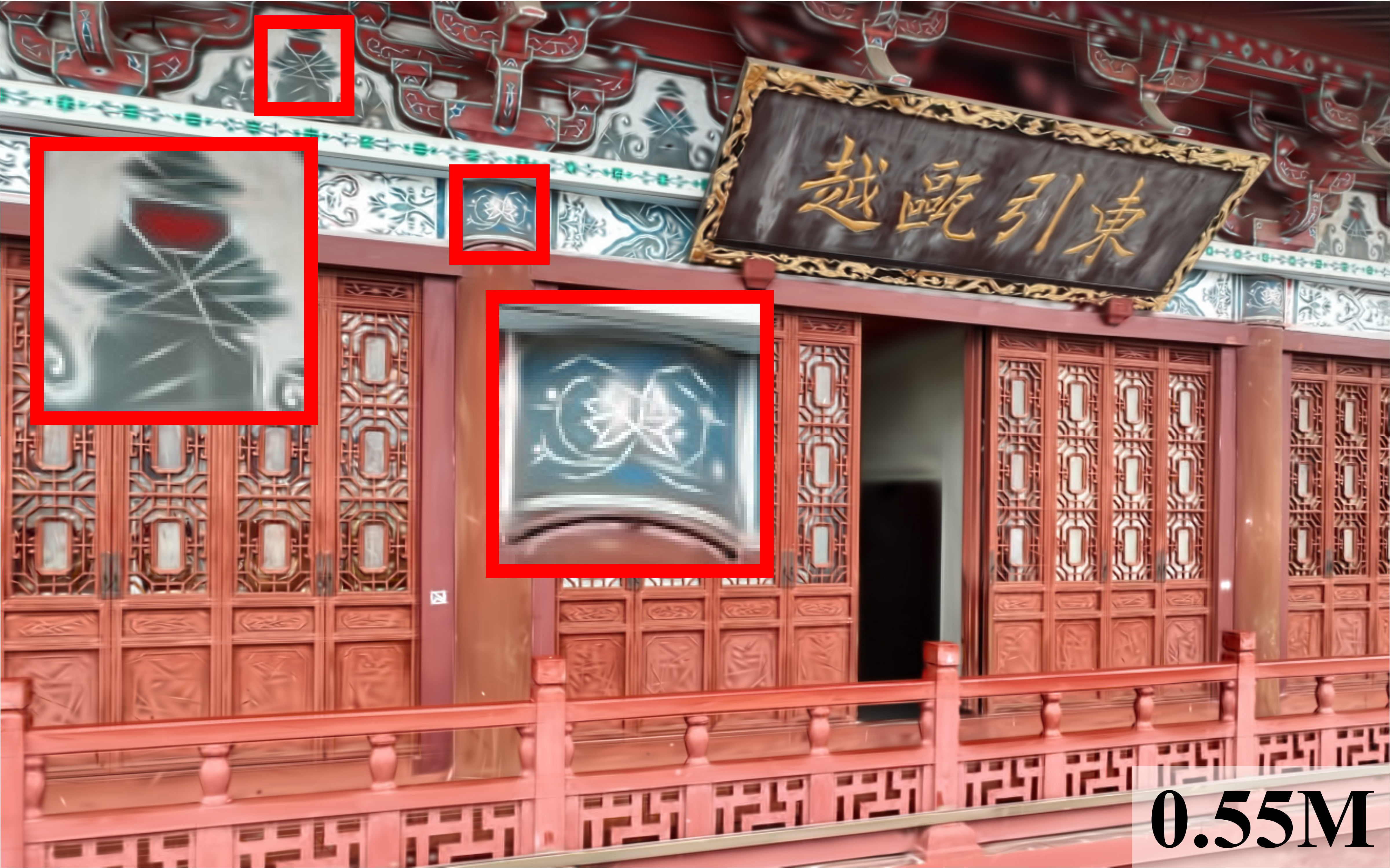}} &
        \fbox{\includegraphics[width=\imgwidth]{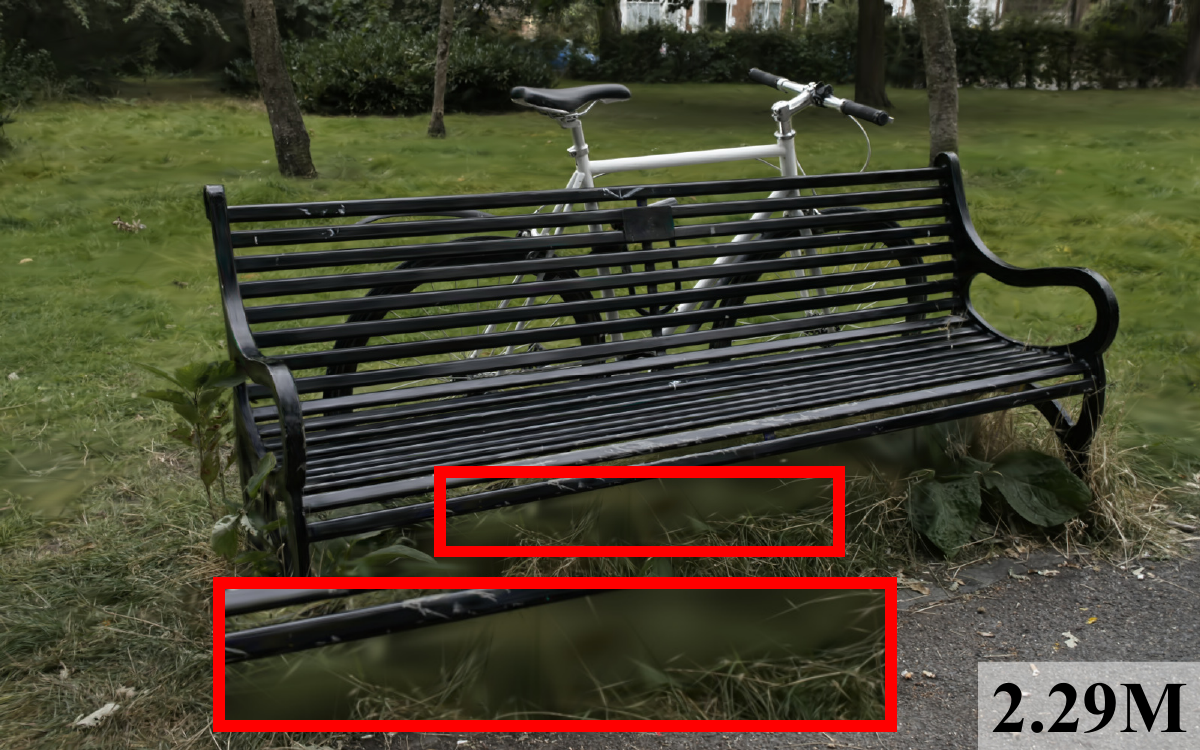}} &
        \fbox{\includegraphics[width=\imgwidth]{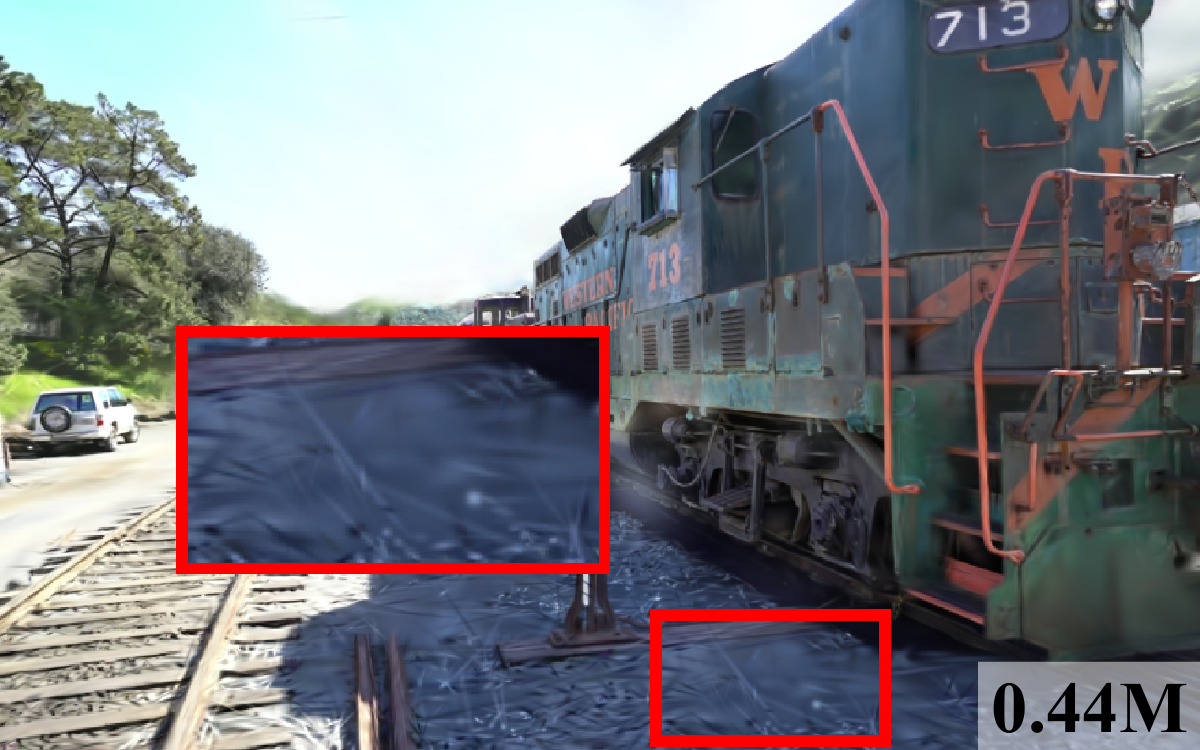}} &
        \fbox{\includegraphics[width=\imgwidth]{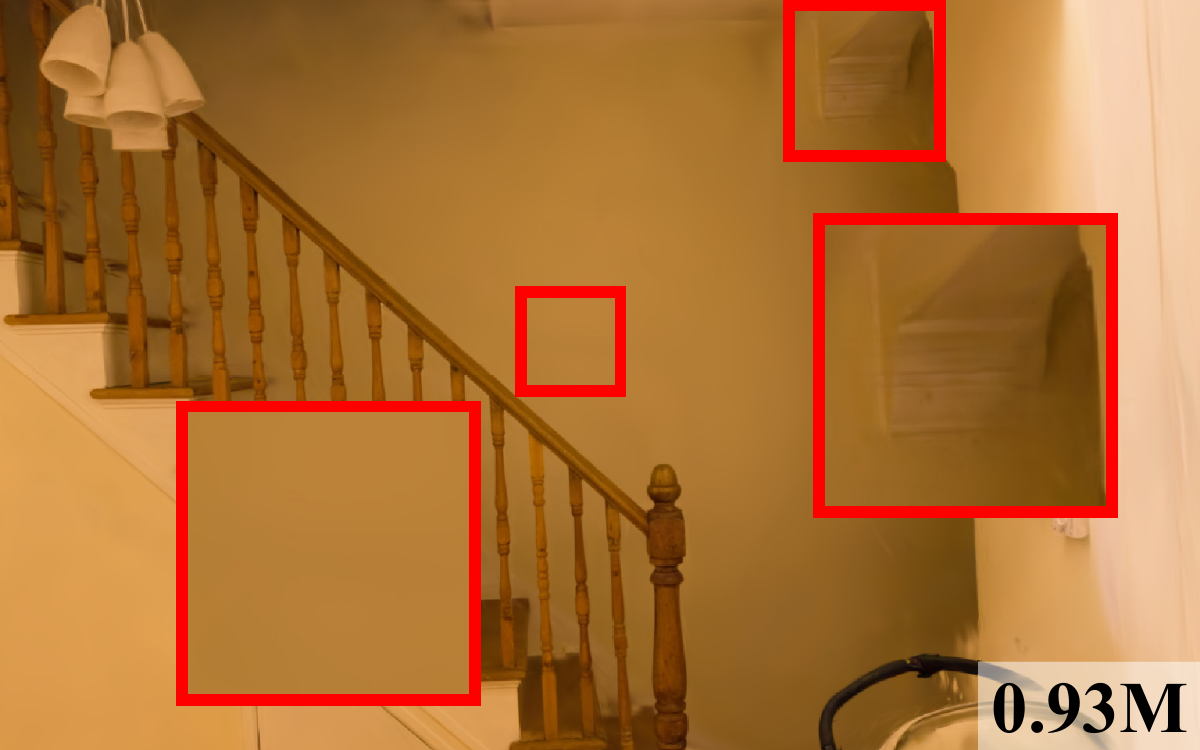}} &
        \fbox{\includegraphics[width=\imgwidth]{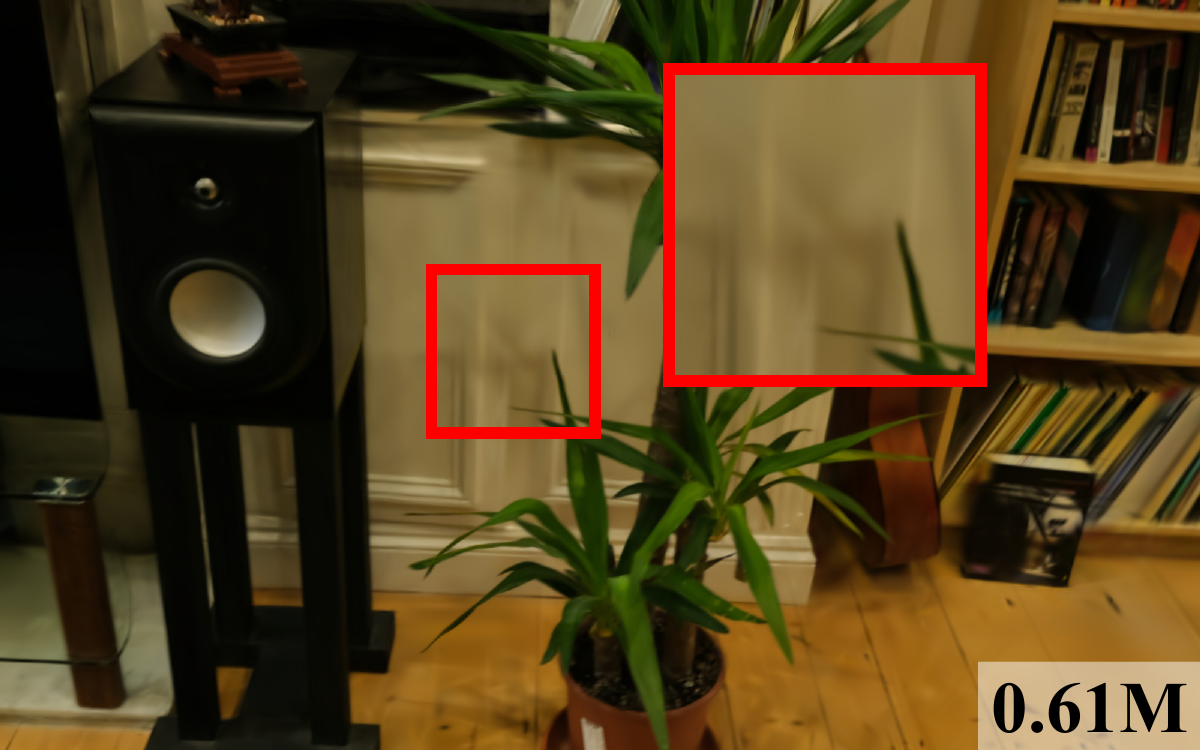}} &
        \fbox{\includegraphics[width=\imgwidth]{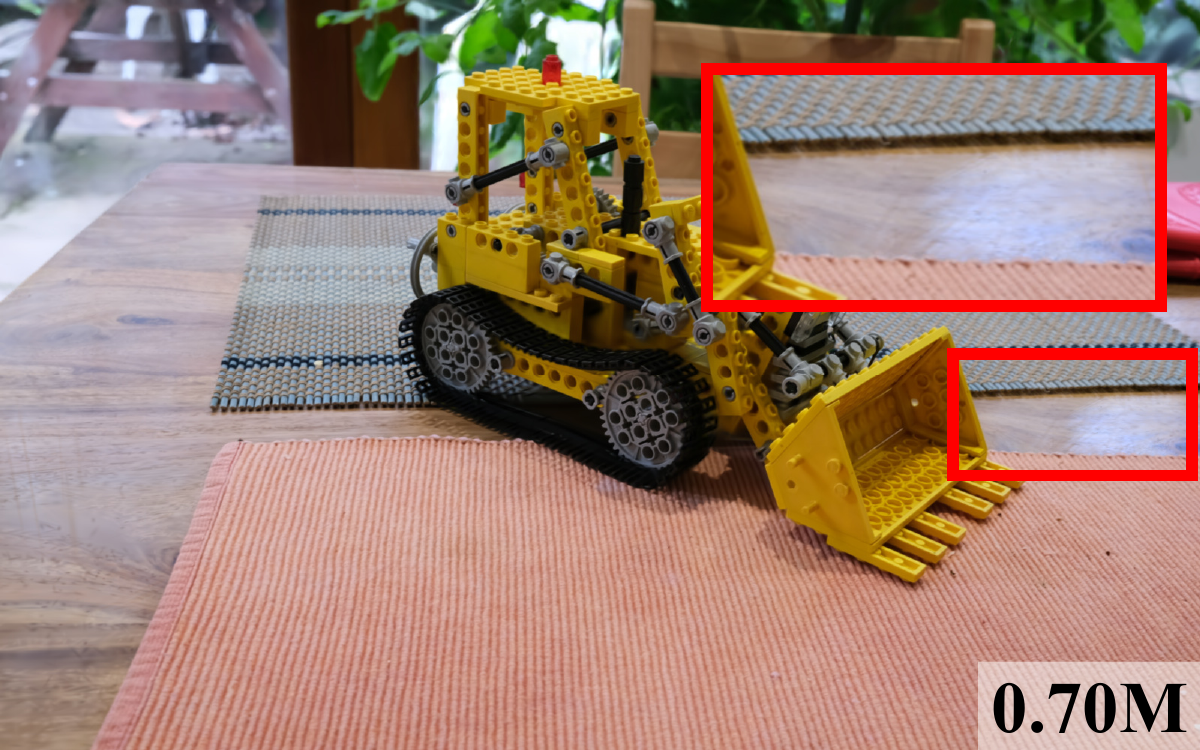}} &
        \fbox{\includegraphics[width=\imgwidth]{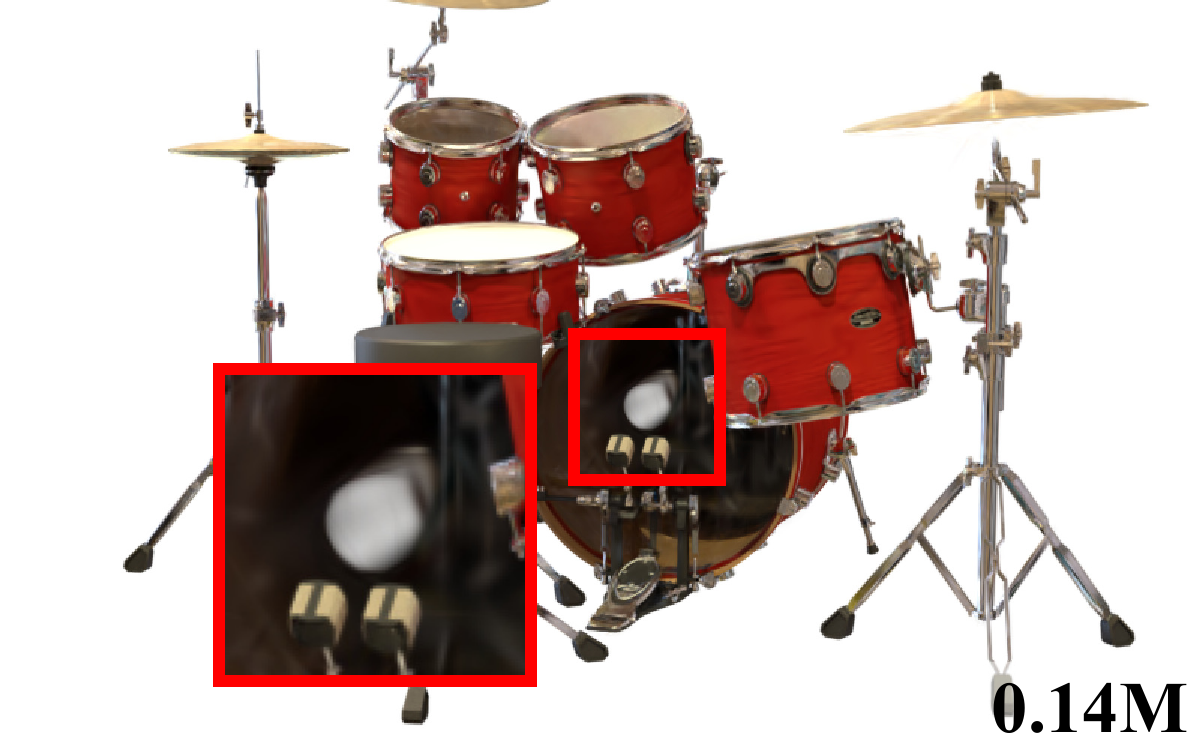}} \\

        \adjustbox{angle=90,lap=0.8em}{\scriptsize\textsf{\textbf{EAGLES}}} &
        \fbox{\includegraphics[width=\imgwidth]{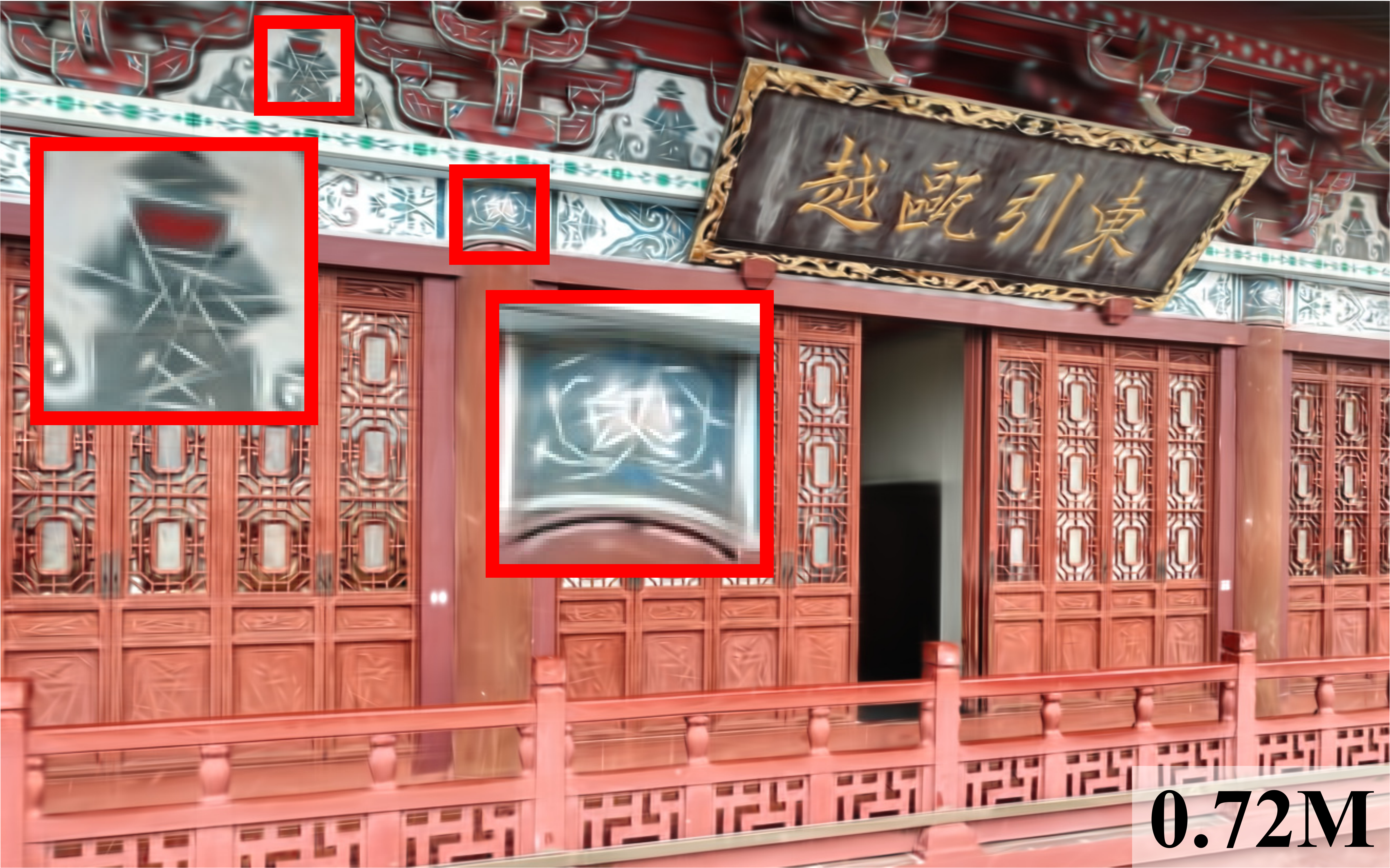}} &
        \fbox{\includegraphics[width=\imgwidth]{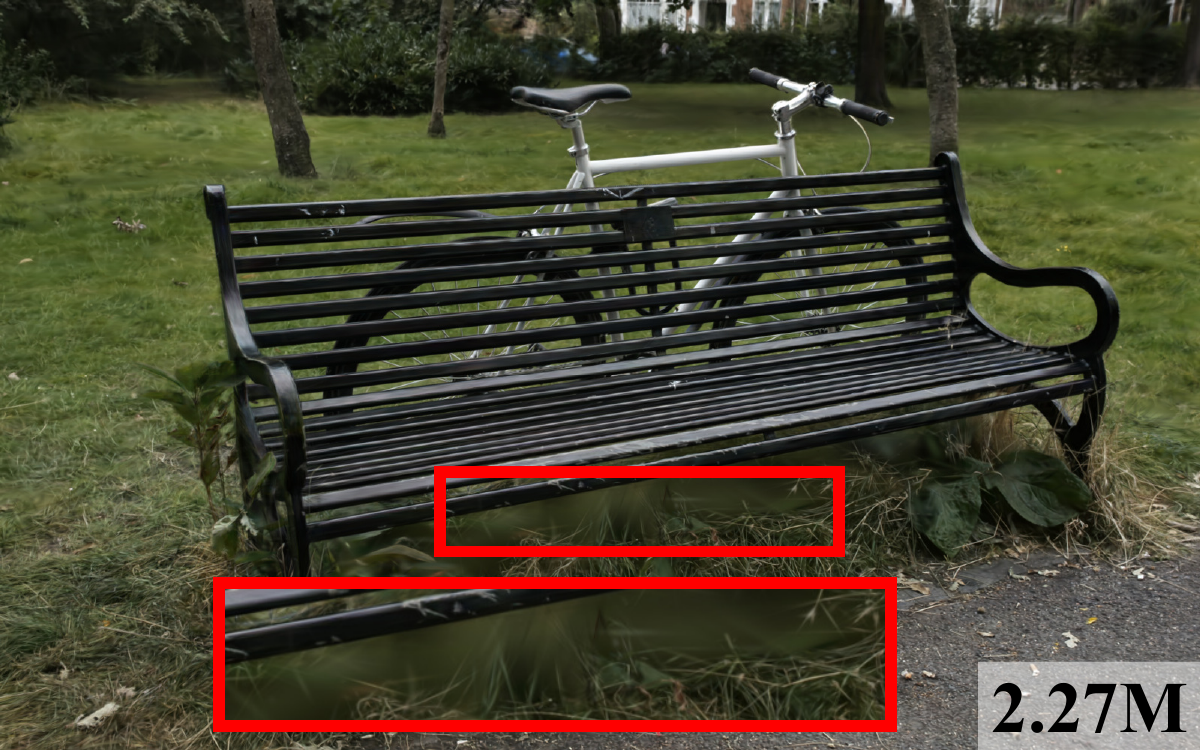}} &
        \fbox{\includegraphics[width=\imgwidth]{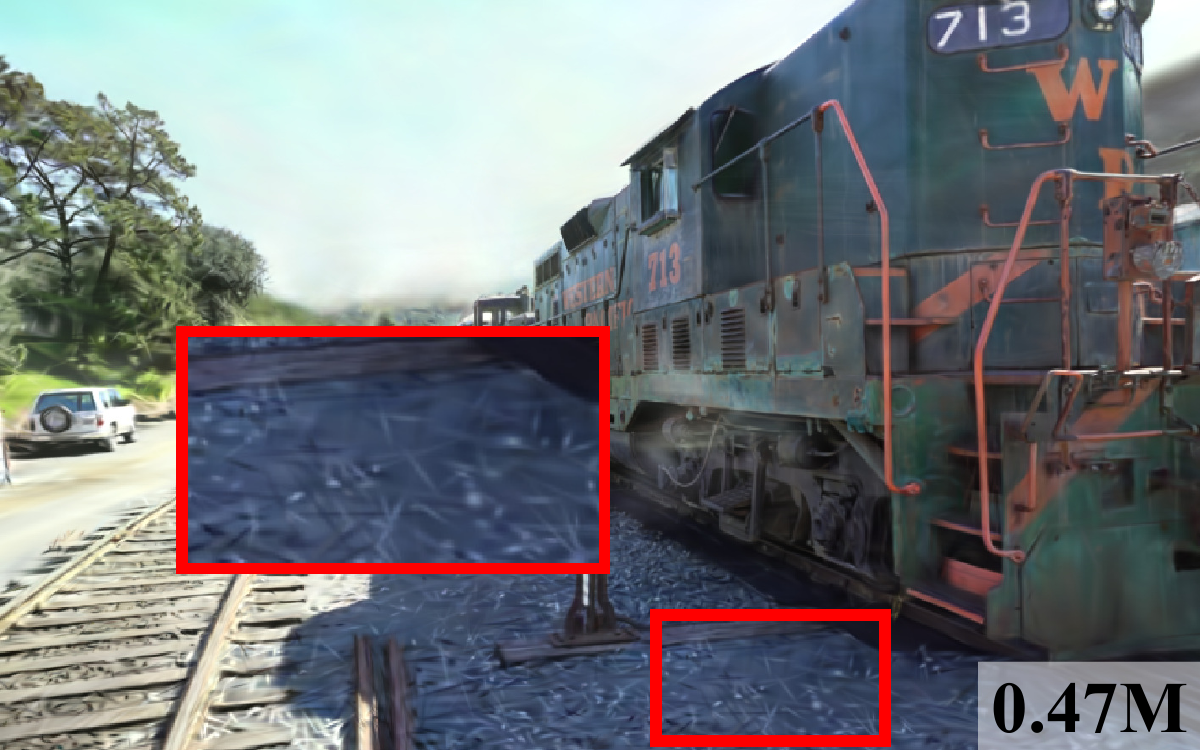}} &
        \fbox{\includegraphics[width=\imgwidth]{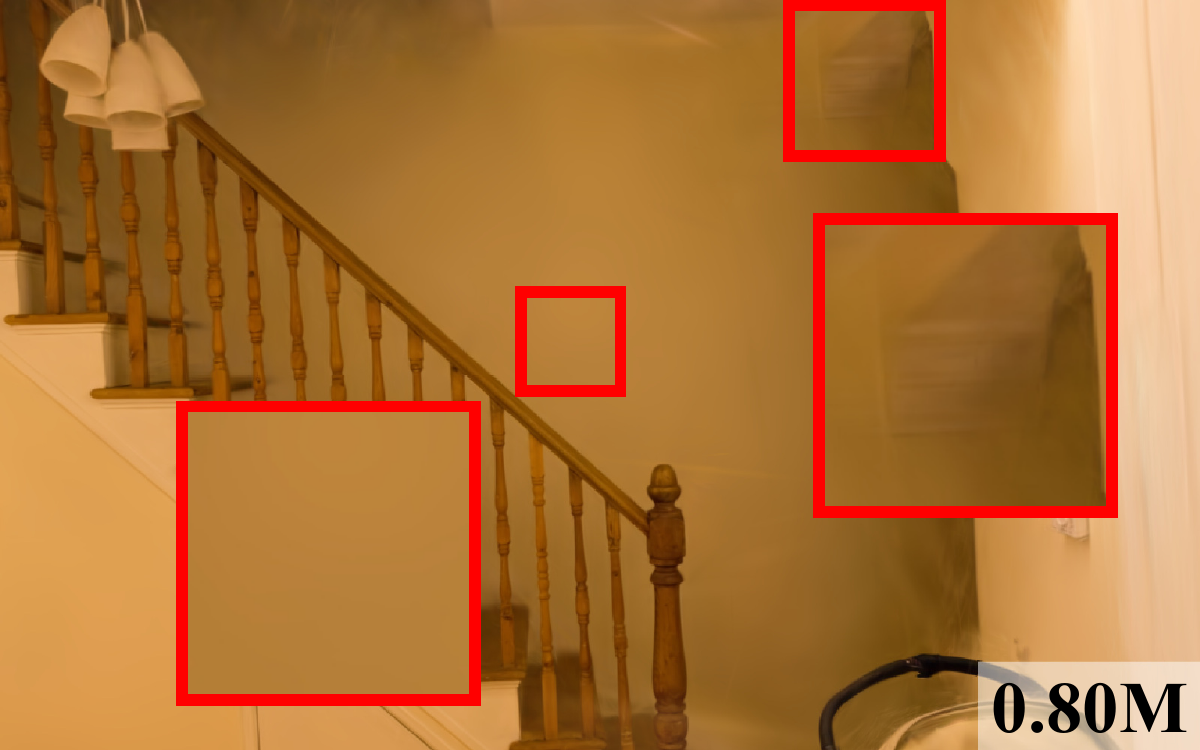}} &
        \fbox{\includegraphics[width=\imgwidth]{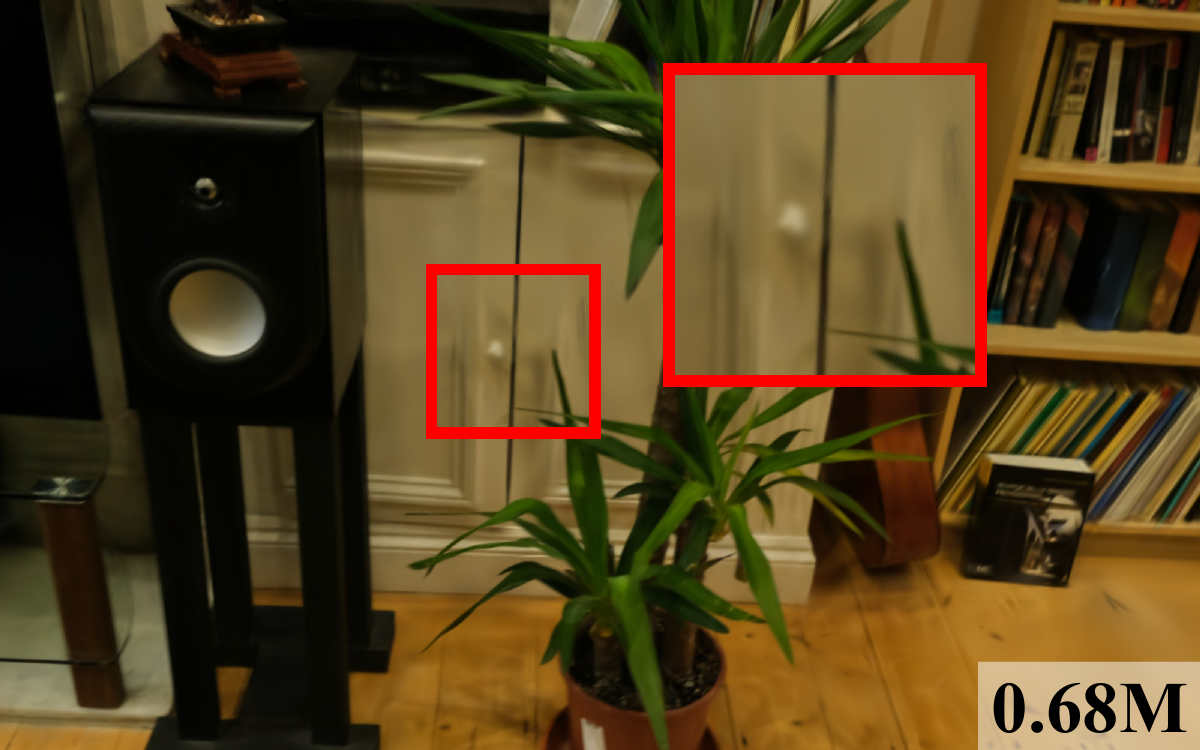}} &
        \fbox{\includegraphics[width=\imgwidth]{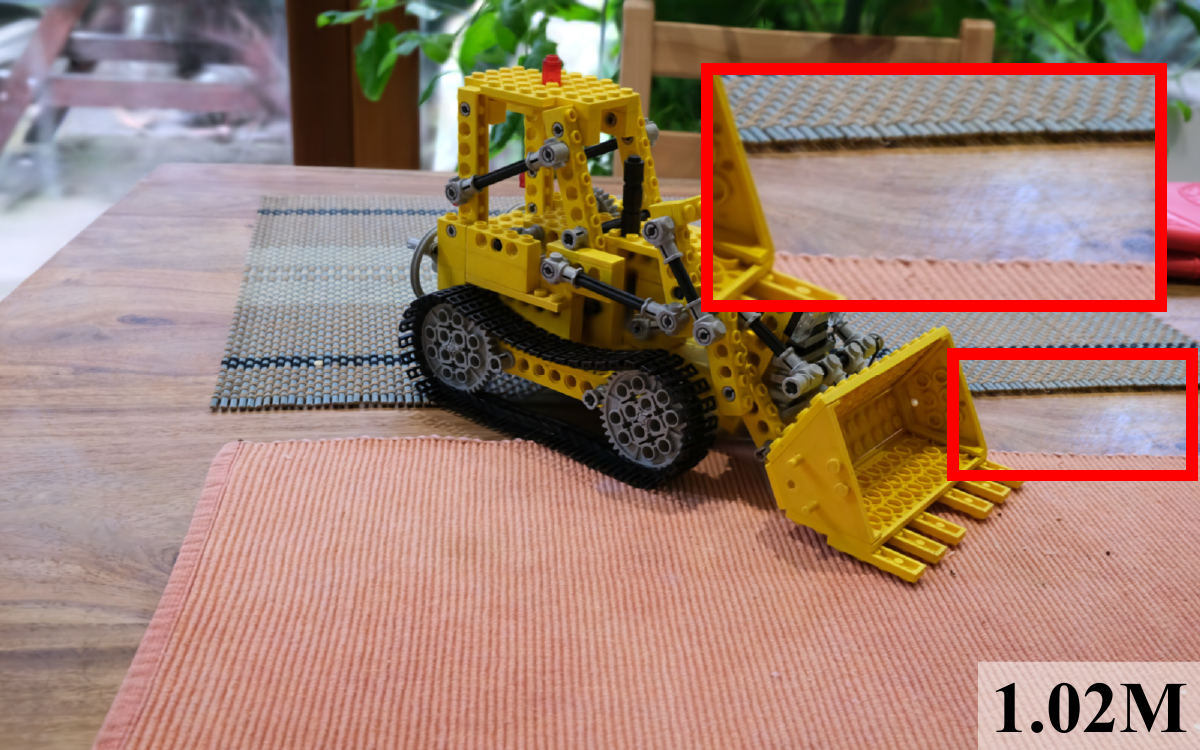}} &
        \fbox{\includegraphics[width=\imgwidth]{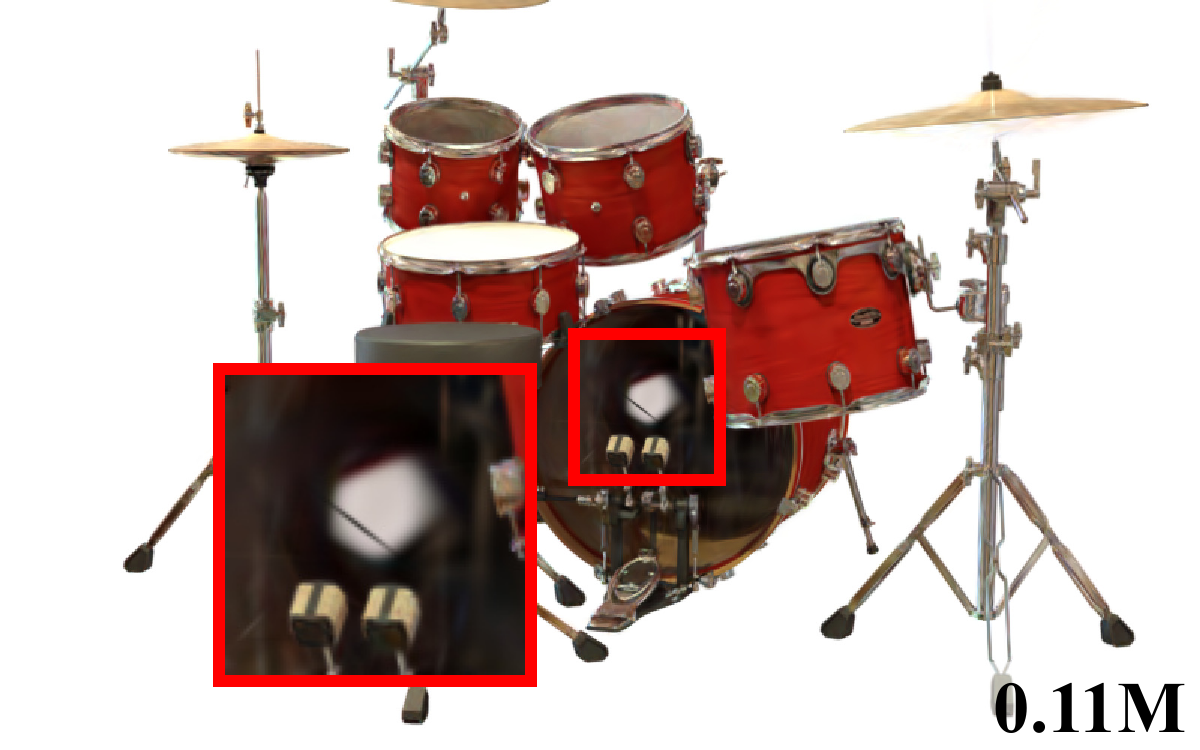}} \\

        & \fixedstack{TW-P-D (View 1) from}{GigaNVS~\cite{GigaNVS}} &
          \fixedstack{Bicycle from}{Mip-NeRF360~\cite{Mip-NeRF360}} &
          \fixedstack{Train from}{Mip-NeRF360~\cite{Mip-NeRF360}} &
          \fixedstack{Playroom from}{Deep Blending~\cite{DeepBlending}} &
          \fixedstack{Room (View 1) from}{Tank \& Temples~\cite{TanksAndTemples}} &
          \fixedstack{Kitchen from}{Mip-NeRF360~\cite{Mip-NeRF360}} &
          \fixedstack{Drums from}{NeRF Synthetic~\cite{NeRF}} \\
          [10pt]
        
        \adjustbox{angle=90,lap=0.8em}{\scriptsize\textsf{\textbf{GT}}} &
        \fbox{\includegraphics[width=\imgwidth]{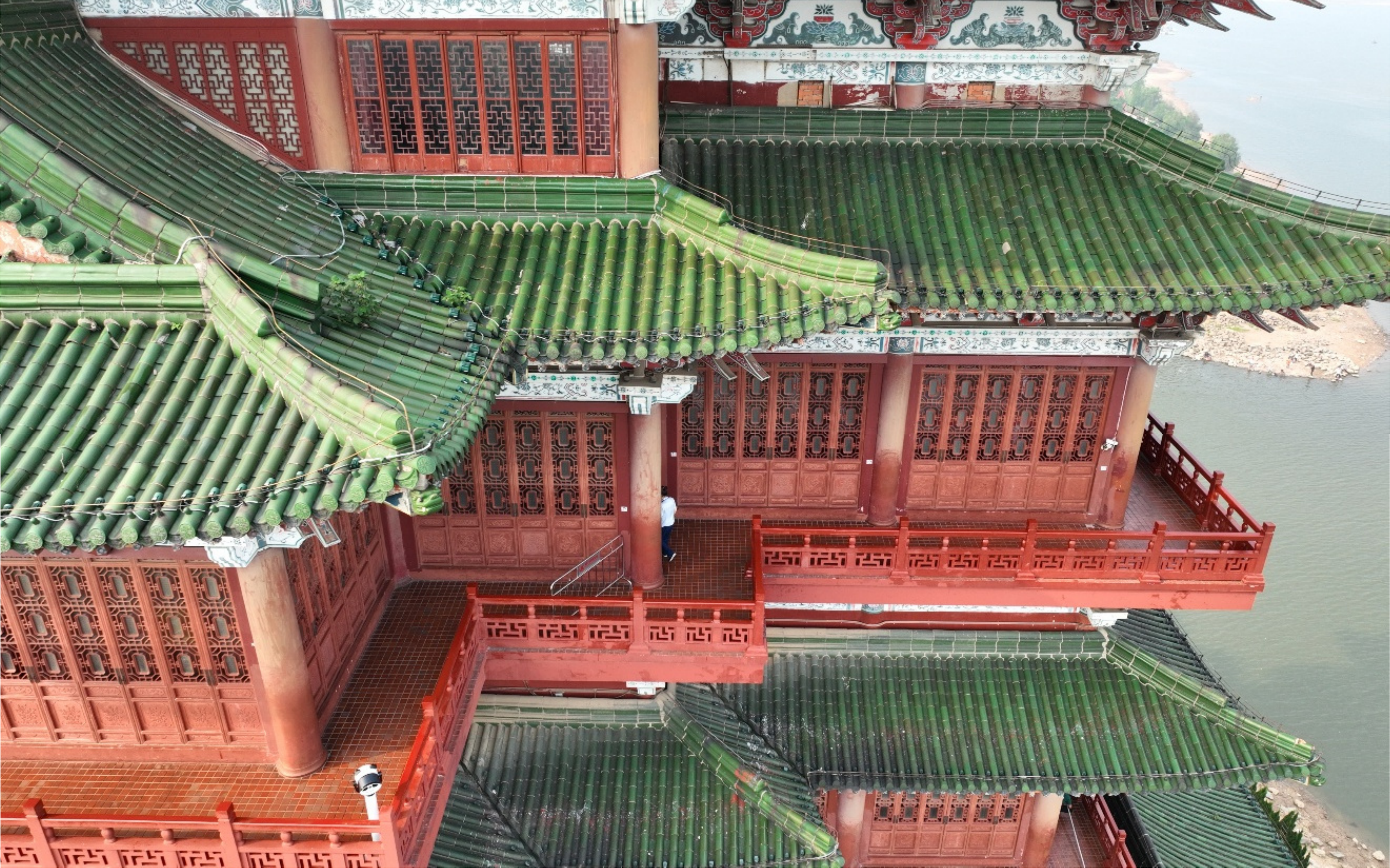}} &
        \fbox{\includegraphics[width=\imgwidth]{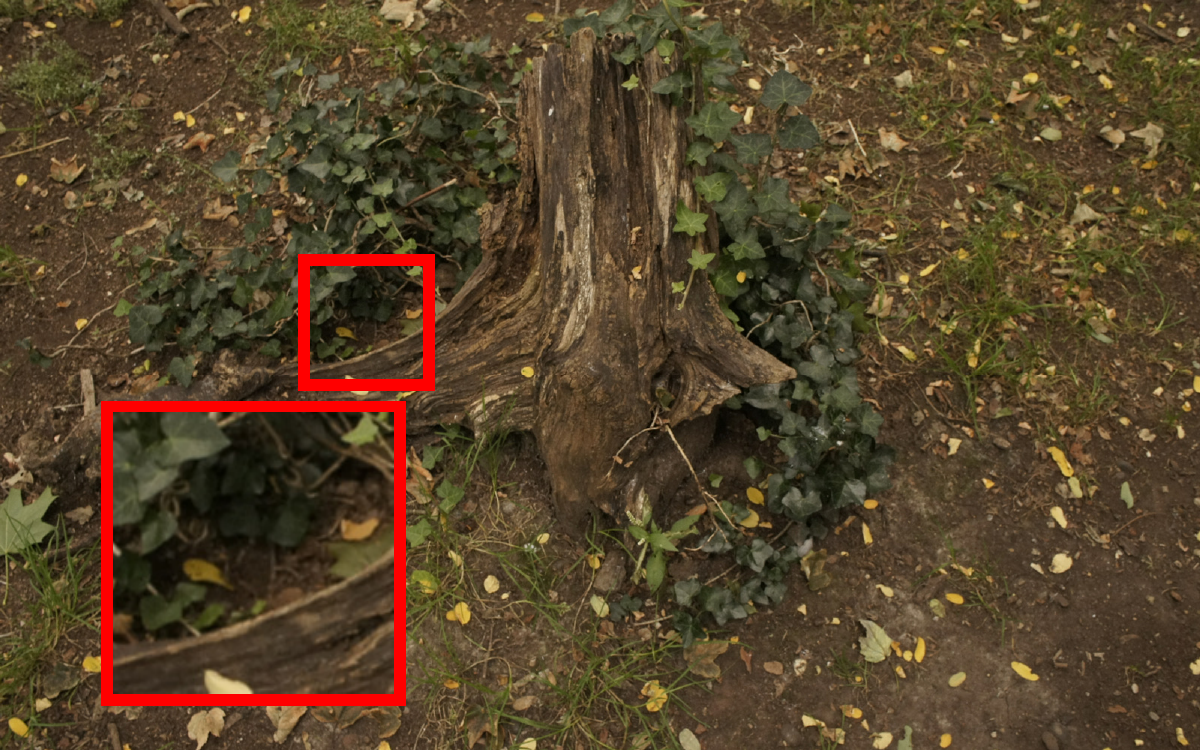}} &
        \fbox{\includegraphics[width=\imgwidth]{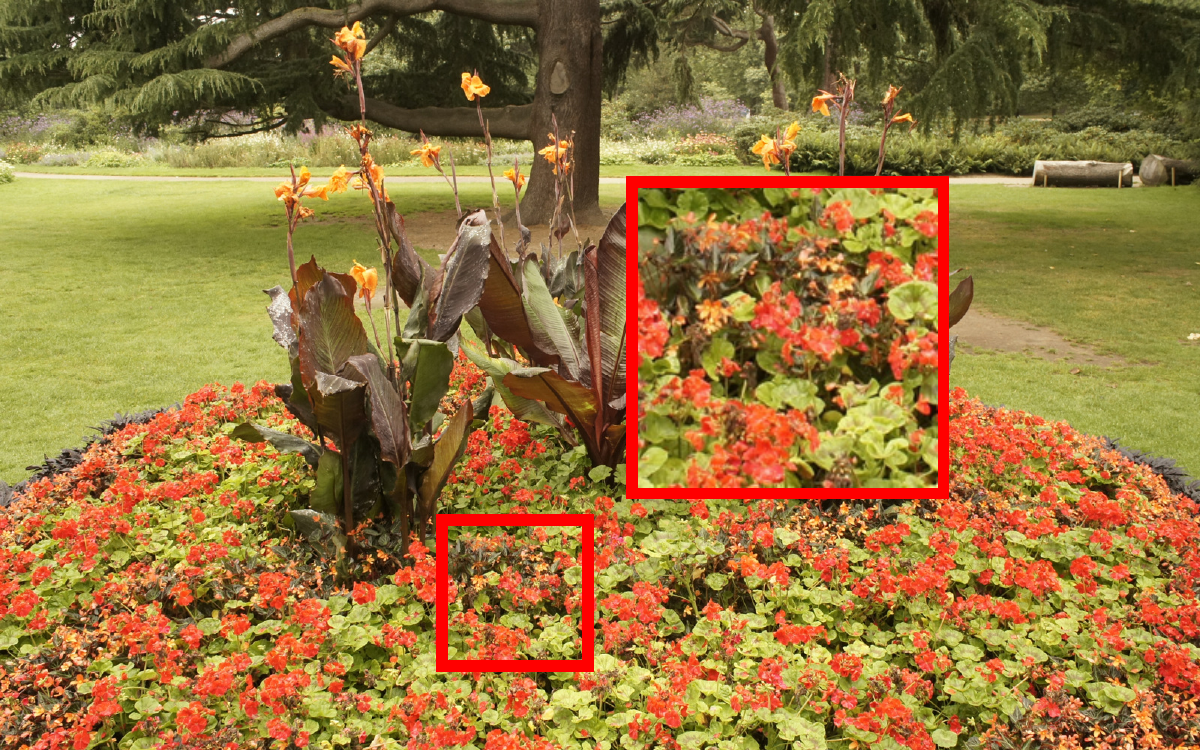}} &
        \fbox{\includegraphics[width=\imgwidth]{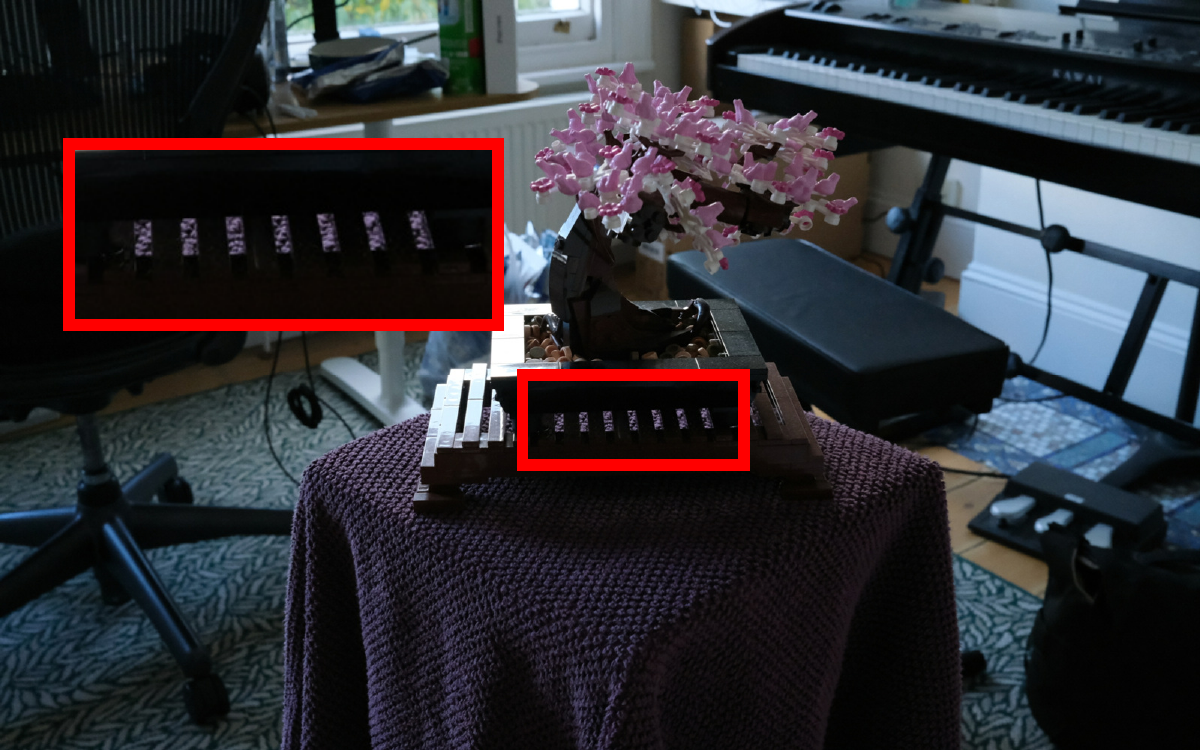}} &
        \fbox{\includegraphics[width=\imgwidth]{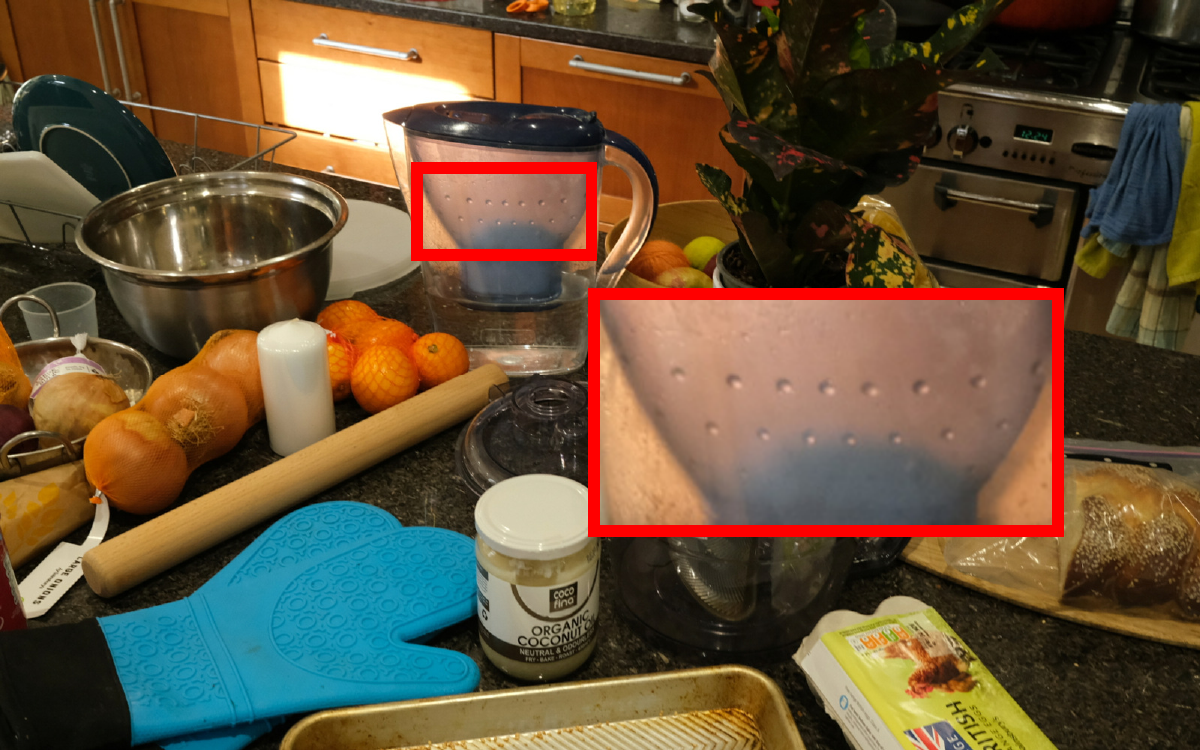}} &
        \fbox{\includegraphics[width=\imgwidth]{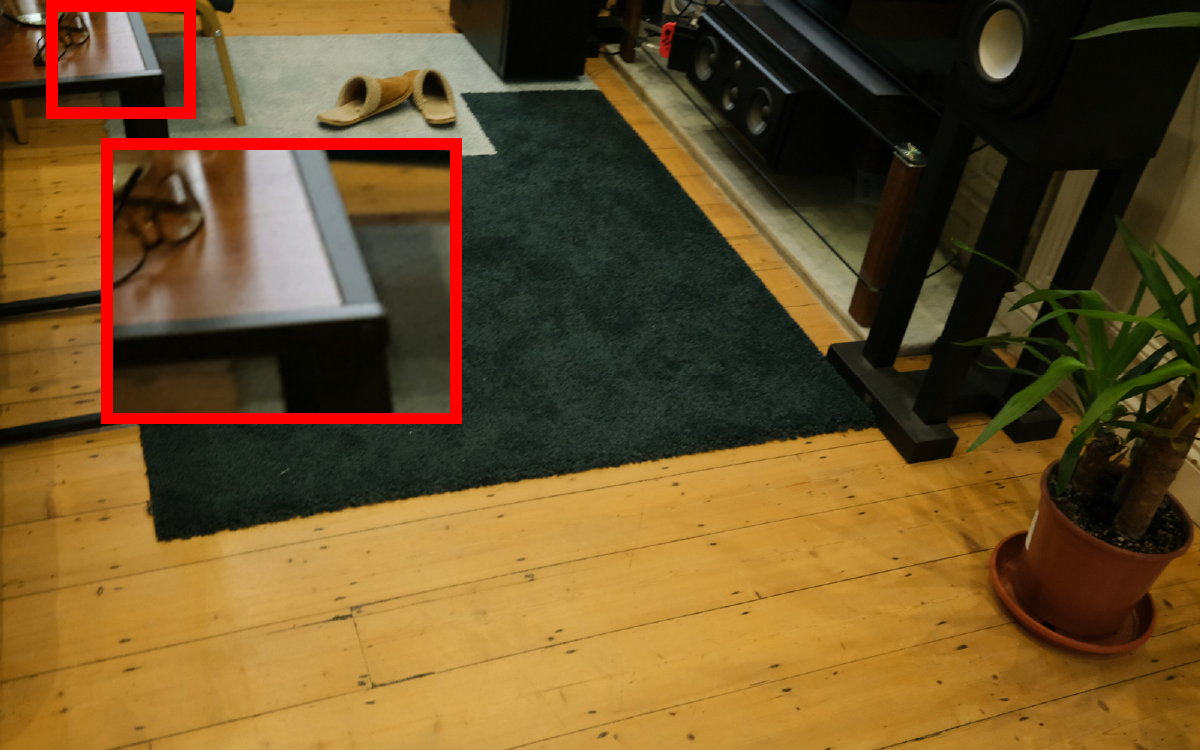}} &
        \fbox{\includegraphics[width=\imgwidth]{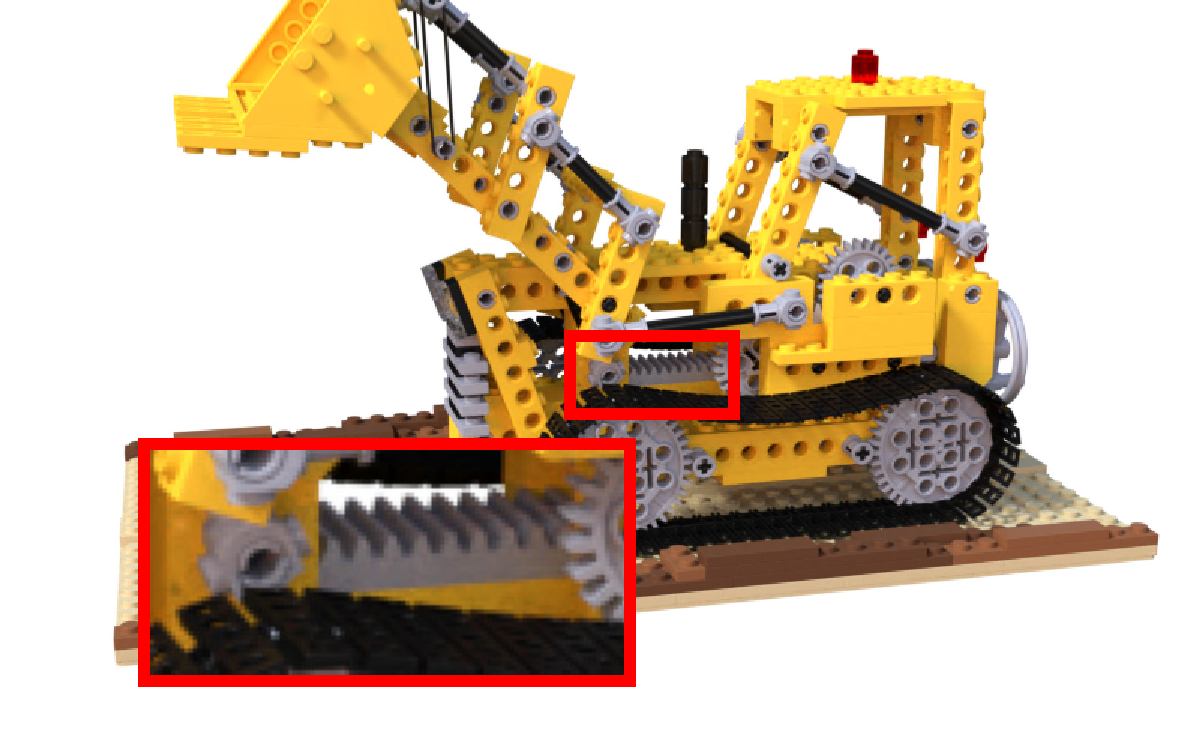}} \\

        \adjustbox{angle=90,lap=0.8em}{\scriptsize\textsf{\textbf{Ours}}} &
        \fbox{\includegraphics[width=\imgwidth]{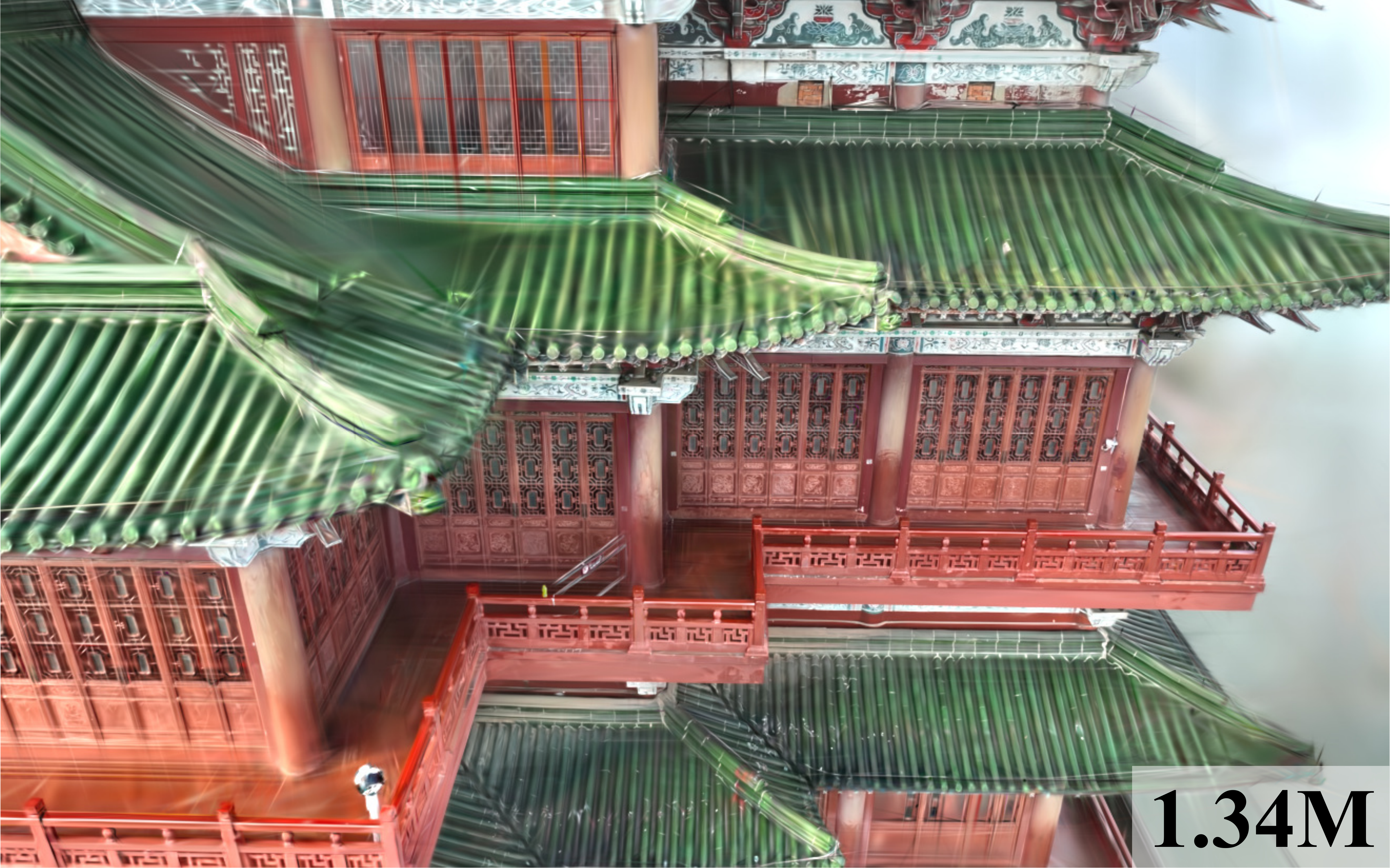}} &
        \fbox{\includegraphics[width=\imgwidth]{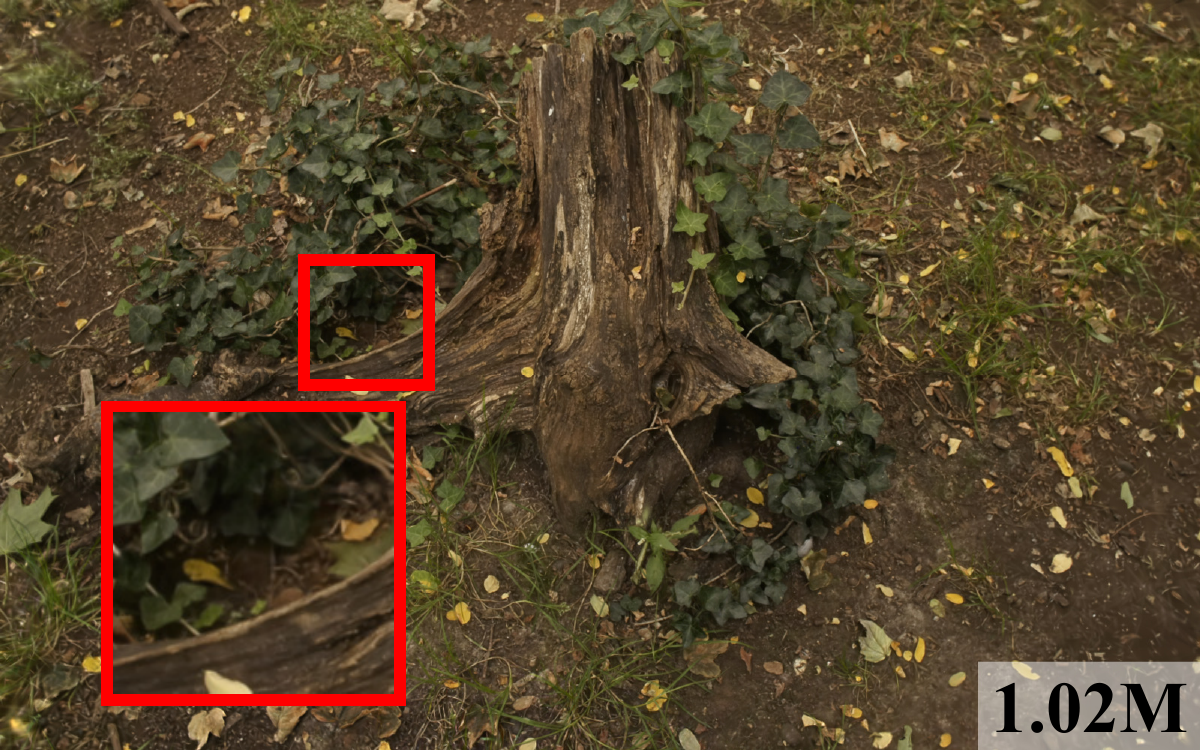}} &
        \fbox{\includegraphics[width=\imgwidth]{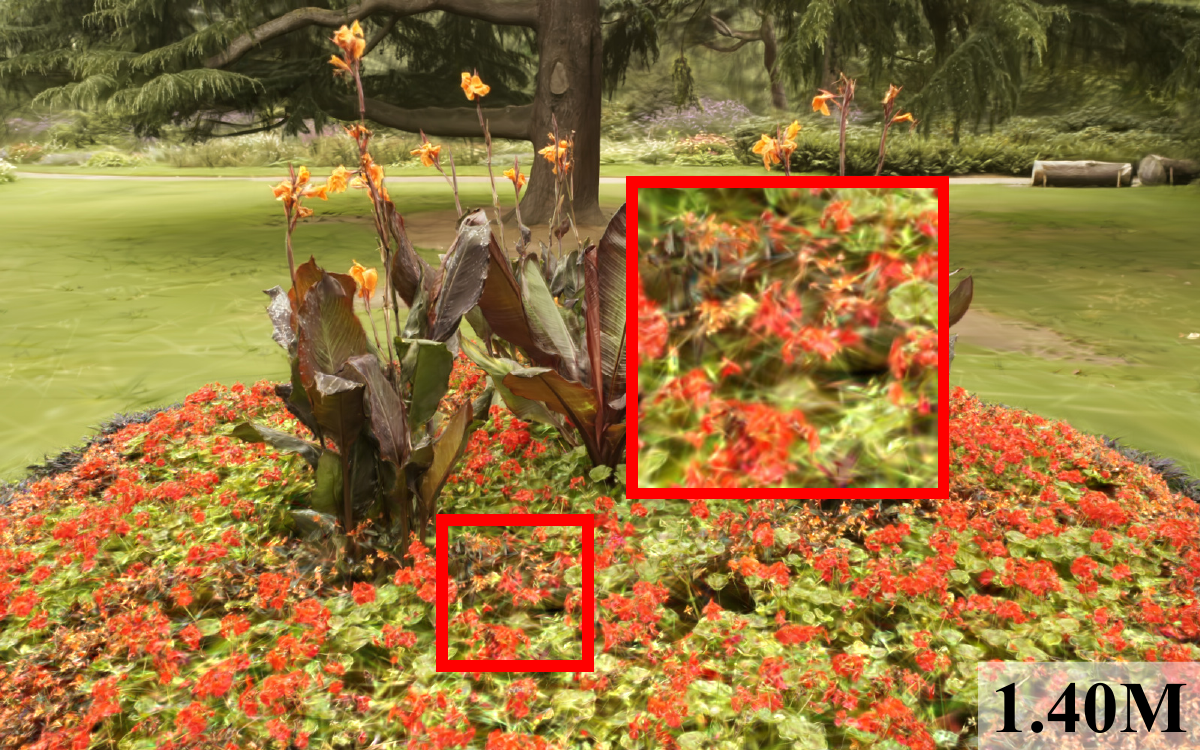}} &
        \fbox{\includegraphics[width=\imgwidth]{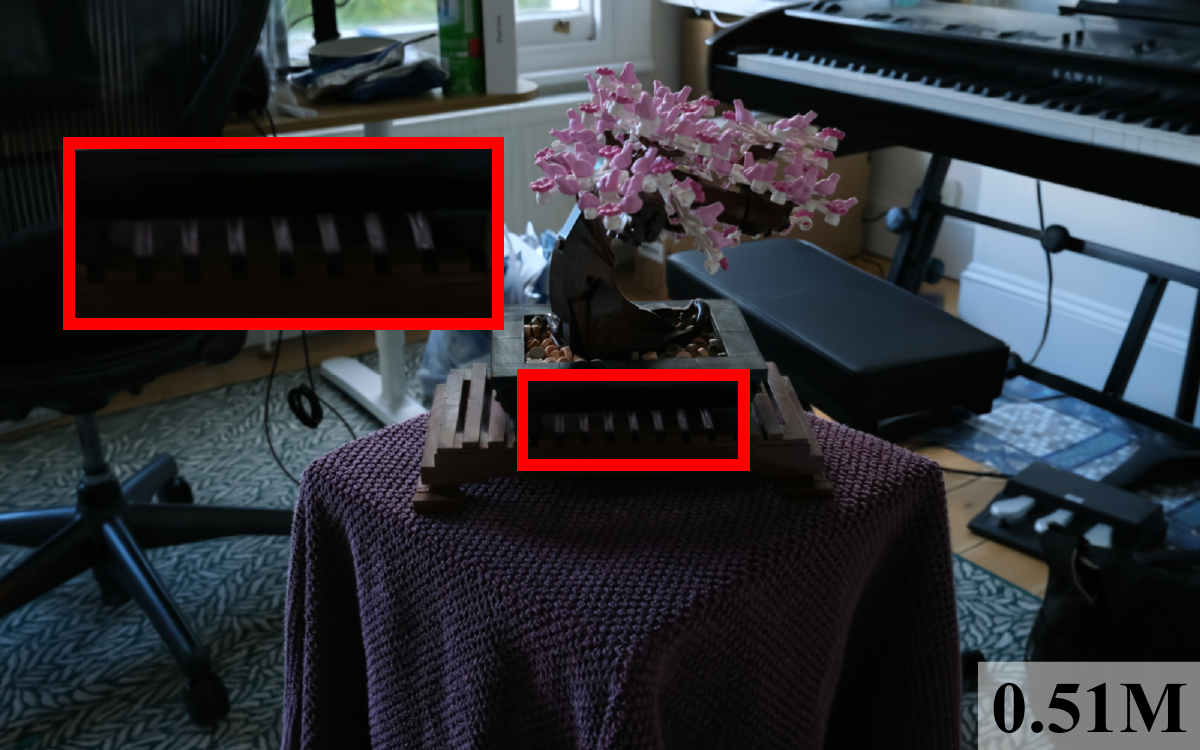}} &
        \fbox{\includegraphics[width=\imgwidth]{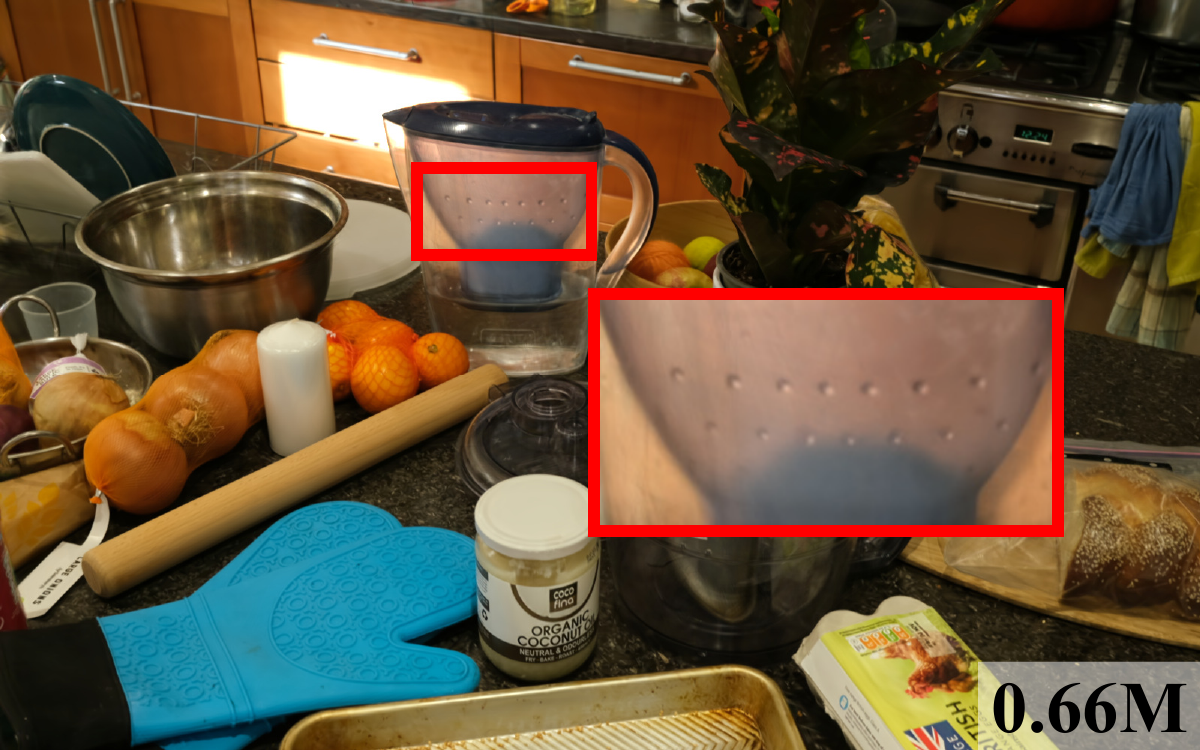}} &
        \fbox{\includegraphics[width=\imgwidth]{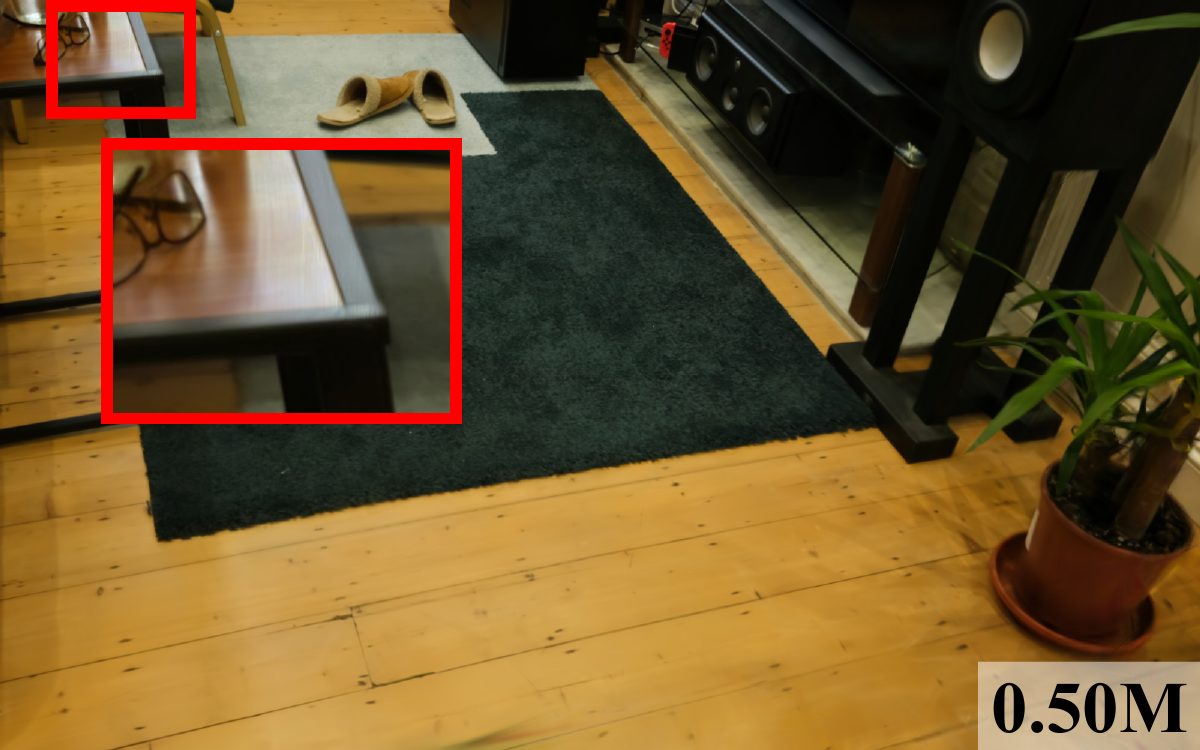}} &
        \fbox{\includegraphics[width=\imgwidth]{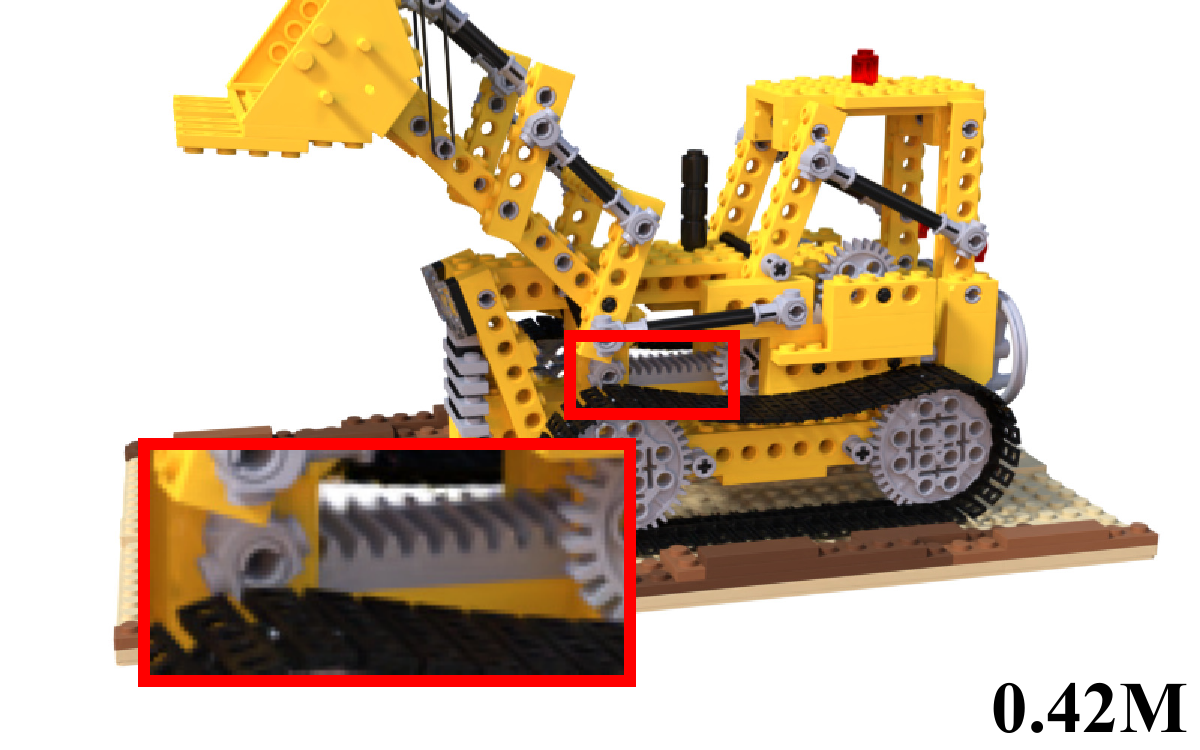}} \\

        \adjustbox{angle=90,lap=0.8em}{\scriptsize\textsf{\textbf{3DGS}}} &
        \fbox{\includegraphics[width=\imgwidth]{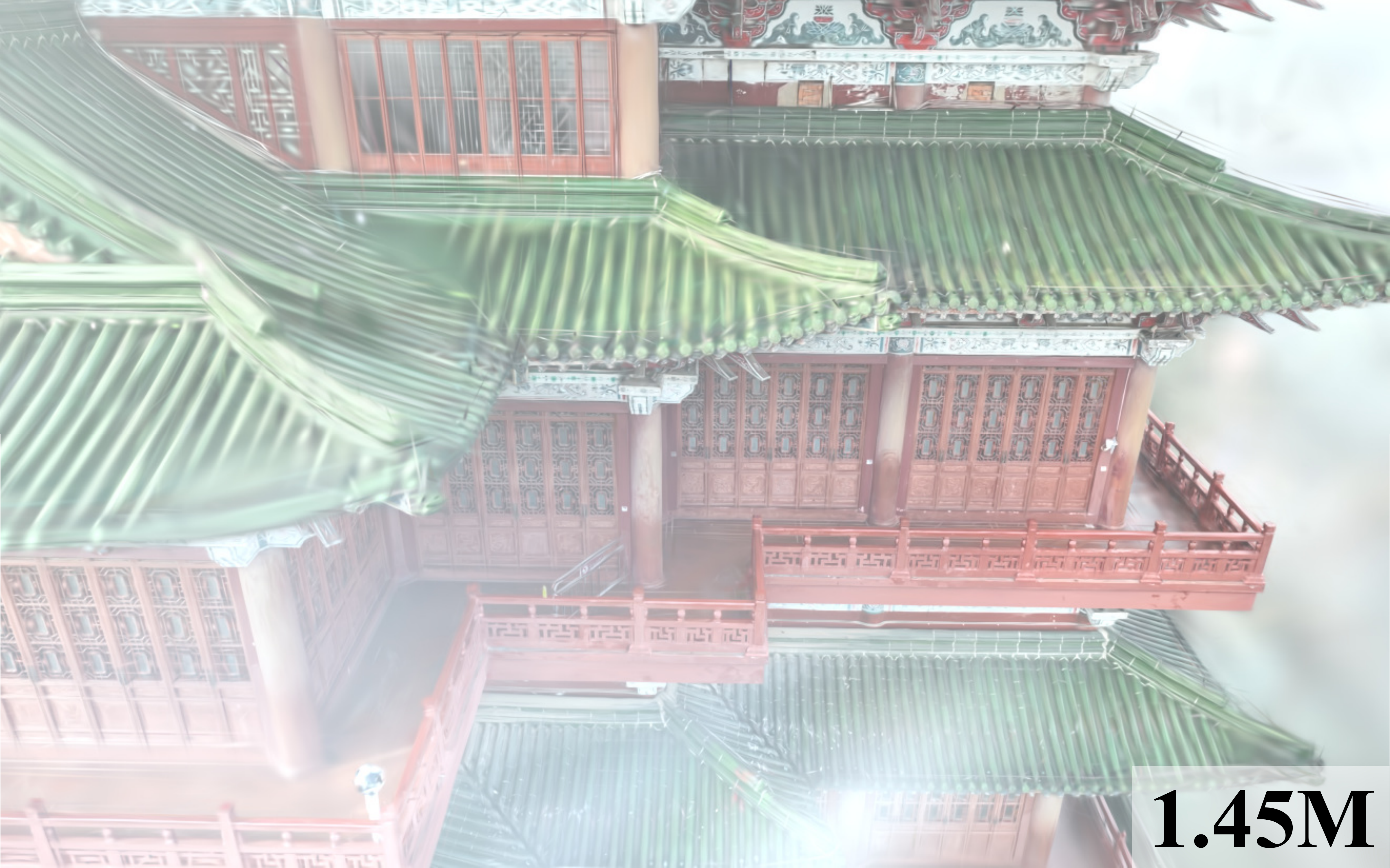}} &
        \fbox{\includegraphics[width=\imgwidth]{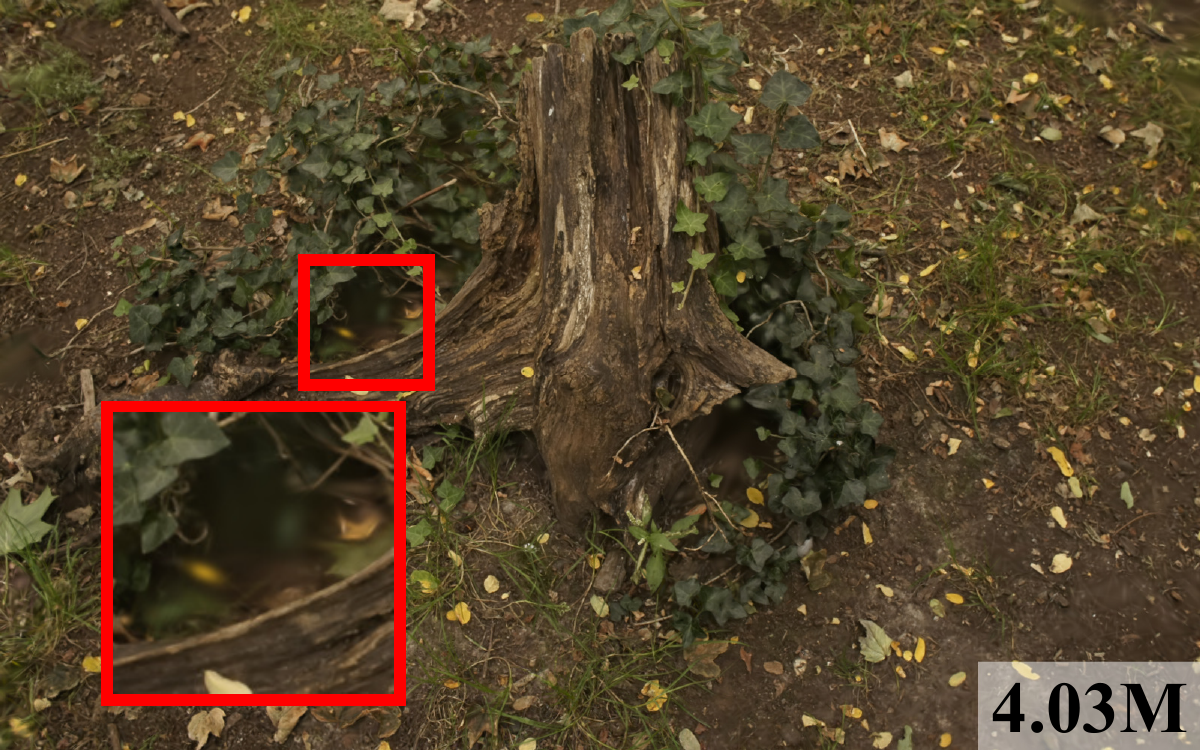}} &
        \fbox{\includegraphics[width=\imgwidth]{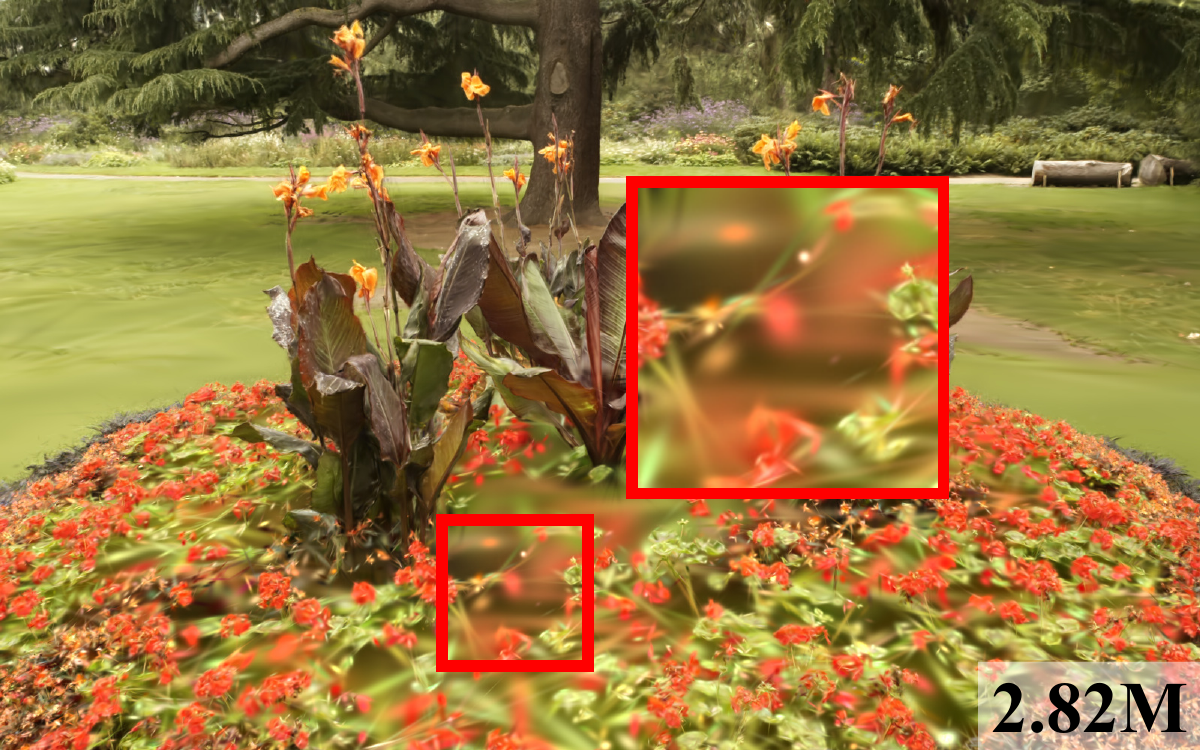}} &
        \fbox{\includegraphics[width=\imgwidth]{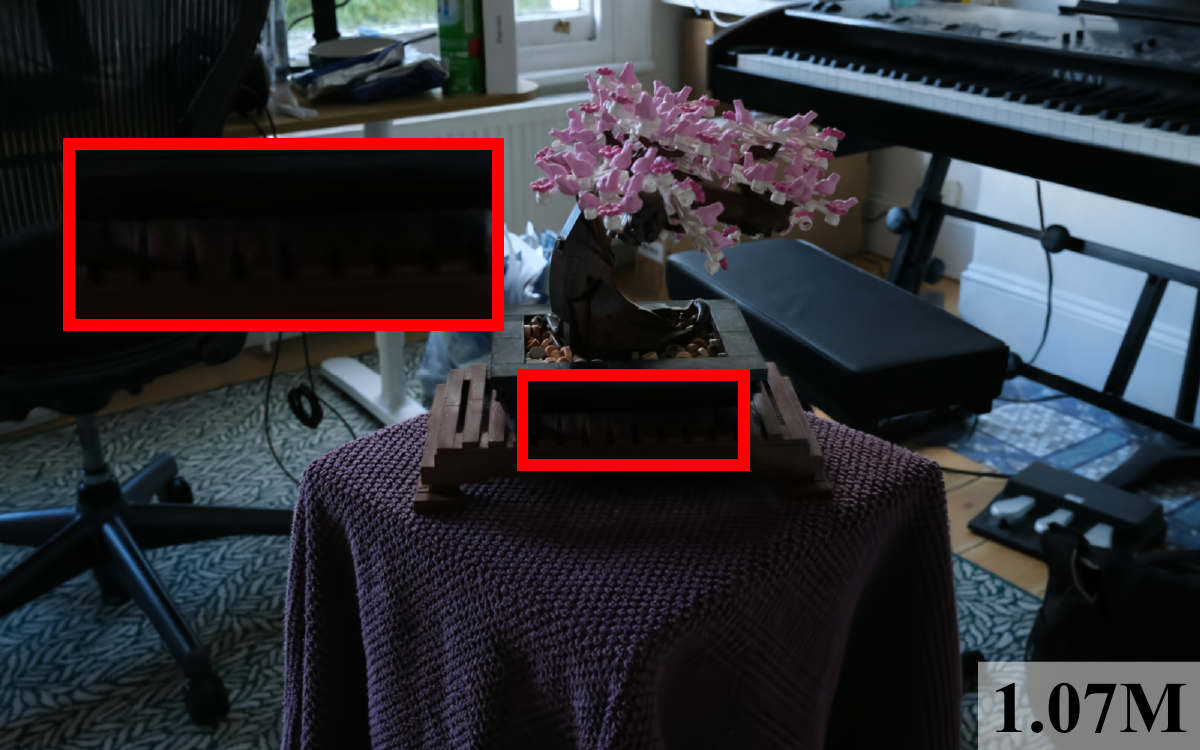}} &
        \fbox{\includegraphics[width=\imgwidth]{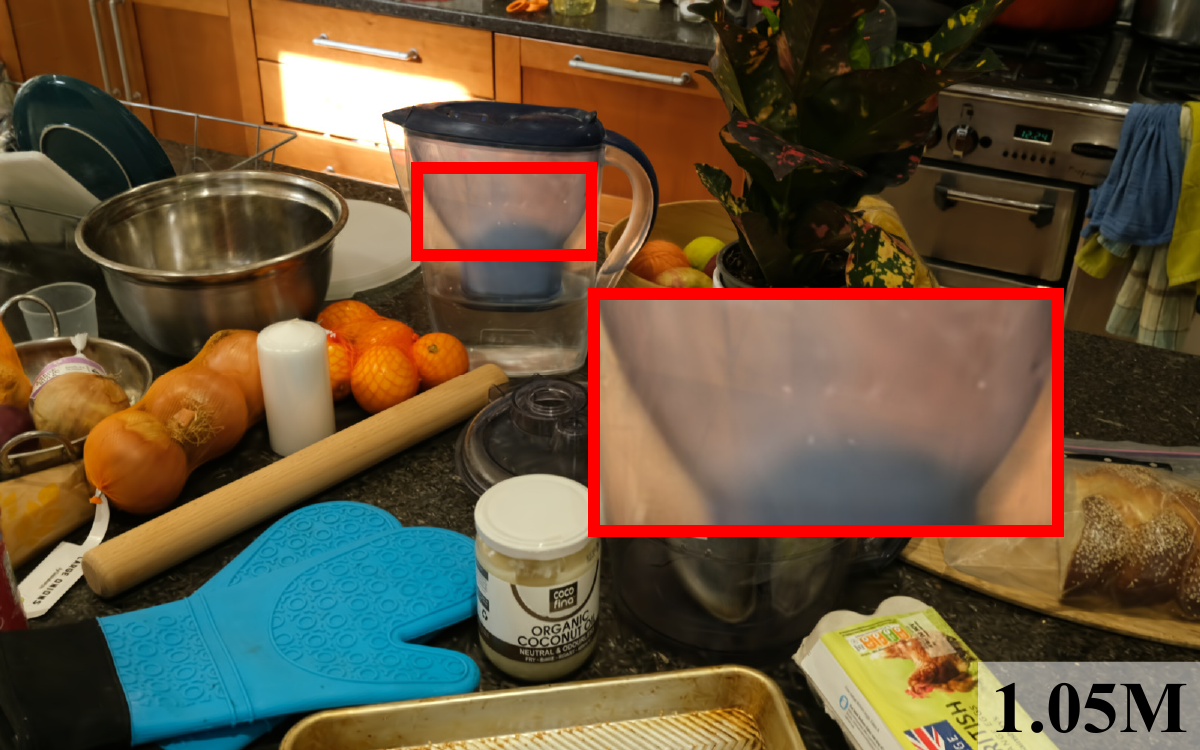}} &
        \fbox{\includegraphics[width=\imgwidth]{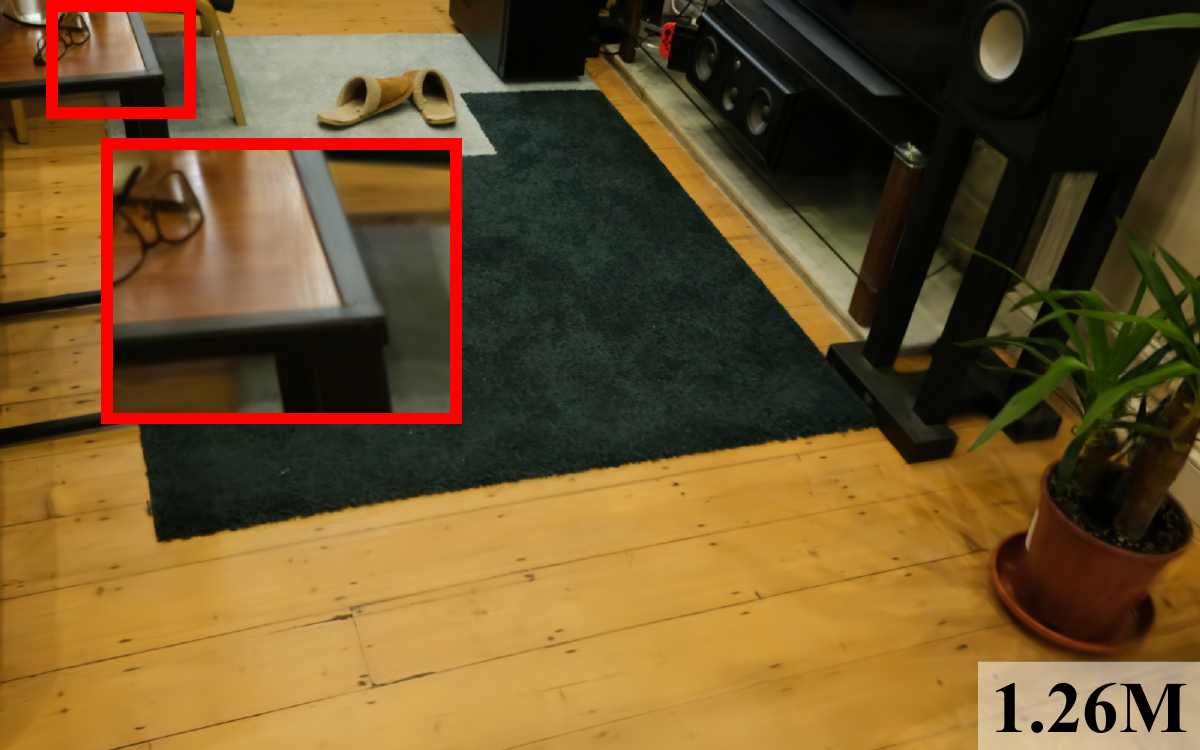}} &
        \fbox{\includegraphics[width=\imgwidth]{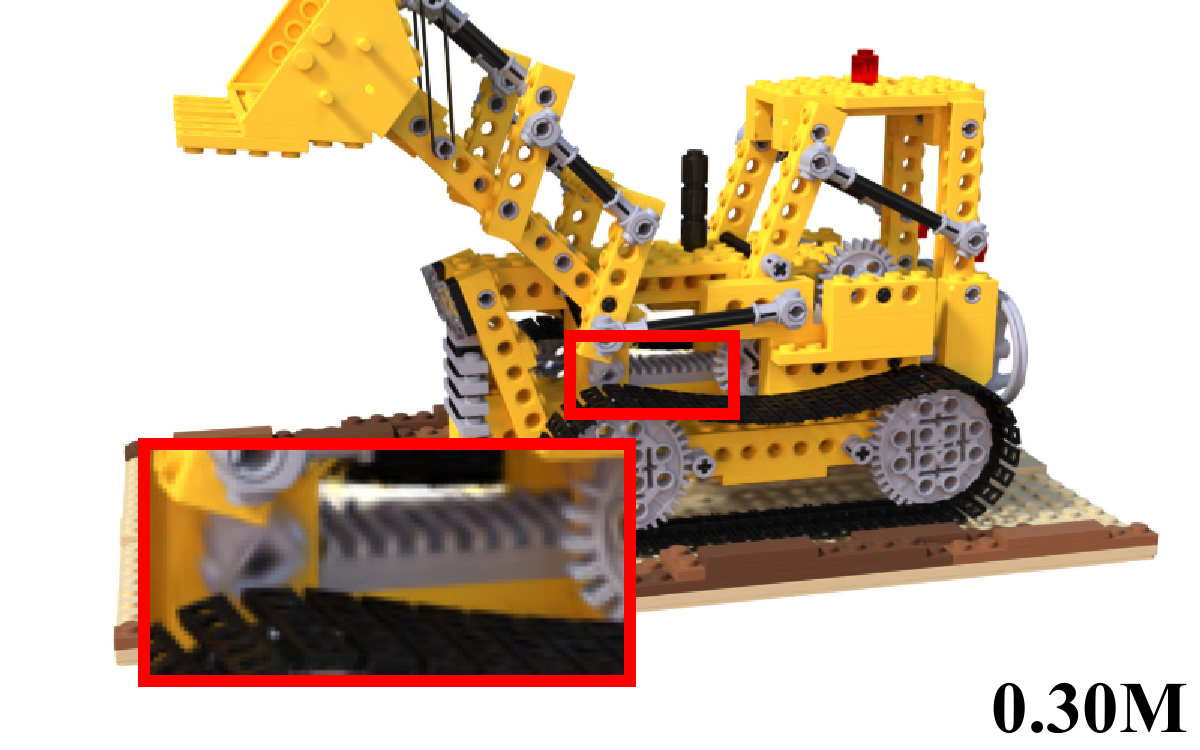}} \\

        \adjustbox{angle=90,lap=0.8em}{\scriptsize\textsf{\textbf{LP-3DGS}}} &
        \fbox{\includegraphics[width=\imgwidth]{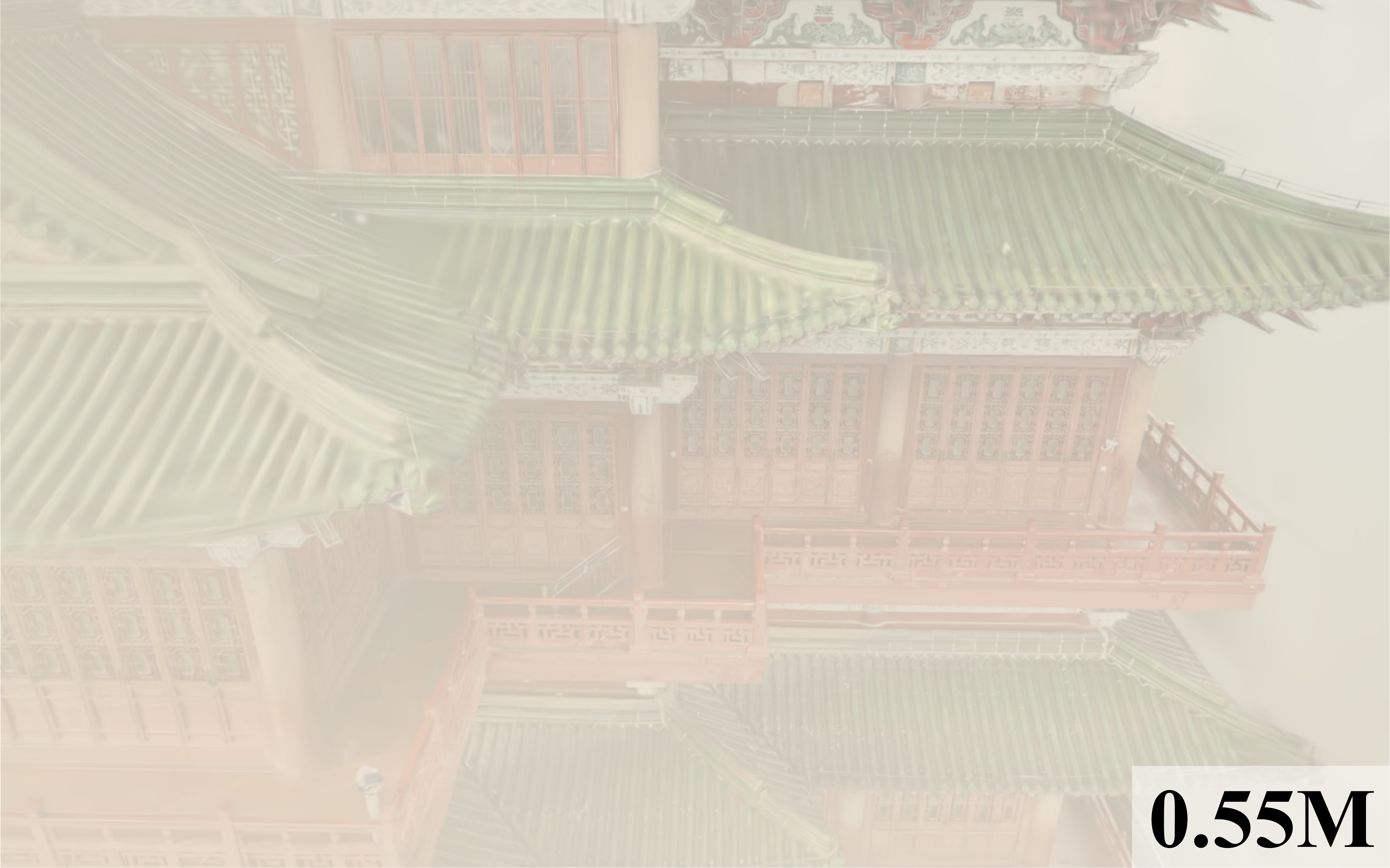}} &
        \fbox{\includegraphics[width=\imgwidth]{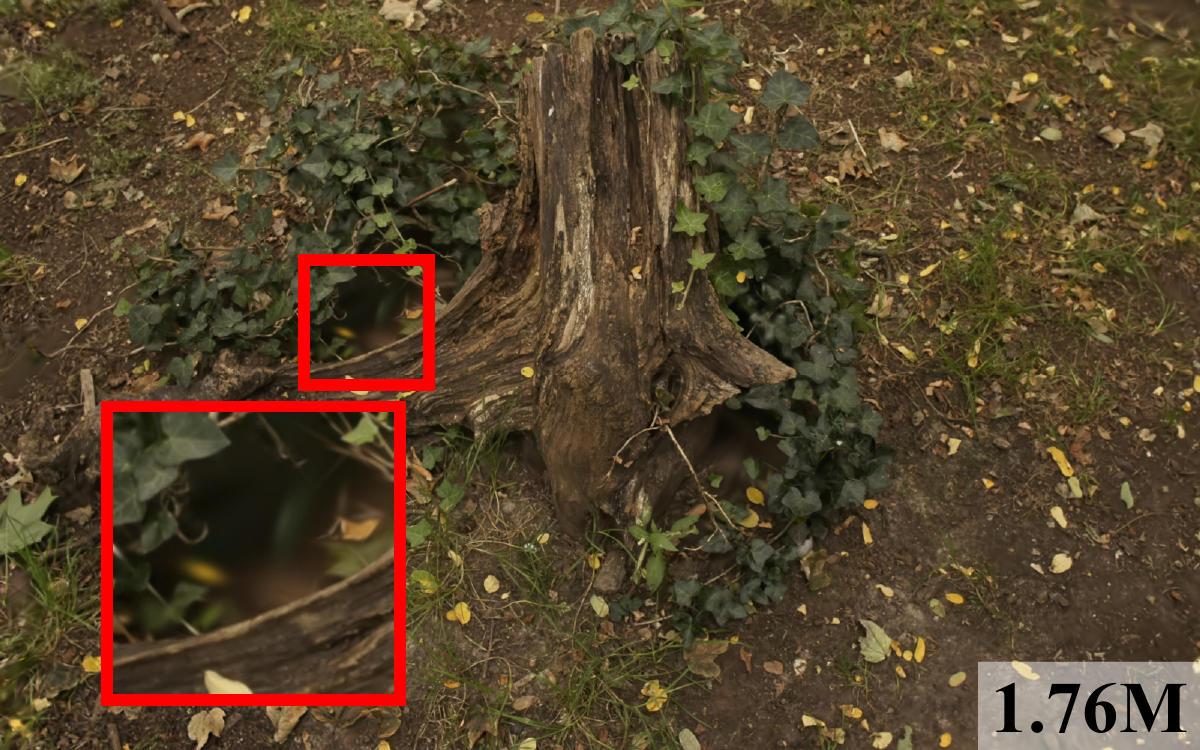}} &
        \fbox{\includegraphics[width=\imgwidth]{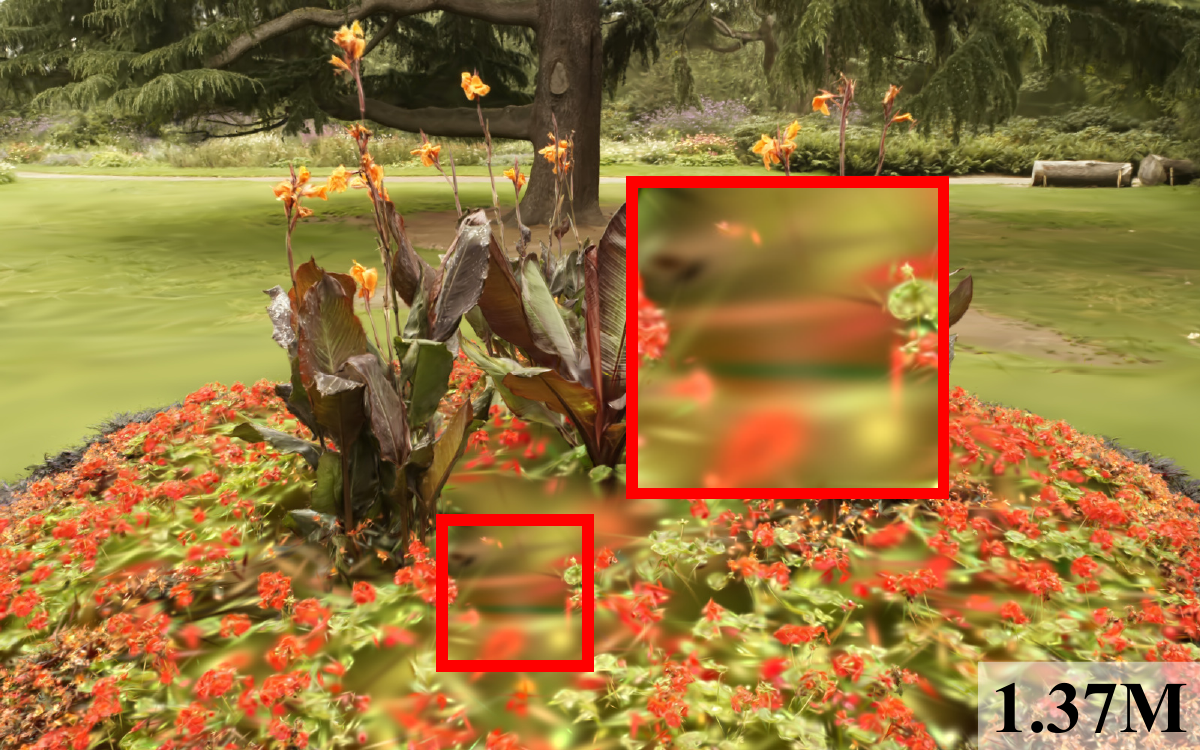}} &
        \fbox{\includegraphics[width=\imgwidth]{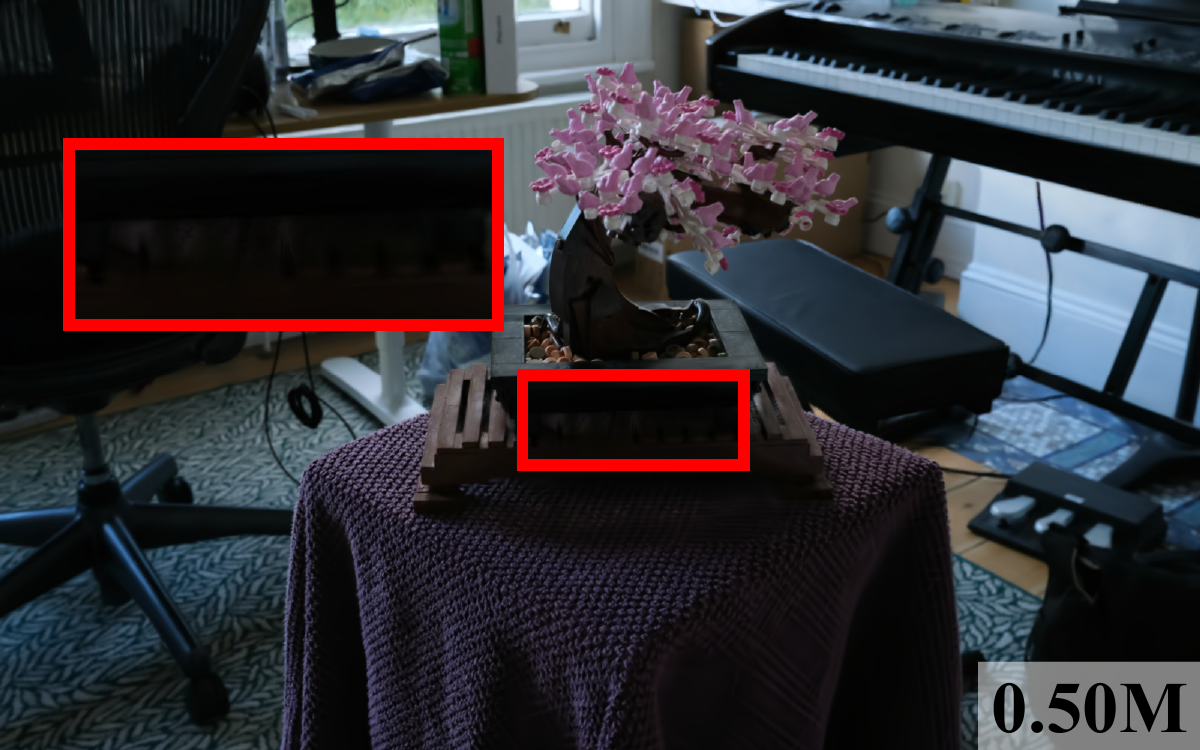}} &
        \fbox{\includegraphics[width=\imgwidth]{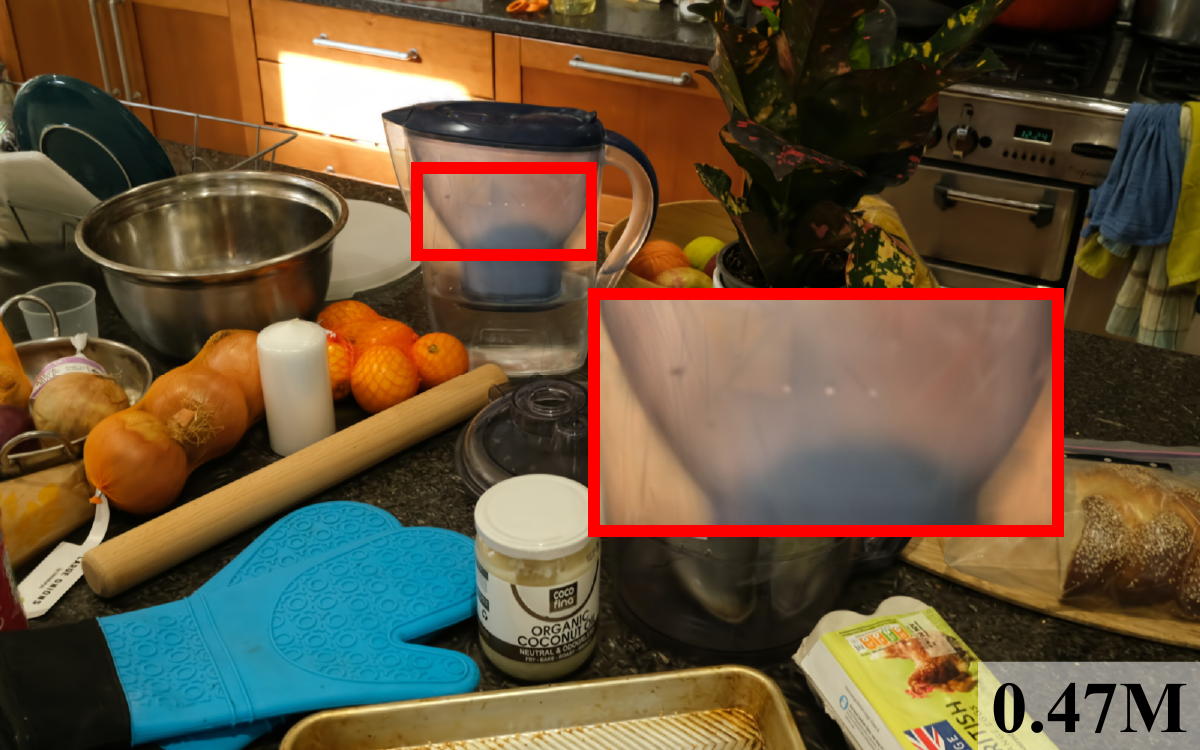}} &
        \fbox{\includegraphics[width=\imgwidth]{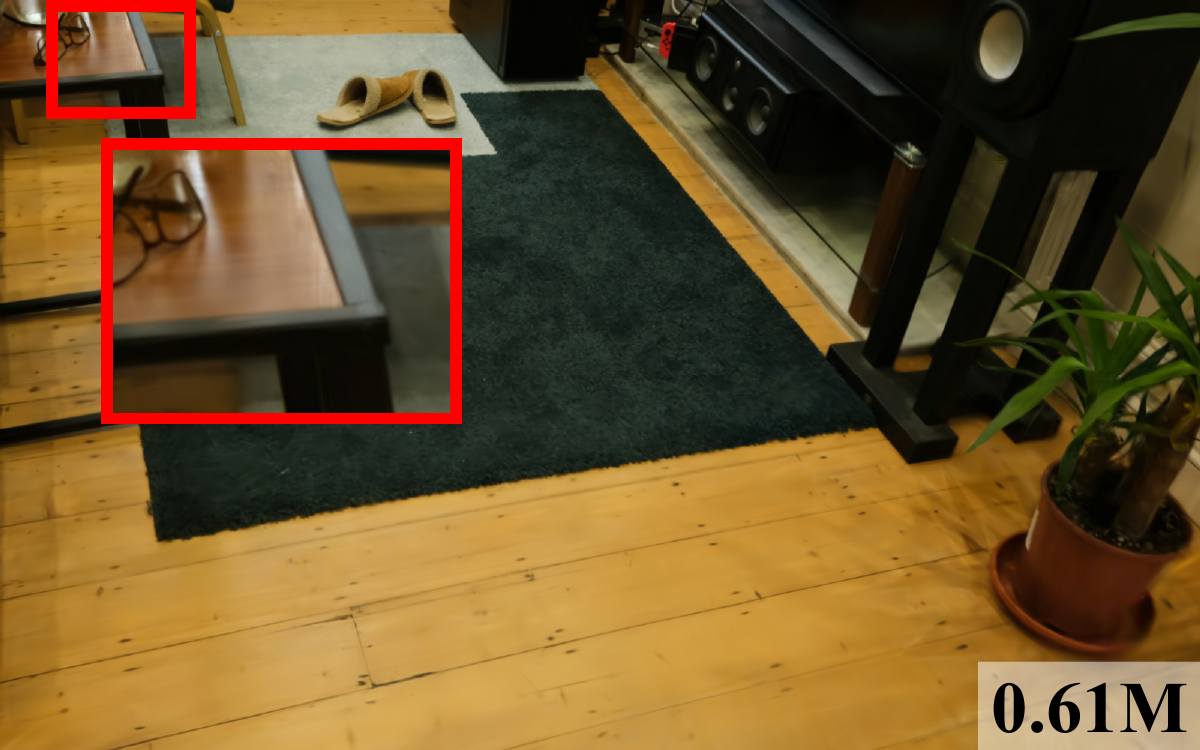}} &
        \fbox{\includegraphics[width=\imgwidth]{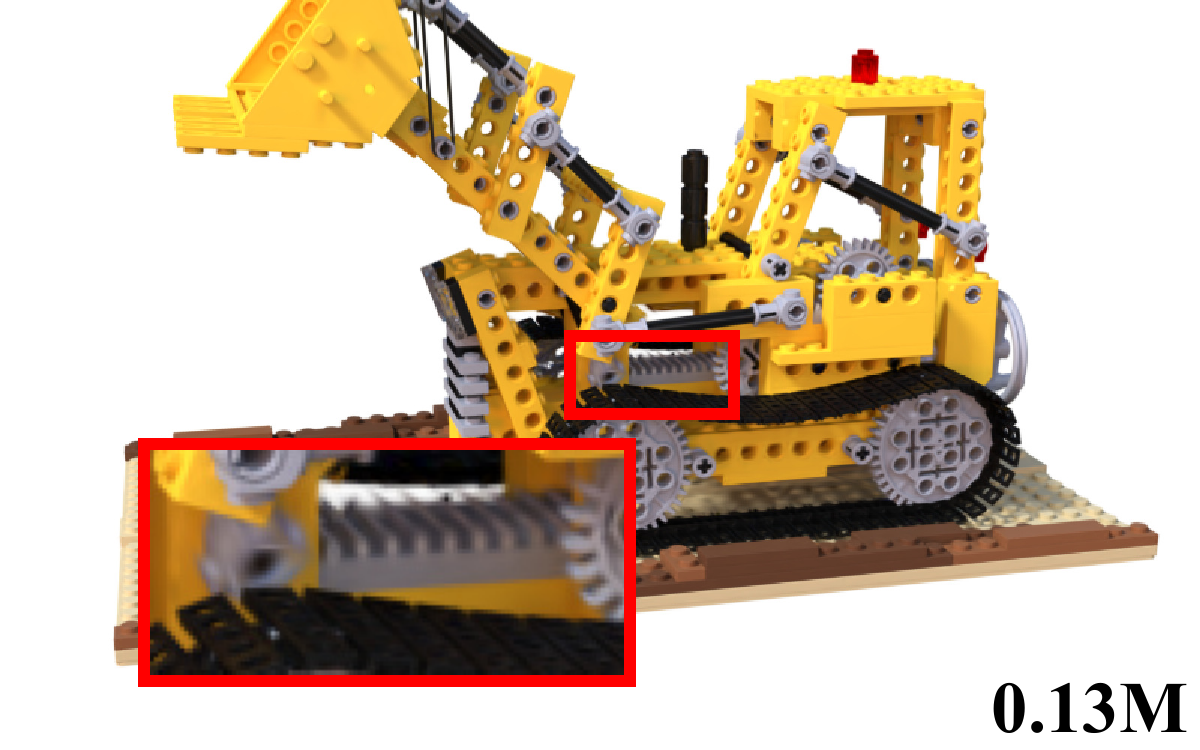}} \\

        \adjustbox{angle=90,lap=0.8em}{\scriptsize\textsf{\textbf{EAGLES}}} &
        \fbox{\includegraphics[width=\imgwidth]{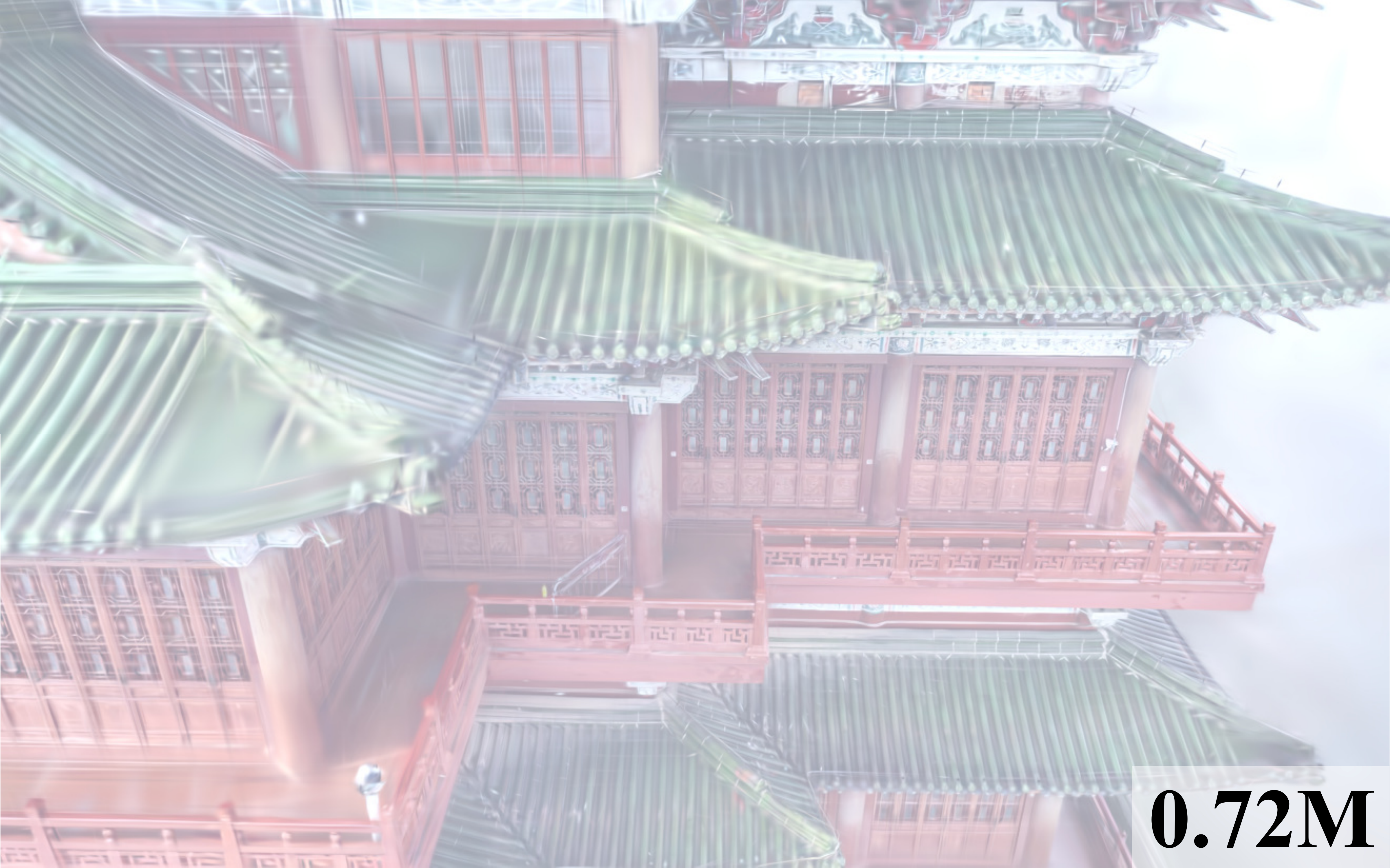}} &
        \fbox{\includegraphics[width=\imgwidth]{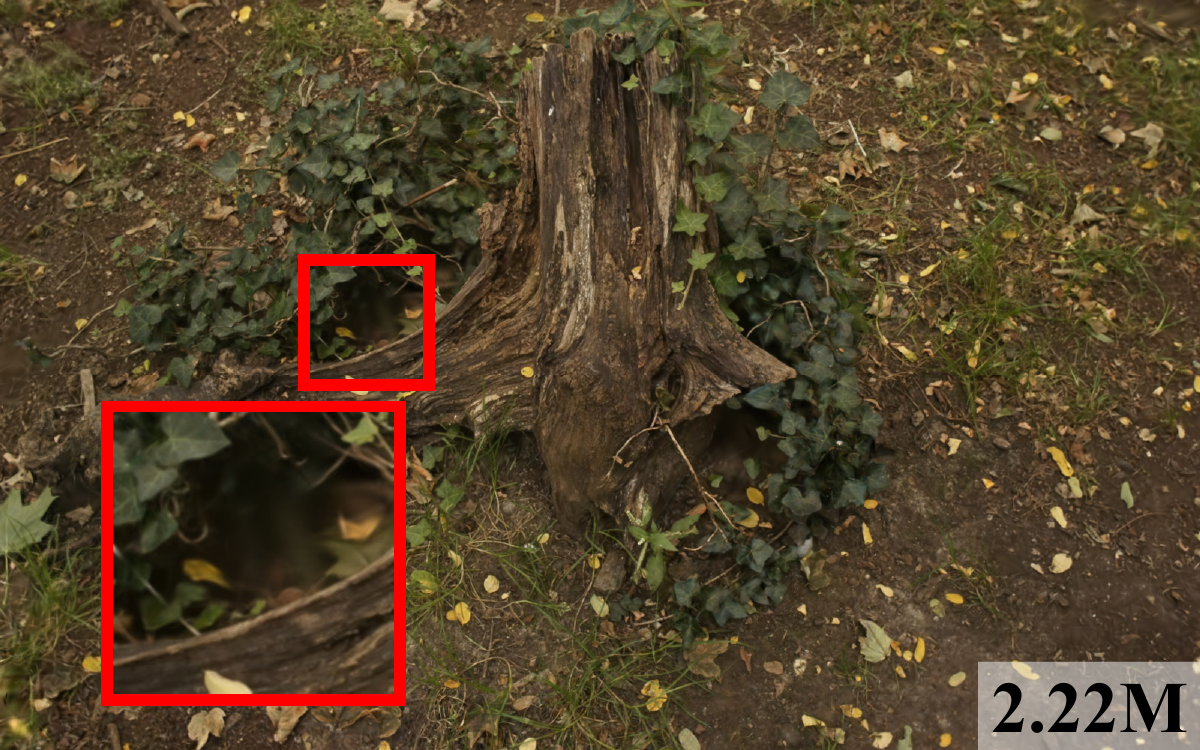}} &
        \fbox{\includegraphics[width=\imgwidth]{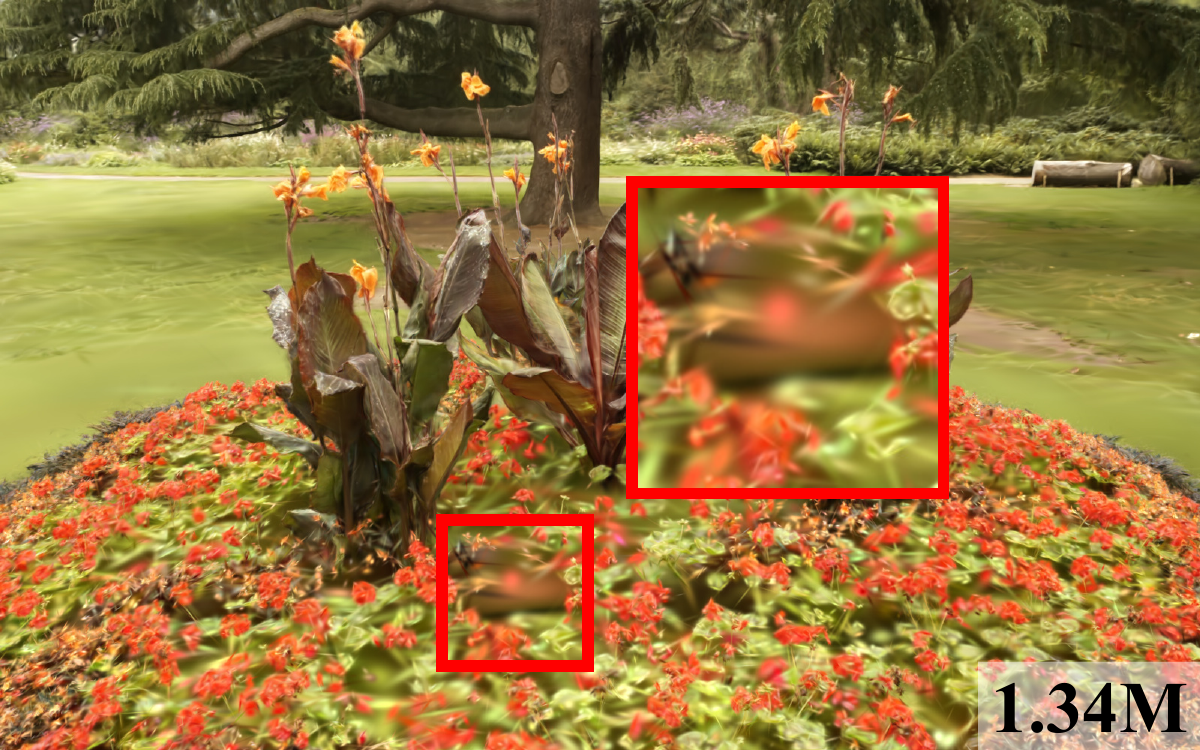}} &
        \fbox{\includegraphics[width=\imgwidth]{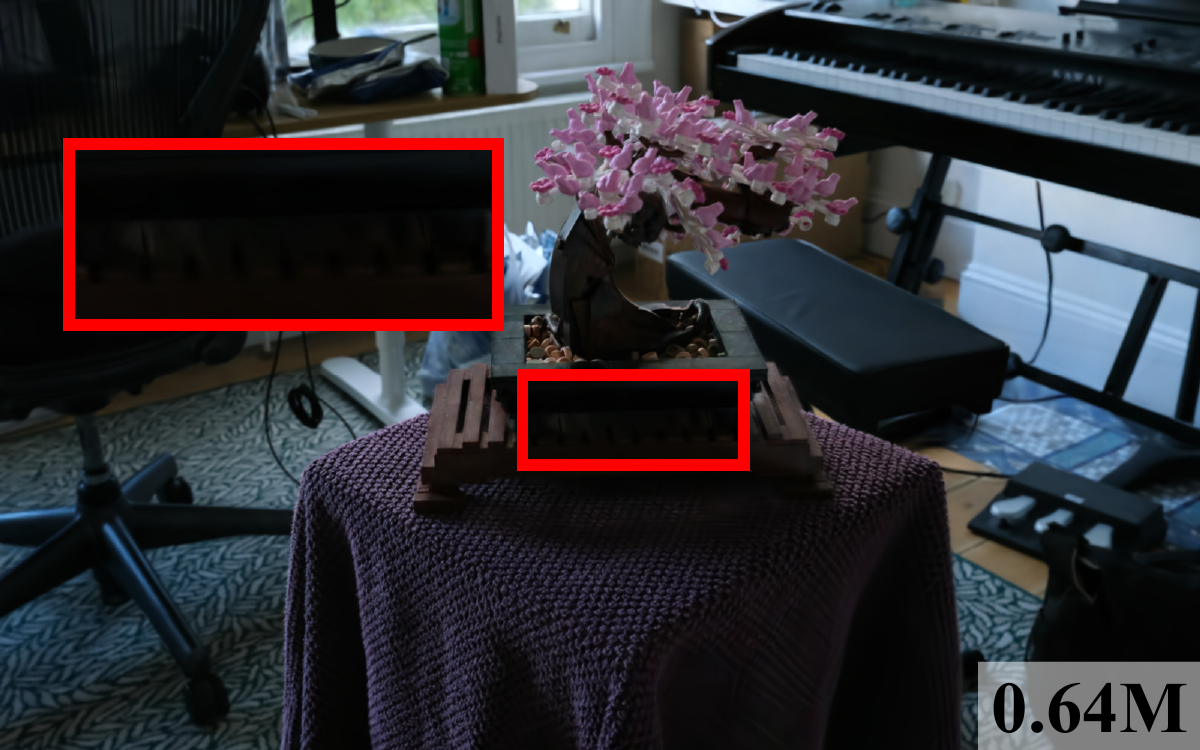}} &
        \fbox{\includegraphics[width=\imgwidth]{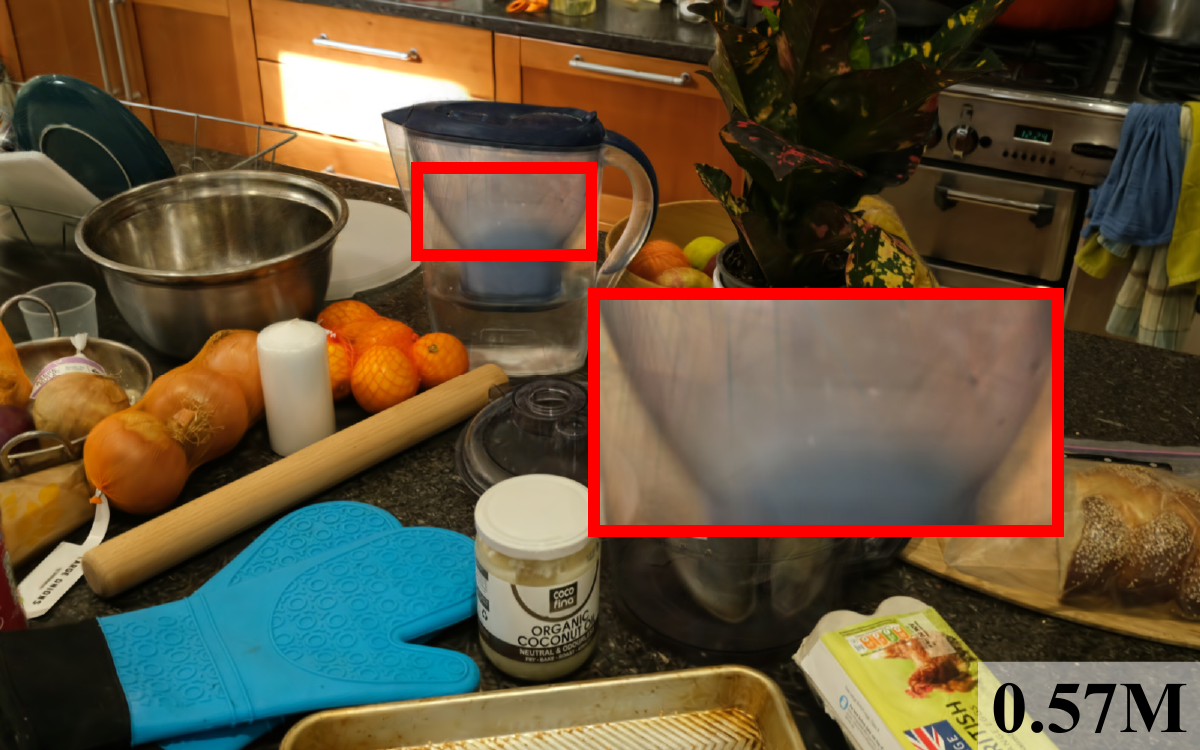}} &
        \fbox{\includegraphics[width=\imgwidth]{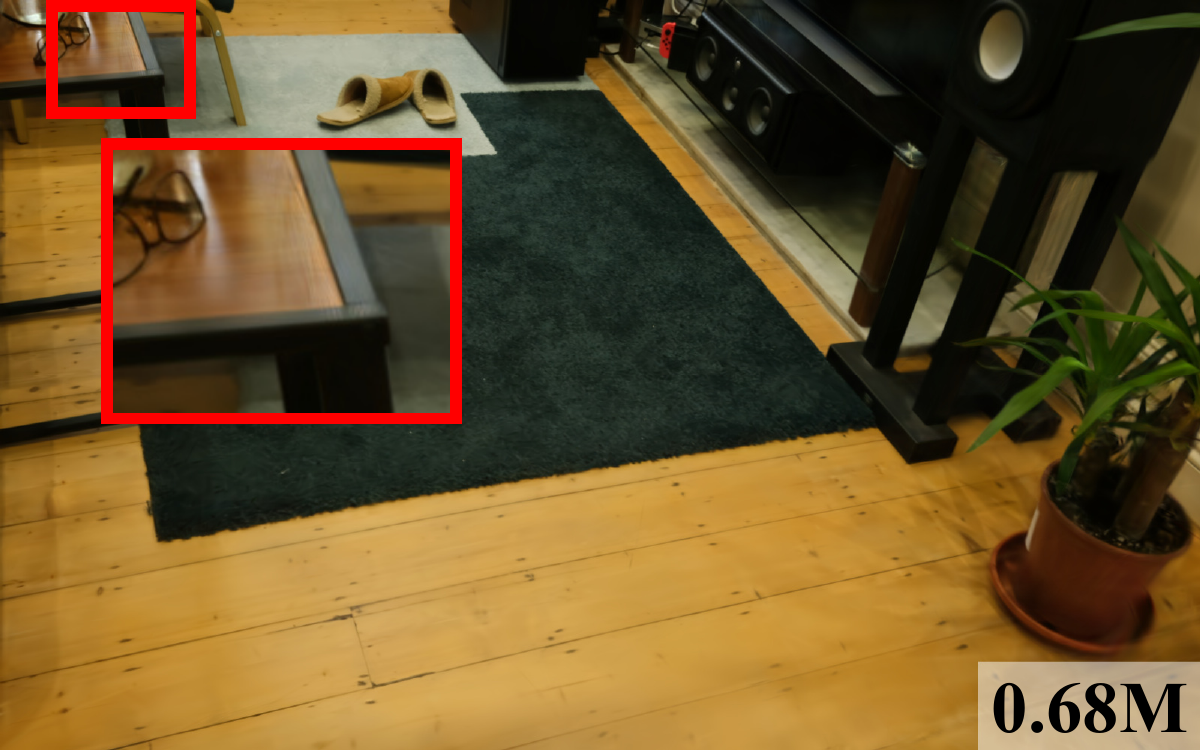}} &
        \fbox{\includegraphics[width=\imgwidth]{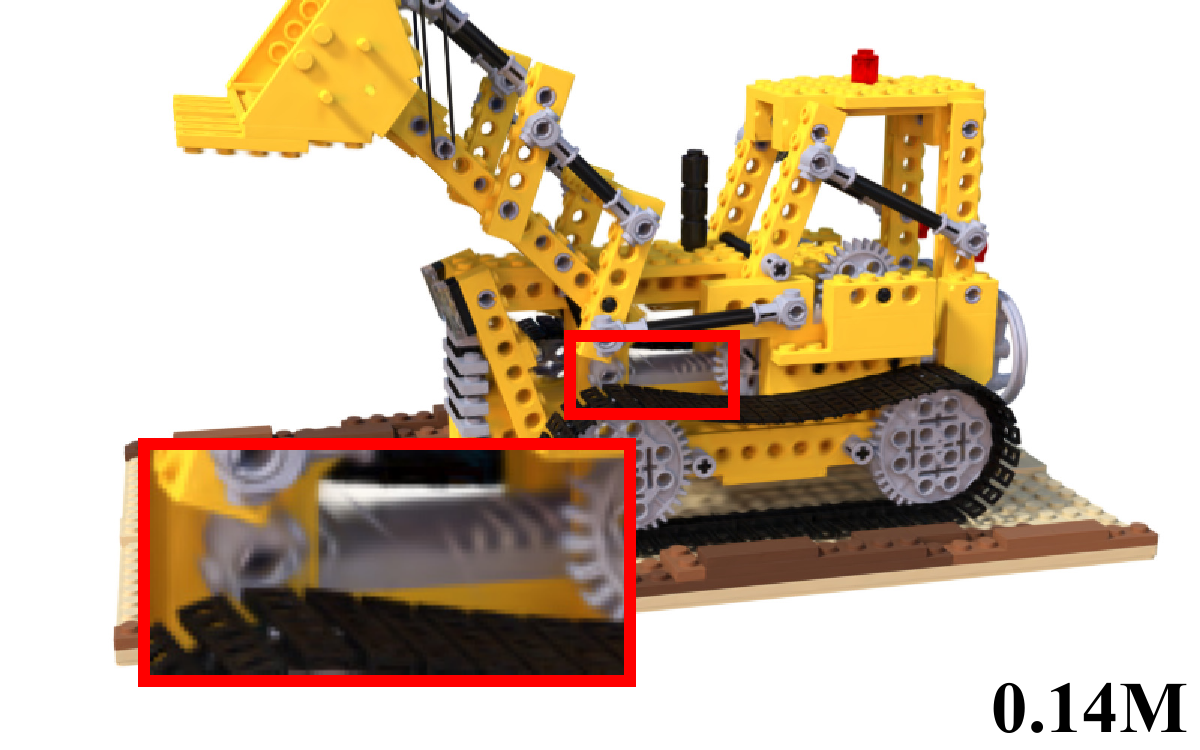}} \\

        & \fixedstack{TW-P-D (View 2) from}{GigaNVS~\cite{GigaNVS}} &
          \fixedstack{Stump from}{Mip-NeRF360~\cite{Mip-NeRF360}} &
          \fixedstack{Flowers from}{Mip-NeRF360~\cite{Mip-NeRF360}} &
          \fixedstack{Bonsai from}{Mip-NeRF360~\cite{Mip-NeRF360}} &
          \fixedstack{Counter from}{Mip-NeRF360~\cite{Mip-NeRF360}} &
          \fixedstack{Room (View 2) from}{Mip-NeRF360~\cite{Mip-NeRF360}} &
          \fixedstack{Lego from}{NeRF Synthetic~\cite{NeRF}}
    \end{tabular}

    \caption{NVS results on unseen test views across multi-scale scenes, comparing our method with $\lambda_\alpha\text{=3e-7}$, vanilla 3DGS~\cite{3DGS}, LP-3DGS~\cite{LP-3DGS} with a default pruning ratio of 0.6, and EAGLES~\cite{EAGLES}. Insets highlight key differences, and the number of Gaussians used by each model is shown in the lower-right corner for reference.
    \textit{Please zoom in to see details.}}
    \label{fig:qualitative_comparison}
\end{figure*}

\textit{First}, as shown in Fig.~\ref{fig:opacity_comp}, introducing the opacity $L_1$ regularization and jointly optimizing it with the reconstruction loss $\mathcal{L}_{\mathrm{RGB}}$ encourages a balance between sparsity and fidelity, leading to a richer distribution of intermediate opacity values rather than a near-binary pattern near 0 or 1. 
According to Eq.~\eqref{eq:alpha-blending}, these intermediate opacities increase the likelihood of each Gaussian contributing to color accumulation across multiple views rather than being occluded by front layers, thereby improving Gaussian utilization efficiency. The resulting richer Gaussian blending also enhances the model’s representational capacity and potentially leads to higher rendering quality.

\textit{Second}, as illustrated in Fig.~\ref{fig:size_evolution_comp}, alternating between uniform splitting and optimization drives a progressive refinement along the frequency dimension: Large Gaussians generated early in training capture global low-frequency structures, while later-introduced smaller Gaussians refine local high-frequency details, forming a coarse-to-fine process that improves Gaussian utilization efficiency. At the same time, each child Gaussian inherits spatial and appearance parameters from its parent, ensuring cross-stage stability and naturally forming a top-down hierarchical structure. This inheritance further introduces a helpful inductive bias: smaller Gaussians, potentially with lower visibility or sparser supervision, are initialized from well-constrained parents with higher visibility, enabling them to maintain reasonable accuracy even under limited supervision.

\textit{Finally}, as demonstrated in Fig.~\ref{fig:self_correcting_mechanism}, the \textit{opacity-based sparsification} mechanism can self-correct the over-splitting issue caused by \textit{uniform splitting}, making the structure more adaptive. According to Eq.~\eqref{eq:final_loss}, new Gaussians that fail to effectively reduce $\mathcal{L}_{\mathrm{RGB}}$ after splitting are gradually suppressed by the $L_1$ opacity regularization, which continuously lowers their $\alpha$ values until they are pruned, guiding the representation back to a sparser configuration with lower overall loss. In contrast, Gaussians that contribute to reducing $\mathcal{L}_{\mathrm{RGB}}$ are retained. By maximizing the improvement of $\mathcal{L}_{\mathrm{RGB}}$, this process adaptively concentrates the Gaussians in structurally or texturally complex areas, thereby enhancing the Gaussian utilization efficiency.

\subsubsection{Qualitative Analysis} 
Fig.~\ref{fig:qualitative_comparison} shows qualitative comparisons between our method and baselines on unseen test views, spanning a variety of scenes from compact objects to large-scale outdoor scenes. The results align with quantitative evaluations: ControlGS achieves higher rendering quality with fewer Gaussians across different scenes. 

In sparsely observed or occluded regions, ControlGS effectively reconstructs fine details and reduces ``floater'' artifacts, such as the grass under the bench in \textit{Bicycle}, vegetation-covered areas in \textit{Stump}, and the edge regions in \textit{TW-P-D~(View~2)}, where other methods fail.
In indoor scenes like \textit{Playroom}, \textit{Room~(View~1)}, and \textit{Counter}, it accurately recovers furnishings and structures with consistent geometry and appearance. 
In complex textured scenes such as \textit{TW-P-D~(View~1)}, \textit{Train} and \textit{Flowers}, it preserves clarity in high-frequency regions like patterns, gravel and vegetation, demonstrating strong texture reconstruction. 
In scenes with complex lighting, such as \textit{Kitchen}, \textit{Bonsai}, \textit{Room~(View~2)}, it accurately reconstructs surface reflections and indirect illumination while maintaining consistent lighting across viewpoints.
In object-scale scenes like \textit{Drums} and \textit{Lego}, it maintains uniform sharpness across the object, avoiding local blurring and distortion.

\subsection{Ablation Study}

To evaluate the contributions of key components, we conducted ablation experiments by individually replacing \textit{uniform splitting}, \textit{attribute inheritance}, and \textit{opacity-based sparsification} in our full method with the corresponding modules from vanilla 3DGS~\cite{3DGS}. All experiments shared identical training configurations, with $\lambda_\alpha$ set to $\text{3e-7}$. Results are summarized in Table~\ref{tab:ablation} and Fig.~\ref{fig:ablation_qualitative}.

\subsubsection{Uniform Splitting}

Replacing uniform splitting with the original \textit{clone-and-split} heuristic results in insufficient Gaussians and introduces local blurring, as it relies on noise-sensitive local criteria that often cause over- or under-reconstruction. Moreover, unlike our octree-style strategy that splits each Gaussian into eight evenly spaced children, 3DGS’s binary splitting produces sparse and uneven candidates, limiting the search space for global optimization.

\subsubsection{Attribute Inheritance}

Replacing attribute inheritance with the original 3DGS initialization, which randomly samples child positions within the parent’s extent and copies opacity, causes a significant inflation in the number of Gaussians and introduces noticeable ``floater'' artifacts. Our octree-style inheritance scheme places children evenly within the parent’s extent, efficiently covering larger areas with fewer and less clustered Gaussians. In contrast, random sampling can yield overly dense or sparse regions, where multiple Gaussians occupy space representable by one. These Gaussians also retain non-negligible contributions, resisting pruning via opacity-based sparsification and introducing redundancy.

\subsubsection{Opacity-based Sparsification}
Replacing opacity-based sparsification with the original opacity-reset-based pruning causes out-of-memory (OOM) conditions. Without effective pruning, low-contributing Gaussians accumulate and, combined with uniform splitting, lead to uncontrolled growth that degrades training efficiency, increases memory consumption, and ultimately causes training to fail.

\newcommand{\barChartF}[1]{%
  \smash{\pgfmathsetmacro{\barlen}{#1/2.63*0.915}%
  \begin{tikzpicture}[baseline=(base)] %
    \node[inner sep=0pt] (base) at (0.9cm, 0.053cm) {};
    \fill[green] (0,0) rectangle (\barlen cm, 0.27cm);
    \node[anchor=center, font=\scriptsize] at (0.45cm, 0.14cm) {#1};
  \end{tikzpicture}%
}}

\begin{table}[h]
\caption{Quantitative ablation results on the Mip-NeRF360 dataset, following the format of Table~\ref{tab:scene_comparison}.}
\label{tab:ablation}
\centering
\scriptsize
\setlength{\tabcolsep}{0pt}
\renewcommand{\arraystretch}{0.75}
\setlength{\extrarowheight}{3pt}
\begin{tabular}{C{3.5cm}|C{0.9cm}C{0.9cm}C{0.9cm}C{1.0cm}}
\toprule
\textbf{Dataset} & \multicolumn{4}{c}{\textbf{Mip-NeRF360 (Indoor/Outdoor)}} \\
Method $\mid$ Metrics & PSNR$\uparrow$ & SSIM$\uparrow$ & LPIPS$\downarrow$ & Num(M) \\
\midrule
3DGS~\cite{3DGS} & \cellcolor{second}27.63 & \cellcolor{second}0.814 & \cellcolor{third}0.222 & \barChartF{2.63} \\
w/o Uniform Splitting & \cellcolor{third}27.55 & \cellcolor{third}0.803 & 0.253 & \barChartF{0.51} \\
w/o Attribute Inheritance & 27.36 & \cellcolor{first}\textbf{0.821} & \cellcolor{first}\textbf{0.202} & \barChartF{1.13} \\
w/o Opacity-based Sparsification & \multicolumn{4}{c}{\cellcolor{gray!35}OOM} \\
\textbf{Full} & \cellcolor{first}\textbf{27.90} & \cellcolor{first}\textbf{0.821} & \cellcolor{second}0.221 & \barChartF{0.83} \\
\bottomrule
\end{tabular}
\vspace{-11pt}
\end{table}

\begin{figure}[h]
  \centering
  \setlength{\tabcolsep}{1pt}  
  \setlength{\fboxrule}{0.2pt}
  \setlength{\fboxsep}{-0.2pt}
  \renewcommand{\arraystretch}{0.6} 
  \newcommand{\imgw}{0.32\linewidth}

  \begin{tabular}{@{}ccc@{}}
    \fbox{\includegraphics[width=\imgw]{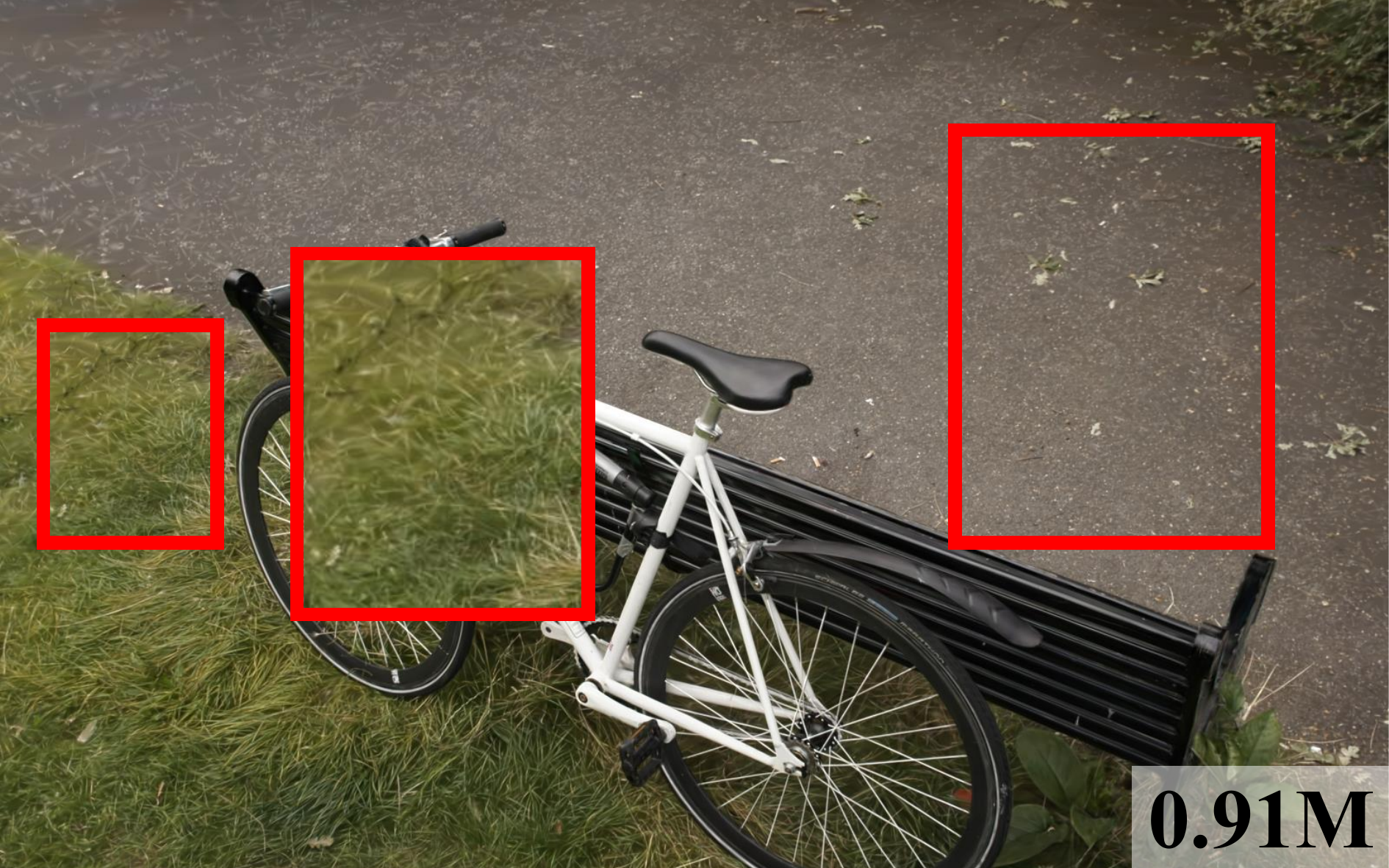}} &
    \fbox{\includegraphics[width=\imgw]{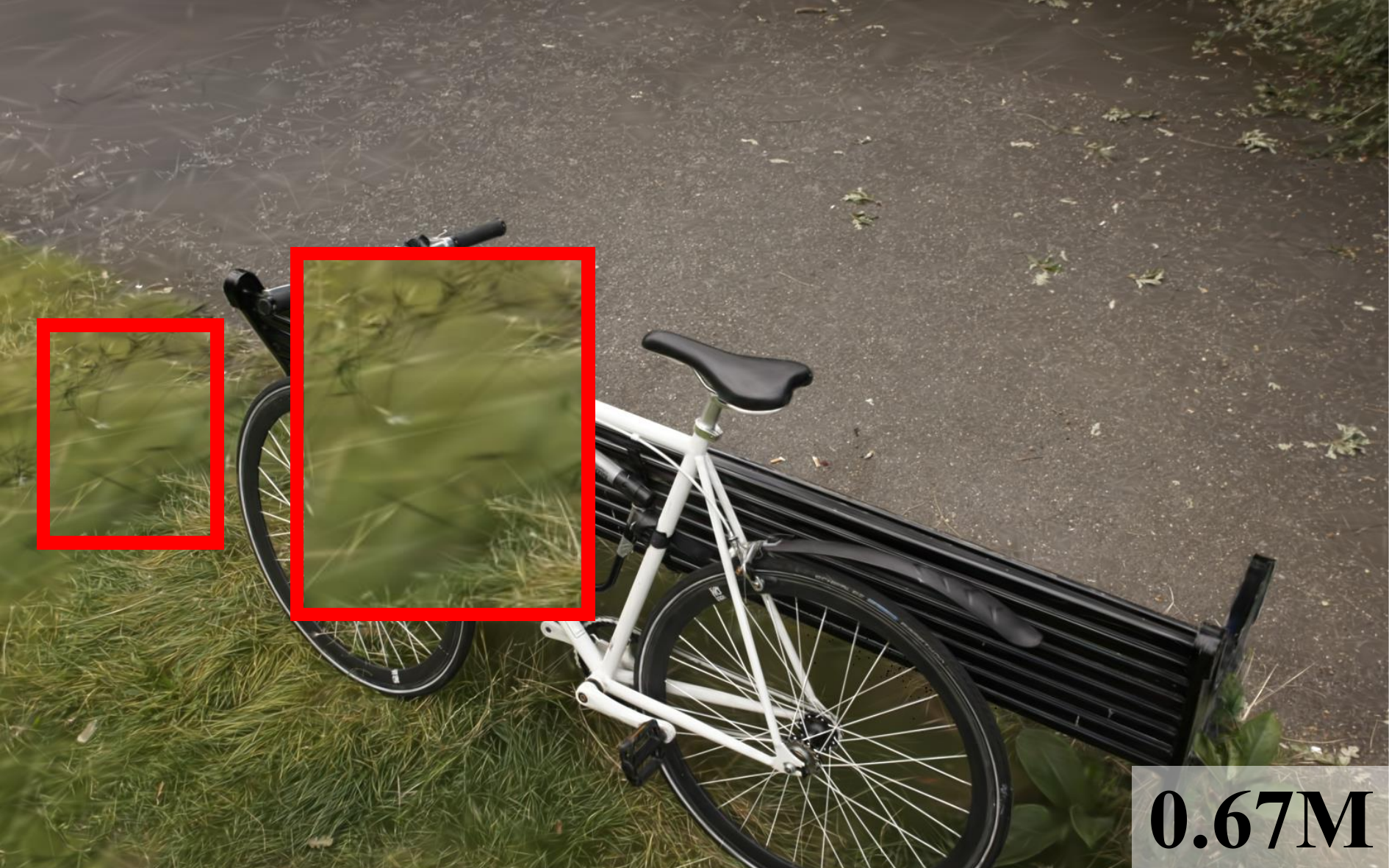}} &
    \fbox{\includegraphics[width=\imgw]{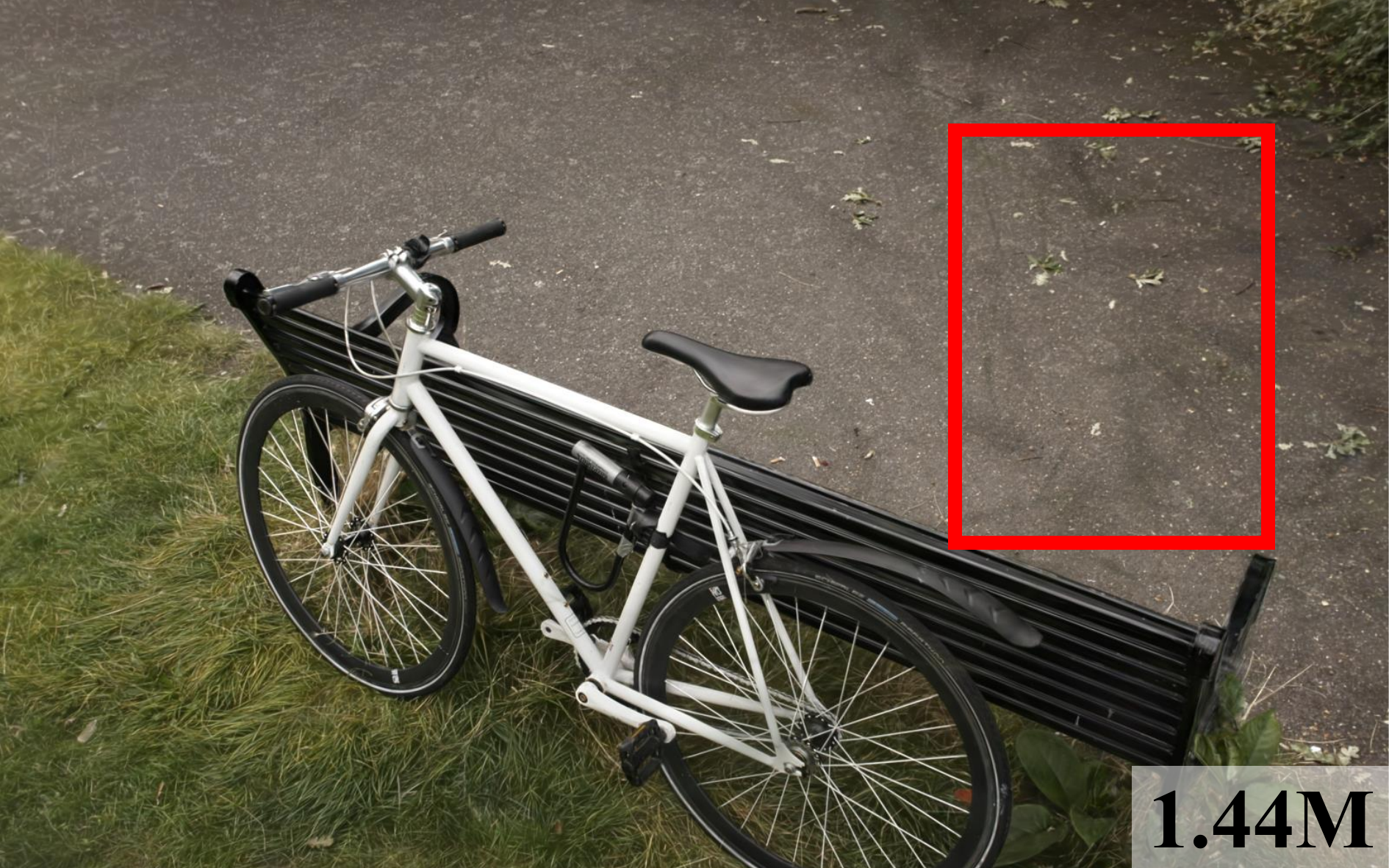}} \\
    \fbox{\includegraphics[width=\imgw]{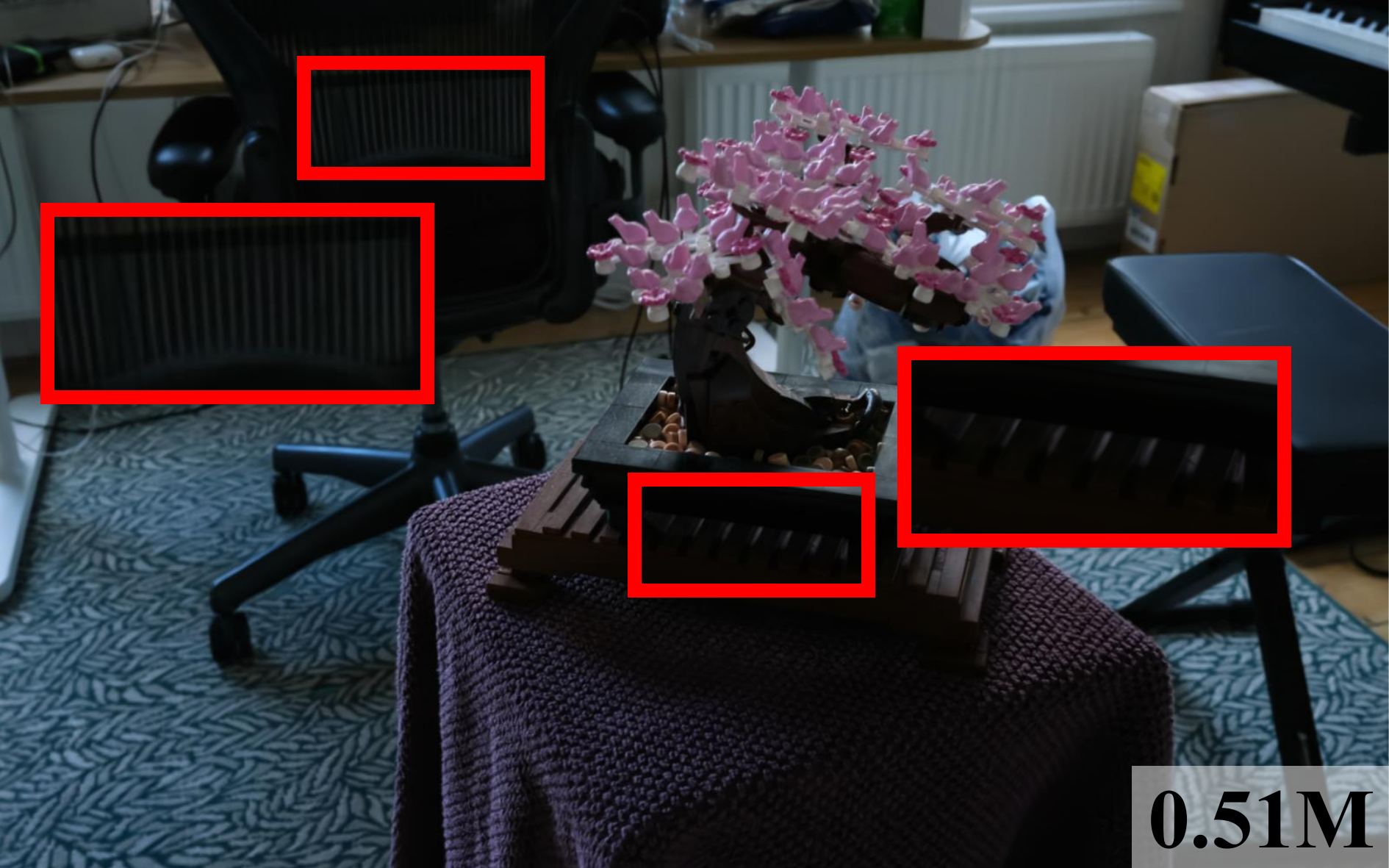}} &
    \fbox{\includegraphics[width=\imgw]{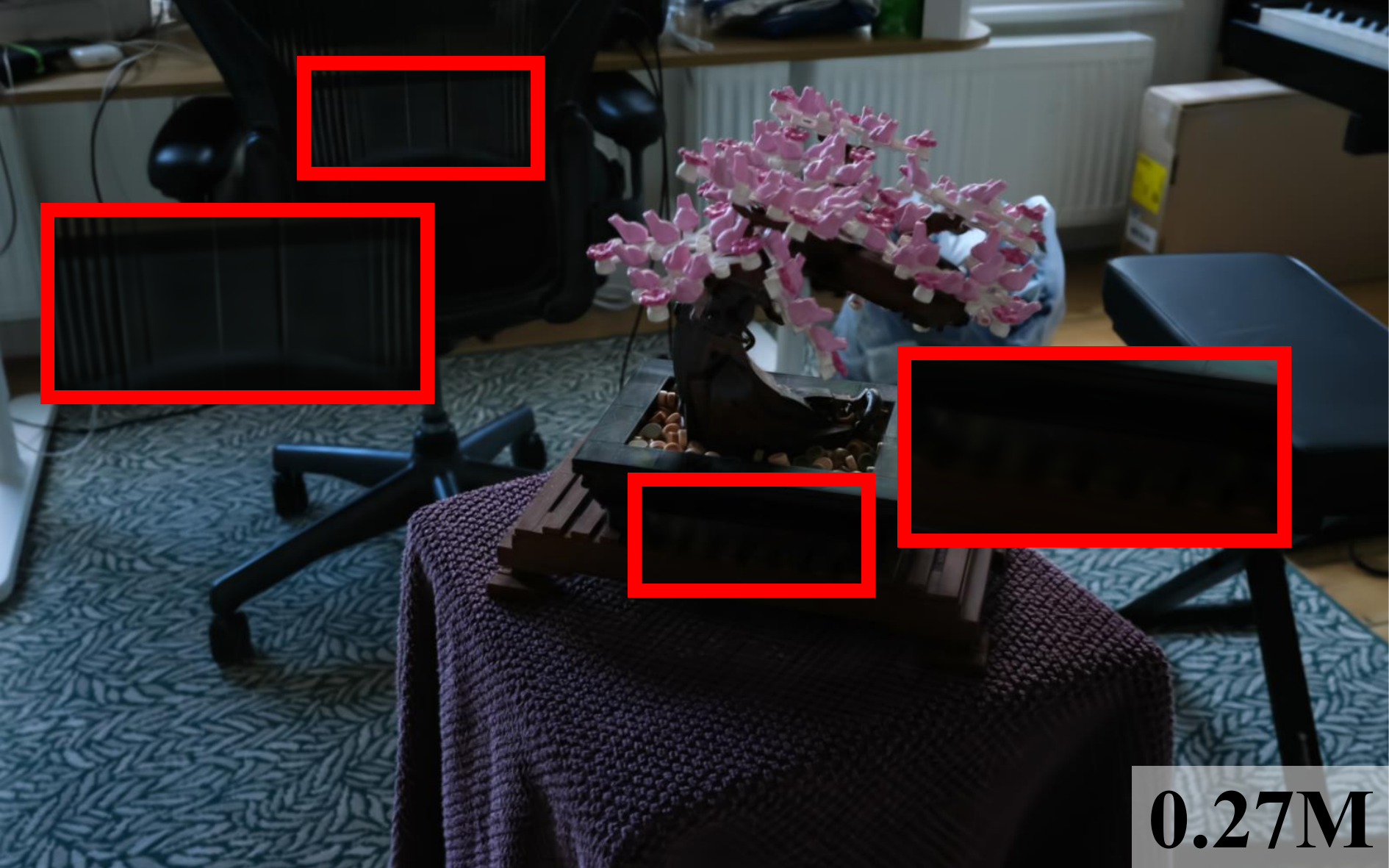}} &
    \fbox{\includegraphics[width=\imgw]{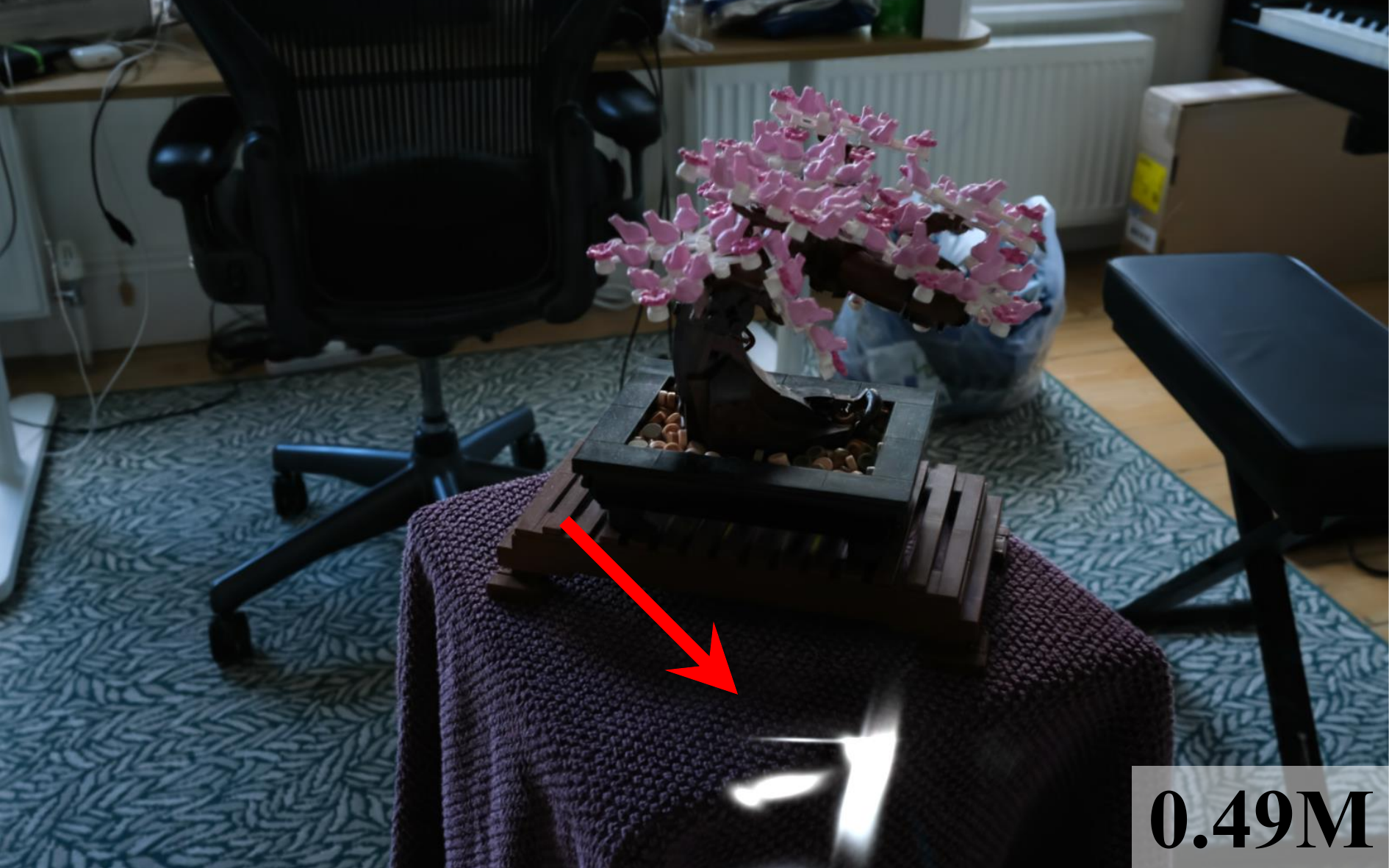}} \\
    [-2pt]
    {\scriptsize \textbf{Full}} &
    {\scriptsize w/o Uniform Splitting} &
    {\scriptsize w/o Attribute Inheritance} \\
  \end{tabular}
  \vspace{-3pt}
  \caption{Qualitative ablation results. 
  Each column shows a different variant: the full model, without uniform splitting, and without attribute inheritance. \textit{Please zoom in to see details.}}
  \label{fig:ablation_qualitative}
  \vspace{-6pt}
\end{figure}

\section{Discussion and Conclusions}

In this work, we present ControlGS, a cross-scene consistent structural compression control framework for 3DGS automated deployment. 
It introduces two core mechanisms: \textit{uniform splitting}, which expands Gaussians without local heuristics, and \textit{opacity-based sparsification}, which prunes Gaussians based only  on their rendering contribution. These mechanisms are unified by a single control hyperparameter~$\lambda_\alpha$, enabling continuous, scene-agnostic, and highly responsive preference control between structural compactness and fidelity, without scene-specific tuning.
Compared to potential competing methods, ControlGS also pushes beyond the Gaussian count--rendering quality Pareto frontier.

Future research directions include: (1) integrating attribute compression for greater compactness; (2) extending to dynamic scenes and video reconstruction; and (3) leveraging ControlGS as a general framework for broader scene representation methods based on explicit primitives.

In summary, ControlGS offers a controllable, broadly applicable, high-performance solution for 3DGS structural compression control. With a simple interface, consistent cross-scene behavior, and strong performance, it enhances the real-world deployment of 3DGS models across varying hardware and bandwidth constraints.

\balance
\bibliographystyle{IEEEtran}
\bibliography{references}

\end{document}